\pdfoutput=1
\documentclass[10pt,twocolumn,letterpaper]{article}
\usepackage[pagenumbers]{cvpr}
\def\confName{CVPR}
\def\confYear{2027}
\usepackage{booktabs}
\usepackage{placeins}
\usepackage{amssymb}
\usepackage{xcolor}
\usepackage{wasysym}
\usepackage[T1]{fontenc}
\usepackage{microtype}
\usepackage{mathtools}
\usepackage{array,tabularx,multirow,colortbl}
\usepackage{enumitem}
\definecolor{realblue}{HTML}{24658A}
\definecolor{realgreen}{HTML}{34786B}
\definecolor{realorange}{HTML}{AB6B25}
\colorlet{catA}{realblue!14}
\colorlet{catB}{realorange!16}
\colorlet{catC}{realgreen!16}
\colorlet{catAll}{black!7}
\colorlet{rowshade}{realblue!4}
\colorlet{ftshade}{realorange!17}
\definecolor{zgreen}{HTML}{4E9A2E}
\colorlet{zebra}{zgreen!12}
\colorlet{ftink}{realorange!85!black}
\usepackage[pagebackref,breaklinks,colorlinks,allcolors=realblue]{hyperref}
\newcommand{\name}{RoboClock}
\hypersetup{pdftitle={\name: A Unified Benchmark for Time-Aware Robotic Intelligence},pdfauthor={Yujie Zhao, Anthony L. Hu, Zimeng Huang, Zhijing Wu, Luobin Wang, Richard Li, Shrivas Sudharsan, Henrik I Christensen, Jishen Zhao},pdfsubject={Preprint}}
\newcolumntype{Y}{>{\raggedright\arraybackslash}X}
\setlist[itemize]{leftmargin=*,itemsep=2pt,topsep=3pt}
\title{\name: A Unified Benchmark for Time-Aware Robotic Intelligence}
\author{%
Yujie Zhao$^{1}$\quad Anthony L. Hu$^{1}$\quad Zimeng Huang$^{1}$\quad Zhijing Wu$^{1}$\\
Luobin Wang$^{1}$\quad Richard Li$^{2}$\quad Shrivas Sudharsan$^{1}$\\
Henrik I Christensen$^{1}$\quad Jishen Zhao$^{1}$\\[4pt]
$^{1}$University of California, San Diego\\
$^{2}$Fremont Union High School District
}
\makeatletter
\newcommand{\cYes}{\textcolor{green!55!black}{\ensuremath{\checkmark}}}
\newcommand{\cNo}{\textcolor{black}{\ensuremath{\times}}}
\newcommand{\cPartial}{\textcolor{black}{\ensuremath{\LEFTcircle}}}
\g@addto@macro\@maketitle{\begin{minipage}{\textwidth}
\centering
\includegraphics[width=0.9\textwidth]{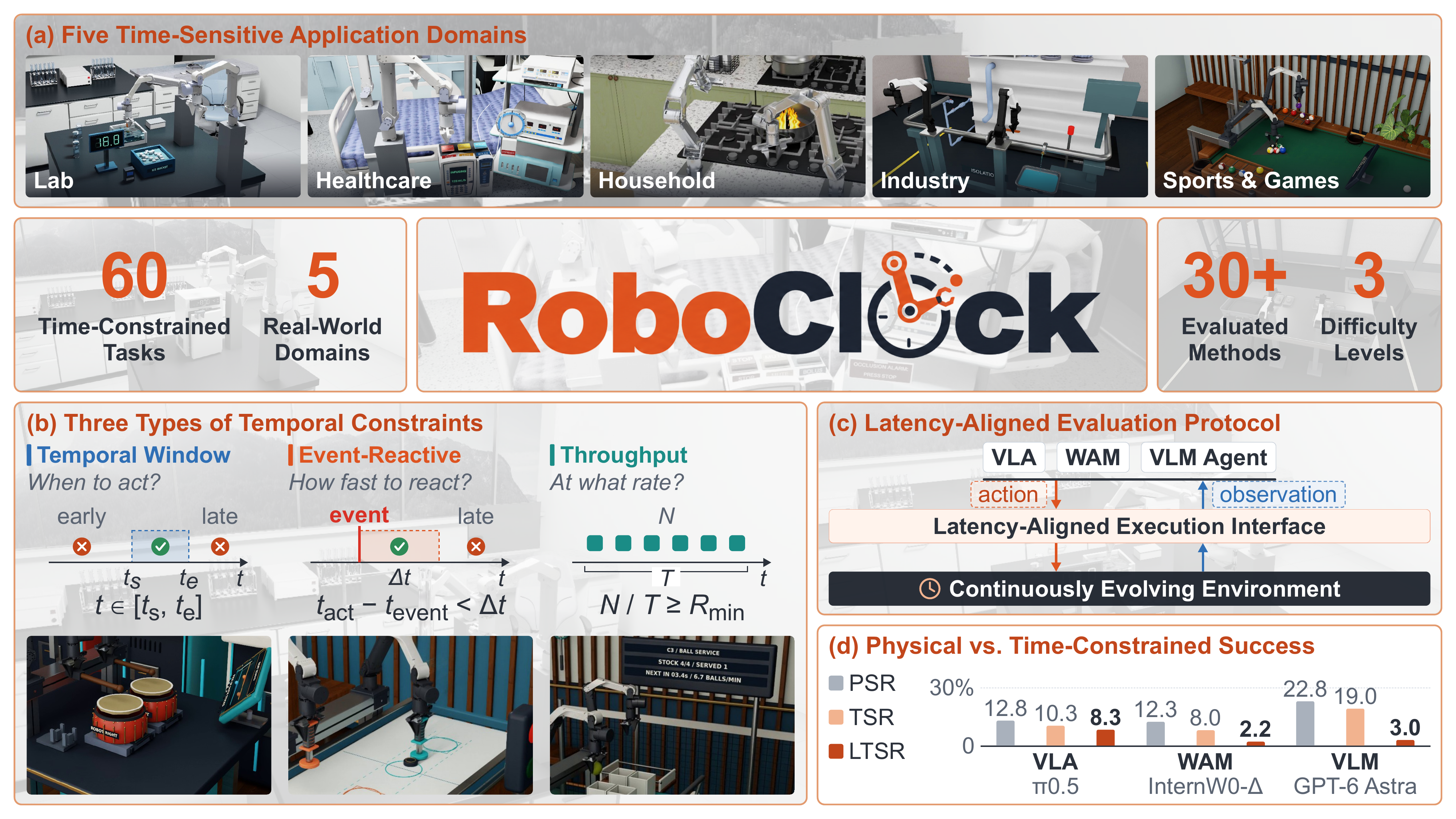}
\captionsetup{hypcap=false}
\captionof{figure}{\textbf{Overview of \name{}.} 60 time-critical manipulation tasks across five application domains, evaluated through a general time interface: any agent (VLA, WAM, or VLM agent) receives observations with the environment clock, and its inference latency advances environment time before its action takes effect.}
\label{fig:time-alignment}
\end{minipage}
\par\vspace{12pt}}
\makeatother
\selectfont
\begin{document}
\maketitle
\begin{abstract}
Time is a fundamental dimension of robotic intelligence:
a robot may fail despite physically correct actions if it
acts outside a valid time window, responds too slowly to
an event, or fails to meet required throughput. Yet existing robotic benchmarks primarily emphasize
spatial correctness and logical task completion,
largely overlooking the temporal constraints of
real-world applications. We present \textbf{RoboClock}, a unified benchmark for time-aware robotic manipulation featuring 60 tasks across five application domains. Directly mirroring these failure modes, RoboClock structures tasks around Three Types of Temporal Constraints: \textit{Temporal Window}, \textit{Event-Reactive}, and \textit{Throughput}. Furthermore, RoboClock incorporates a Latency-Aligned evaluation protocol that continuously advances environment time during model inference and control-interface delays, paired with time-sensitive metrics.
Evaluating over 30 methods, we find that the
best-performing approach is the trained $\pi_{0.5}$, yet it
achieves only 8.3\% temporal success even on Easy-level
tasks, with persistent challenges in
generalization and transfer.
These findings reveal fundamental limitations in
current robotic systems and highlight the need to
jointly optimize physical competence, temporal
reasoning, and execution latency.
\end{abstract}

\section{Introduction}
\label{sec:intro}

Time is a fundamental dimension of physical intelligence. 
Operating in dynamic, continuously evolving environments, embodied agents require not only spatial reasoning and manipulation skills, but also timely decision-making and execution. 
In the physical world, a correct action becomes futile if delivered too late -- a reality reflected in a recent survey of 800 industry leaders~\cite{intel2026robotics}, where over 75\% view sub-100-ms latency as essential for critical robotic applications. 
Meanwhile, embodied AI has advanced through VLA models, WAMs, and VLM-based agents~\cite{black2025pi05,kim2026cosmos,chu2026openrua}, reaching 98.5\%~\cite{kim2026cosmos} on LIBERO~\cite{liu2023libero} and 68.87\% on RoboDojo~\cite{chen2026robodojo}.

Yet a critical gap separates these benchmark triumphs from real-world deployment: \textbf{physical time}. 
Conventional evaluation protocols typically 
assess whether a task is eventually completed in the correct logical sequence, decoupling the agent's computational and actuation speed from environmental dynamics. 
By implicitly pausing the simulated world during policy computation, this turn-based paradigm artificially \textbf{inflates task success rates through two unrealistic relaxations}: first, it permits agentic systems to exploit evaluation loops via exhaustive trial-and-error recovery, action rollbacks, and repeated invocation retries; second, it accommodates unbounded inference latency, allowing foundation-model policies to extensively deliberate at each step.

\begin{figure}[t]
  \centering
  \includegraphics[width=\linewidth]{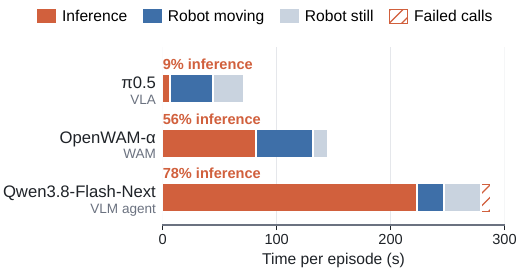}
  \caption{\textbf{Time breakdown per episode across \name{} tasks.}}
  \label{fig:intro-breakdown}
\end{figure}

Empirical measurements expose the severity of these hidden temporal costs. 
A recent NVIDIA study reveals that model inference alone can consume up to 97.8\% of total execution time in language-model-based policies, stretching a 5.2-second physical demonstration into nearly four minutes~\cite{yang2026astranomy}; similarly, ORA~\cite{yang2026ora} achieves 83.5\% success on LIBERO~\cite{liu2023libero} at the cost of a 13.8-minute median episode duration~\cite{yang2026ora}. 
Our profiling across \name{} tasks (Fig.~\ref{fig:intro-breakdown}) confirms that this overhead diverges starkly across policy families: while the reactive VLA $\pi_{0.5}$ completes episodes in roughly 70 seconds with only 9\% spent on inference, inference time surges to 56\% for the WAM OpenWAM-$\alpha$ ($\sim$145\,s total) and 78\% for the VLM agent Qwen3.8-Flash-Next ($\sim$285\,s total), compounded by repeated failed calls and prolonged periods when the robot stands still.
In the real world, time is more than an efficiency metric: agent speed relative to object dynamics decides task success.

\begin{table*}[t]
\centering
\caption{\textbf{Comparison of temporal robotic benchmarks.} \cYes: supported; \cNo: not supported; \cPartial: partial or indirect.}
\label{tab:temporal_comparison}

\scriptsize
\setlength{\tabcolsep}{5pt}
\renewcommand{\arraystretch}{1.18}

\begin{tabular}{@{}l c ccc cc@{}}
\toprule
& &
\multicolumn{3}{c}{\textbf{Temporal Requirements}} &
\multicolumn{2}{c}{\textbf{Evaluation}} \\
\cmidrule(lr){3-5}
\cmidrule(l){6-7}

\textbf{Benchmark} &
\textbf{\#Tasks} &
\shortstack{\textbf{Execution}\\\textbf{Windows}} &
\shortstack{\textbf{Response}\\\textbf{Deadlines}} &
\shortstack{\textbf{Throughput}\\\textbf{Targets}} &
\shortstack{\textbf{Inference Delay}\\\textbf{Included}} &
\shortstack{\textbf{Success}\\\textbf{Breakdown}} \\
\midrule

ReflexVLA (ReflexBench)~\cite{chen2026reflex}
& 6
& \cYes & \cPartial & \cNo
& \cYes & \cNo \\

DynamicManip~\cite{liao2026dynamicmanip}
& 5+4
& \cYes & \cPartial & \cNo
& \cYes & \cNo \\

DynamicVLA (DOM)~\cite{xie2026dynamicvla}
& 16$^{*}$
& \cYes & \cNo & \cNo
& \cPartial & \cNo \\

DOMINO~\cite{fang2026domino}
& 35
& \cYes & \cNo & \cNo
& \cNo & \cNo \\

RoboDyna~\cite{li2026robodyna}
& 30
& \cYes & \cYes & \cNo
& \cYes & \cNo \\

LIBERO-Industry~\cite{sung2026liberoindustry}
& 3$^{\dagger}$
& \cYes & \cNo & \cPartial
& \cNo & \cNo \\

SafeManip (v1)~\cite{huang2026safemanip}
& 50
& \cNo & \cNo & \cNo
& \cNo & \cNo \\

\midrule

\textbf{RoboClock (Ours)}
& \textbf{60}
& \cYes & \cYes & \cYes
& \cYes & \cYes \\

\bottomrule
\end{tabular}

\vspace{1mm}
\parbox{\textwidth}{\scriptsize
\textbf{Notes.}
SafeManip (v1) uses 50 RoboCasa365 household tasks.
\textit{Inference Delay Included}: model inference latency advances the simulation clock.
\textit{Success Breakdown}: separate reporting of physical completion, temporal validity, and latency-inclusive success.
$^{\dagger}$LIBERO-Industry defines three workflows (S2D, D2S, D2D) across 216 parameterized configurations.
$^{*}$DynamicVLA's DOM benchmark reports 16 real-robot tasks.
}
\vspace{-2mm}
\end{table*}

To bridge this gap, we introduce \name{}, a benchmark of 60 time-constrained manipulation tasks across five application domains and three difficulty levels (Fig.~\ref{fig:time-alignment}a).
Temporal parameters draw on manufacturer documentation, empirical studies, sports rules, and public standards where such sources exist.
Tasks fall into three primary types (Fig.~\ref{fig:time-alignment}b):
(1)~Temporal Window, requiring actions within a specified window (\eg, timely film removal from a developer bath)~\cite{sung2026liberoindustry};
(2)~Event-Reactive, requiring rapid responses to events with randomized onset times (\eg, an emergency stop after a sudden fault warning)~\cite{li2026robodyna,chen2026reflex,xiong2026react}; and
(3)~Throughput, requiring a minimum completion rate (\eg, replenishing a ball feeder to meet repeated demand)~\cite{li2026embodiedefficiency,sung2026liberoindustry}.
Each task has one primary type and may include secondary temporal constraints (Sec.~\ref{sec:temporal-taxonomy}).

We introduce a Latency-Aligned evaluation protocol
(Fig.~\ref{fig:time-alignment}c) that accounts for model
inference and control-interface delays.
Observations are timestamped using the environment clock,
and environment time advances during model inference
before actions take effect.
We report physical progress and success under different
timing conditions through four metrics: Progress Score (PS),
Physical Success Rate (PSR), Temporal Success Rate (TSR),
and Latency-Aligned TSR (LTSR)
(Fig.~\ref{fig:time-alignment}d).

Using this protocol, we systematically evaluate over
30 representative methods spanning VLAs, WAMs, and
VLM agents, revealing distinct limitations across paradigms.
VLAs benefit substantially from fine-tuning,
yet even the best-performing model achieves only
8.3\% LTSR, with limited generalization to unseen tasks.
WAMs achieve higher physical success (14.8\% PSR)
but exhibit greater latency sensitivity, with temporal
success dropping from 10.0\% to 3.8\% when inference
latency is included.
VLM agents demonstrate strong zero-shot transfer
capabilities, achieving up to 19.0\% TSR, yet inference
latency reduces this to at most 8.2\% LTSR.
These findings reveal that physical competence,
generalization, and real-time responsiveness remain
distinct challenges that must be addressed jointly
for time-aware robotic intelligence.

Our main contributions are summarized as follows:
\begin{itemize}
      \item \textbf{\name{} Benchmark.}
    We introduce a time-sensitive robotics benchmark comprising
    60 manipulation tasks across five real-world application
    domains and three difficulty levels, covering three types
    of temporal constraints: Temporal Window, Event-Reactive,
    and Throughput.

    \item \textbf{Latency-Aligned Evaluation Framework.}
    We develop a standardized evaluation protocol that accounts
    for model inference and control delays, with four metrics
    (PS, PSR, TSR, and LTSR) that summarize physical progress,
    manipulation success, and temporal compliance with and
    without inference latency.

\item \textbf{Systematic Evaluation of Robotic Agent Systems.}
We systematically evaluate over 30 representative methods
spanning VLAs, WAMs, and VLM agents, revealing distinct
limitations across paradigms.
Fine-tuned VLAs achieve up to 8.3\% LTSR but struggle
with generalization.
VLM agents demonstrate zero-shot transfer capabilities,
yet inference latency reduces their temporal success
from up to 19.0\% to at most 8.2\%.
\end{itemize}

\section{Related Work}
\label{sec:related_work}

\subsection{End-to-End Policies and Robot Agent Systems}
\label{sec:related_work_methods}

End-to-end policies map multimodal observations to actions with different latency--generalization trade-offs.
Autoregressive VLAs, such as RT-2~\cite{brohan2023rt2} and OpenVLA~\cite{kim2024openvla}, transfer web-scale semantic knowledge but incur sequential decoding costs.
Flow-matching VLAs, including $\pi_0$/$\pi_{0.5}$~\cite{black2024pi0,black2025pi05}, produce continuous action chunks for fast control, while WAMs such as DreamZero~\cite{ye2026dreamzero} and OpenWAM~\cite{wang2026openwam,yu2026openwam} jointly predict future states and actions at greater computational cost.

VLM-based robot agents couple high-level reasoning with learned policies or programmatic executors.
Code as Policies~\cite{liang2022codeaspolicies}, VoxPoser~\cite{huang2023voxposer}, and CaP-X~\cite{fu2026capx} respectively generate policies, construct value maps, and orchestrate action tools.
Such systems support compositional planning but introduce reasoning and tool-call delays.
\name{} evaluates how these trade-offs affect task-level temporal compliance.

\subsection{Robotic Benchmark Evaluation}
\label{sec:related_work_benchmarks}

Standard manipulation benchmarks have progressed from diverse skill learning in RLBench~\cite{james2020rlbench} to long-horizon language conditioning in CALVIN~\cite{mees2022calvin}, lifelong transfer in LIBERO~\cite{liu2023libero}, large-scale household tasks in RoboCasa365~\cite{nasiriany2026robocasa365}, and memory-dependent execution in RoBoMME~\cite{dai2026robomme}. 
While Embodied First Exam~\cite{wang2026embodiedfirstexam} evaluates coding agents under real-time inference delays, it relies on coarse overall episode budgets.
Overall, existing benchmarks prioritize spatial-semantic generalization and physical completion, treating time merely as a generous step limit. 
They fail to capture fine-grained temporal requirements, where success depends not only on \textit{what} actions are executed, but strictly on \textit{when} and \textit{how fast} they occur.

\subsection{Dynamic Robotic Manipulation and Evaluation}
\label{sec:related_work_temporal}

Recent efforts have begun exploring time-sensitive manipulation. 
DOMINO~\cite{fang2026domino}, DynamicVLA~\cite{xie2026dynamicvla}, and LIBERO-Dynamic~\cite{ren2026liberodynamic} study policy execution around moving objects, while Real-Time Chunking~\cite{black2025rtc} and ReflexVLA~\cite{chen2026reflex} mitigate inference latency for reactive control. 
At the task level, RoboDyna~\cite{li2026robodyna} introduces periodic physical opportunities, LIBERO-Industry~\cite{sung2026liberoindustry} models industrial transfer windows, and SafeManip~\cite{huang2026safemanip} enforces temporal logic for safety.

However, these works often couple timing with environmental dynamics or focus on narrower task setups, with limited separate reporting of physical and temporal success (Tab.~\ref{tab:temporal_comparison}).
In contrast, \textbf{RoboClock} 
enables a comprehensive evaluation centered on task-level temporal compliance across 60 tasks in five domains. 
By unifying execution windows, response deadlines, and throughput targets, while supporting timing-only difficulty variation and reporting success with and without inference latency, RoboClock enables comparisons of temporal compliance and latency sensitivity across end-to-end policies and System 1/System 2 agents.

\section{Problem Formulation}
\subsection{Time Model and Latency-Aligned Execution}
\label{sec:time_model}

\begin{figure}[t]
  \centering
  \includegraphics[width=\linewidth]{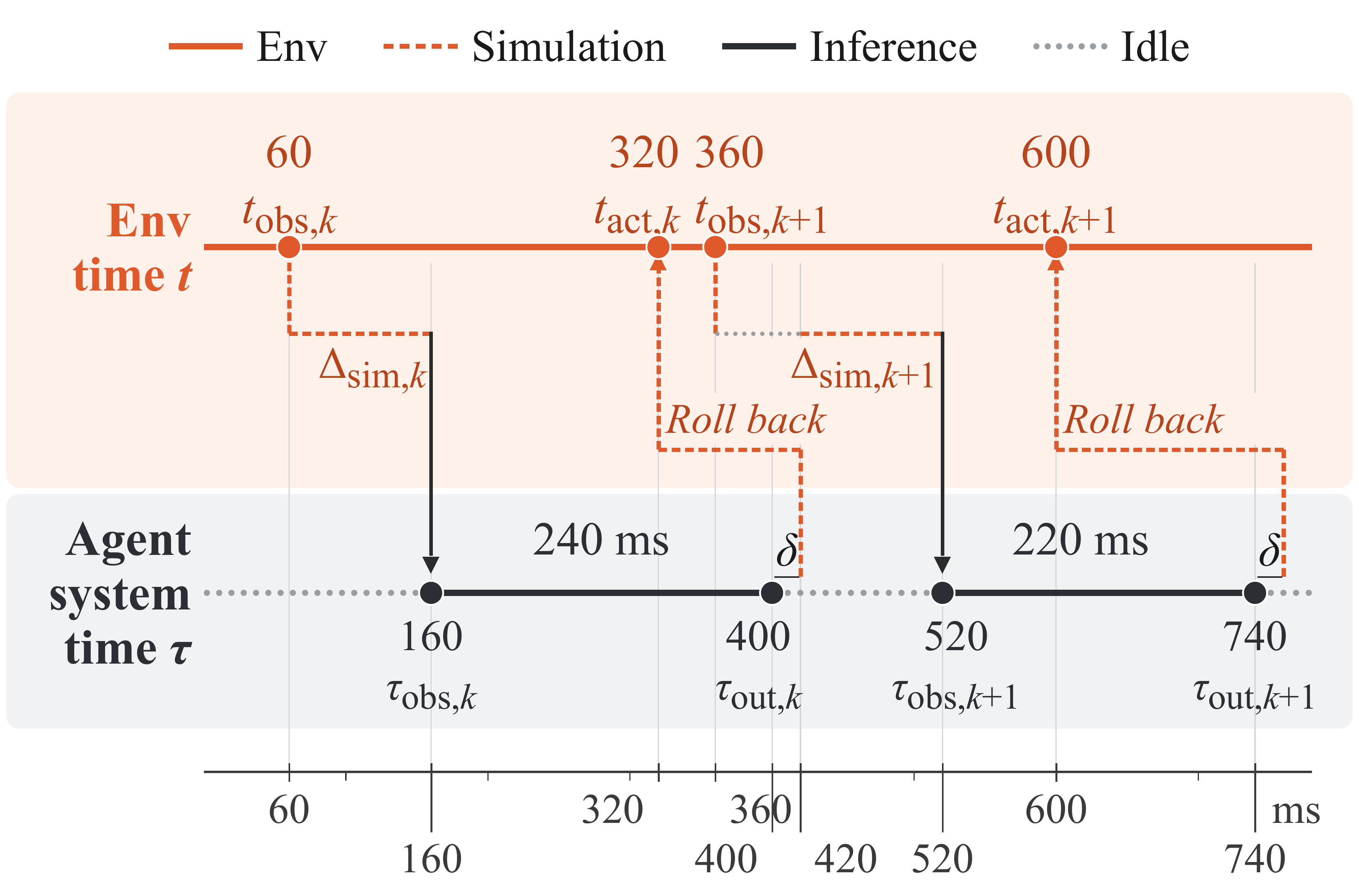}
  \caption{\textbf{Temporal alignment between environment and agent.} Environment time $t$ advances with inference, control delay, and actions; simulation overhead $\Delta_{\mathrm{sim},k}$ affects only agent time $\tau$.}
  \label{fig:block_and_replay}
\end{figure}

We separate environment time $t$ from agent wall-clock time $\tau$
to account for inference latency
without including hardware-dependent simulator overhead~\cite{mittal2025isaaclab,tao2024maniskill3}
(Fig.~\ref{fig:block_and_replay}).
We adopt the synchronous, Latency-Aligned action-chunk execution
natively supported by official VLA/WAM implementations.
Simulation, rendering, and observation-delivery overhead
$\Delta_{\mathrm{sim},k}$ affects only $\tau$.
The environment advances at fixed physics steps
$\Delta t_{\mathrm{phys}}$, with commands updated every
$\Delta t_{\mathrm{ctrl}}$.
In cycle $k$, an observation is captured at $t_{\mathrm{obs},k}$
and delivered at $\tau_{\mathrm{obs},k}$.
The policy produces an action chunk at $\tau_{\mathrm{out},k}$,
with inference latency measured as
\begin{equation}
  L_{\mathrm{infer},k}
  = \tau_{\mathrm{out},k} - \tau_{\mathrm{obs},k}.
  \label{eq:latency}
\end{equation}
This includes policy-side processing from observation reception
to action generation, excluding simulation, rendering, and observation-delivery overhead.

We use a block-and-replay protocol:
save the environment state at $t_{\mathrm{obs},k}$ and suspend
physics during inference; then restore the state and advance it
under previously committed controls for $L_{\mathrm{infer},k}+\delta$,
where $\delta$ is the fixed control-interface delay.
The new action chunk takes effect at
\begin{equation}
  t_{\mathrm{act},k}
  = t_{\mathrm{obs},k}
  + L_{\mathrm{infer},k} + \delta,
  \label{eq:t_act}
\end{equation}
rounded up to the next control timestep $\Delta t_{\mathrm{ctrl}}$.
After executing $N_{\mathrm{act},k}$ actions at frequency
$f_{\mathrm{act}}$, the next observation is sampled at
\begin{equation}
  t_{\mathrm{obs},k+1}
  = t_{\mathrm{act},k}
  + \frac{N_{\mathrm{act},k}}{f_{\mathrm{act}}},
  \label{eq:t_obs_next}
\end{equation}
and the cycle repeats.
Execution is thus synchronous: each inference uses the observation
after the preceding chunk finishes. During the replayed delay,
the scene evolves before the new chunk takes effect, exposing
the consequences of acting on a stale observation.
With deterministic replay, this preserves physical action timing
up to control-step discretization, independently of simulator speed.

\section{Benchmark Design and Construction}
\label{sec:benchmark}

\begin{figure*}[t]
\centering
\includegraphics[width=\textwidth]{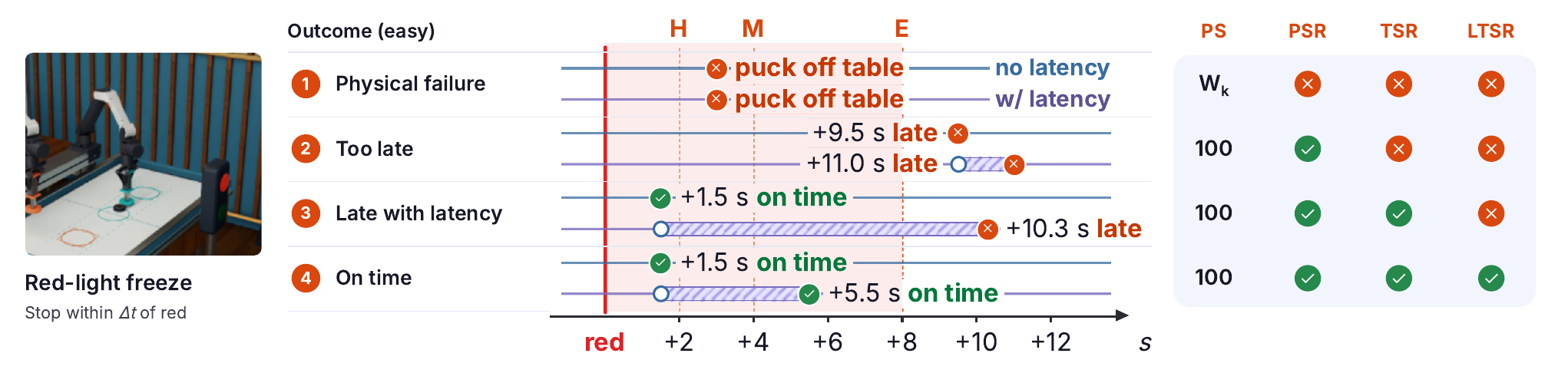}
\caption{\textbf{Illustrative outcomes under the four metrics in Red-light freeze.} Times are relative to red-light onset; H/M/E mark the hard/medium/easy response limits (2/4/8\,s), with outcomes evaluated at the easy limit ($\Delta t=8$\,s). Blue and purple timelines show execution with inference paused and with latency advancing environment time, respectively. (1) Physical failure retains partial PS but fails all success criteria. (2) A late stop achieves full PS and physical success (PSR), but fails TSR and LTSR. (3) A stop that is timely only with inference paused satisfies TSR but fails LTSR. (4) A timely stop even with latency satisfies all criteria.}
\label{fig:metric-examples}
\end{figure*}

\subsection{Three Categories of Temporal Requirements}
\label{sec:temporal-taxonomy}
\name{} organizes tasks by the temporal requirement that determines whether a physically correct action is useful. Each task belongs to one of three primary categories, set by its main temporal obligation regardless of domain. Any further timed clause (\eg, an overall time limit) is a secondary constraint; temporal success requires all.

\noindent\textbf{(1) Temporal Window:} a target event must occur within $[t_s,t_e]$; early and late actions fail (\eg, closing the faucet while the collected water is inside a target band that the flow crosses in 6/4/2\,s). Four windows open at the start or a rule-defined cue, so only $t_e$ binds (App.~\ref{app:task-catalog}).
\textbf{(2) Event-Reactive:} an unannounced trigger at randomized $t_{\mathrm{event}}$ requires $0 \le t_{\mathrm{act}}-t_{\mathrm{event}} < \Delta t$ (\eg, pressing the emergency stop within 3\,s of a motor-fault warning).
\textbf{(3) Throughput:} correct completions must satisfy $N/T \ge R_{\min}$, with rolling checks for sustained service (\eg, stocking balls for six demands spaced 12\,s apart without starvation).

\raggedbottom
\subsection{Benchmark Construction}
\label{sec:construction}

\paragraph{Domains, scenes, and assets.}
Each domain draws timing constraints from its native workflows (laboratory protocols, clinical devices, household cooking and safety, industrial safety, and games) and has dedicated Isaac Sim scenes, \eg, a kitchen built on Lightwheel Kitchen and a hospital ward converted from an artist-made model. Objects are sourced from public repositories (\eg, RoboCasa and Sketchfab) or custom-modeled; each is converted to USD and articulated as needed for manipulation. Effects beyond rigid-body physics, such as indicator color changes or tea steeping, are governed by process models that dynamically update visible geometry and materials. The 60 task scenes use 111 object and fixture models (excluding robots) plus procedural props such as displays, signal lamps, and timers (App.~\ref{app:task-catalog}).

\paragraph{Task design and randomization.}
Process quantities follow manufacturer documents, peer-reviewed studies, sports rulebooks, and public standards where a source exists. Of the 42 time-limited tasks, 8 take their easy limit from a documented time; the rest, including all 17 Throughput tasks, are calibrated against a reference controller that keeps at least 20\% or 0.3\,s of slack at hard (sources and abstractions in App.~\ref{app:timing}).
Every task passed human review by domain authors, who verified that its workflow and cue are realistic and its time limits achievable. Three difficulty levels tighten the primary temporal constraint: the window width, response limit, or required rate. Independently of difficulty, each seed uniformly samples object layouts, event onsets, and physical and process parameters from predefined ranges. Because event onsets and latent parameters are revealed solely through visual cues, policies must actively monitor the scene rather than rely on a fixed schedule.

\paragraph{Demonstrations from ground truth.}
For each task, a scripted oracle (i)~reads the current ground-truth simulator state (poses, contacts, process variables, and logged event onsets) without access to future events; (ii)~converts it via inverse kinematics into 16-D joint targets; (iii)~schedules motions to satisfy the temporal clauses; and (iv)~retains an episode only if the task evaluator confirms both physical and temporal success. Demonstrations are recorded and verify that each task is feasible on time (App.~\ref{app:demo-examples}).

\paragraph{Feasibility validation.}
The oracle is our feasibility reference: it reads privileged state, so it has no perception latency, but it moves at the rig's joint speeds and holds its start pose until the cue is shown. Run on seeds 0--9 at every released difficulty level with inference paused, it reaches joint 99.2\% physical and temporal success. Human operators also solve the tasks by teleoperation under the same temporal constraints, reaching 100\% PSR and 96\% TSR, which shows that the tasks are achievable by an agent that perceives the cues and reacts in time.

\subsection{Evaluation Metrics}
\label{sec:metrics}
We report four complementary metrics for physical progress and success under different timing conditions (Fig.~\ref{fig:metric-examples}). With inference paused and timing constraints relaxed, PS, following RoboDojo~\cite{chen2026robodojo}, measures progress across ordered physical milestones, assigning 100 only upon reaching the final goal. PSR measures the percentage of episodes that achieve this goal.

TSR measures the percentage of inference-paused episodes that achieve the physical goal while satisfying all temporal constraints (execution windows, response deadlines, throughput targets). LTSR applies the same success criteria in Latency-Aligned mode, where inference latency advances environment time. %

\section{Experimental Setup}
\label{sec:experiments}
Additional experimental details, a temporal-difficulty study (App.~\ref{app:emh}), and a real-robot pilot (App.~\ref{app:real-robot}) are provided in the appendix.
\label{sec:baselines}
\begin{table*}[t]
\centering\small
\caption{End-to-end policies on all 60 tasks: released RoboDojo checkpoints and our fine-tuned models (Sec.~\ref{sec:finetune}). A/B/C: Temporal Window, Event-Reactive, Throughput; metrics (Sec.~\ref{sec:metrics}) in \%, macro-averaged over tasks; latency on eight A100s. Shaded: fine-tuned; \textbf{bold}/\underline{underline}: best/second best.}
\label{tab:main-results}
\setlength{\tabcolsep}{5pt}
\begin{tabular}{@{}ll*{12}{c}@{}}
\toprule
& & \multicolumn{3}{c}{\cellcolor{catA}A} & \multicolumn{3}{c}{\cellcolor{catB}B} & \multicolumn{3}{c}{\cellcolor{catC}C} & \multicolumn{3}{c}{\cellcolor{catAll}Overall} \\
\cmidrule(lr){3-5}\cmidrule(lr){6-8}\cmidrule(lr){9-11}\cmidrule(l){12-14}
Class & Model & \cellcolor{catA}PSR & \cellcolor{catA}TSR & \cellcolor{catA}LTSR & \cellcolor{catB}PSR & \cellcolor{catB}TSR & \cellcolor{catB}LTSR & \cellcolor{catC}PSR & \cellcolor{catC}TSR & \cellcolor{catC}LTSR & PSR & TSR & \textbf{LTSR} \\
\midrule
\multicolumn{14}{@{}l}{\emph{Released checkpoints without fine-tuning}} \\
\midrule
\multirow{2}{*}{\textcolor{realblue}{\textbf{VLA}}} & WALL-OSS-1.0-lite~\cite{zhai2025walloss} & 0.0 & 0.0 & 0.0 & 0.0 & 0.0 & 0.0 & 0.0 & 0.0 & 0.0 & 0.0 & 0.0 & 0.0 \\
 & Galaxea G0.5~\cite{liu2026g05} & 2.1 & 0.0 & 0.0 & 0.0 & 0.0 & 0.0 & 0.0 & 0.0 & 0.0 & 0.7 & 0.0 & 0.0 \\
\midrule
 & VPP2~\cite{guo2026vpp2} & \cellcolor{rowshade}5.3 & \cellcolor{rowshade}1.6 & \cellcolor{rowshade}0.0 & \cellcolor{rowshade}0.0 & \cellcolor{rowshade}0.0 & \cellcolor{rowshade}0.0 & \cellcolor{rowshade}0.0 & \cellcolor{rowshade}0.0 & \cellcolor{rowshade}0.0 & \cellcolor{rowshade}1.7 & \cellcolor{rowshade}0.5 & \cellcolor{rowshade}0.0 \\
\multirow{-2}{*}{\textcolor{realgreen}{\textbf{WAM}}} & Awomo-0.5~\cite{awomo2026awomo05} & \cellcolor{rowshade}0.0 & \cellcolor{rowshade}0.0 & \cellcolor{rowshade}0.0 & \cellcolor{rowshade}0.0 & \cellcolor{rowshade}0.0 & \cellcolor{rowshade}0.0 & \cellcolor{rowshade}0.0 & \cellcolor{rowshade}0.0 & \cellcolor{rowshade}0.0 & \cellcolor{rowshade}0.0 & \cellcolor{rowshade}0.0 & \cellcolor{rowshade}0.0 \\
\midrule
\multicolumn{14}{@{}l}{\emph{Released checkpoints and fine-tuned counterparts}} \\
\midrule
 & $\pi_{0.5}$~\cite{black2025pi05} & 0.0 & 0.0 & 0.0 & 0.0 & 0.0 & 0.0 & 0.0 & 0.0 & 0.0 & 0.0 & 0.0 & 0.0 \\
 & \quad\textcolor{ftink}{\textit{+ fine-tuned (ours)}} & \cellcolor{ftshade}\textbf{15.8} & \cellcolor{ftshade}\textbf{10.5} & \cellcolor{ftshade}\underline{5.3} & \cellcolor{ftshade}19.6 & \cellcolor{ftshade}17.5 & \cellcolor{ftshade}\textbf{16.7} & \cellcolor{ftshade}0.0 & \cellcolor{ftshade}0.0 & \cellcolor{ftshade}0.0 & \cellcolor{ftshade}\underline{12.8} & \cellcolor{ftshade}\textbf{10.3} & \cellcolor{ftshade}\textbf{8.3} \\
 & X-VLA~\cite{zheng2026xvla} & 1.6 & 0.0 & 0.0 & 4.6 & 0.0 & 0.0 & 0.0 & 0.0 & 0.0 & 2.3 & 0.0 & 0.0 \\
\multirow{-4}{*}{\textcolor{realblue}{\textbf{VLA}}} & \quad\textcolor{ftink}{\textit{+ fine-tuned (ours)}} & \cellcolor{ftshade}8.9 & \cellcolor{ftshade}\underline{8.4} & \cellcolor{ftshade}\textbf{6.8} & \cellcolor{ftshade}18.3 & \cellcolor{ftshade}10.0 & \cellcolor{ftshade}6.7 & \cellcolor{ftshade}0.0 & \cellcolor{ftshade}0.0 & \cellcolor{ftshade}0.0 & \cellcolor{ftshade}10.2 & \cellcolor{ftshade}6.7 & \cellcolor{ftshade}\underline{4.8} \\
\midrule
 & InternW0-$\Delta$~\cite{miao2026internw0delta} & 5.3 & 0.0 & 0.0 & 1.7 & 0.0 & 0.0 & 0.0 & 0.0 & 0.0 & 2.3 & 0.0 & 0.0 \\
 & \quad\textcolor{ftink}{\textit{+ fine-tuned (ours)}} & \cellcolor{ftshade}\underline{10.5} & \cellcolor{ftshade}2.1 & \cellcolor{ftshade}1.6 & \cellcolor{ftshade}\underline{22.5} & \cellcolor{ftshade}\underline{18.3} & \cellcolor{ftshade}4.2 & \cellcolor{ftshade}0.0 & \cellcolor{ftshade}0.0 & \cellcolor{ftshade}0.0 & \cellcolor{ftshade}12.3 & \cellcolor{ftshade}8.0 & \cellcolor{ftshade}2.2 \\
 & OpenWAM-$\alpha$~\cite{wang2026openwam} & 3.7 & 0.0 & 0.0 & 0.0 & 0.0 & 0.0 & 0.0 & 0.0 & 0.0 & 1.2 & 0.0 & 0.0 \\
\multirow{-4}{*}{\textcolor{realgreen}{\textbf{WAM}}} & \quad\textcolor{ftink}{\textit{+ fine-tuned (ours)}} & \cellcolor{ftshade}7.4 & \cellcolor{ftshade}3.7 & \cellcolor{ftshade}1.6 & \cellcolor{ftshade}\textbf{29.6} & \cellcolor{ftshade}\textbf{20.8} & \cellcolor{ftshade}\underline{8.3} & \cellcolor{ftshade}\textbf{2.4} & \cellcolor{ftshade}\textbf{1.8} & \cellcolor{ftshade}0.0 & \cellcolor{ftshade}\textbf{14.8} & \cellcolor{ftshade}\underline{10.0} & \cellcolor{ftshade}3.8 \\
\bottomrule
\end{tabular}
\end{table*}

\begin{table}[t]
\centering\small
\caption{Seen (45 fine-tuning) vs.\ unseen (15 held-out) tasks, in \%. Shaded: fine-tuned; \textbf{bold}/\underline{underline}: best/second best.}
\label{tab:seen-unseen}
\setlength{\tabcolsep}{4.2pt}
\begin{tabular}{@{}l*{6}{c}@{}}
\toprule
& \multicolumn{3}{c}{Seen (45)} & \multicolumn{3}{c}{Unseen (15)} \\
\cmidrule(lr){2-4}\cmidrule(l){5-7}
Model & PSR & TSR & LTSR & PSR & TSR & LTSR \\
\midrule
$\pi_{0.5}$ & 0.0 & 0.0 & 0.0 & 0.0 & 0.0 & 0.0 \\
\quad\textcolor{ftink}{\textit{+ fine-tuned (ours)}} & \cellcolor{ftshade}\underline{17.1} & \cellcolor{ftshade}\textbf{13.8} & \cellcolor{ftshade}\textbf{11.1} & \cellcolor{ftshade}0.0 & \cellcolor{ftshade}0.0 & \cellcolor{ftshade}0.0 \\
\addlinespace[2pt]
X-VLA & 0.9 & 0.0 & 0.0 & \textbf{6.7} & 0.0 & 0.0 \\
\quad\textcolor{ftink}{\textit{+ fine-tuned (ours)}} & \cellcolor{ftshade}12.7 & \cellcolor{ftshade}8.2 & \cellcolor{ftshade}\underline{6.0} & \cellcolor{ftshade}2.7 & \cellcolor{ftshade}\textbf{2.0} & \cellcolor{ftshade}\textbf{1.3} \\
\addlinespace[2pt]
InternW0-$\Delta$ & 2.4 & 0.0 & 0.0 & 2.0 & 0.0 & 0.0 \\
\quad\textcolor{ftink}{\textit{+ fine-tuned (ours)}} & \cellcolor{ftshade}14.9 & \cellcolor{ftshade}10.7 & \cellcolor{ftshade}2.9 & \cellcolor{ftshade}\underline{4.7} & \cellcolor{ftshade}0.0 & \cellcolor{ftshade}0.0 \\
\addlinespace[2pt]
OpenWAM-$\alpha$ & 1.6 & 0.0 & 0.0 & 0.0 & 0.0 & 0.0 \\
\quad\textcolor{ftink}{\textit{+ fine-tuned (ours)}} & \cellcolor{ftshade}\textbf{19.8} & \cellcolor{ftshade}\underline{13.3} & \cellcolor{ftshade}5.1 & \cellcolor{ftshade}0.0 & \cellcolor{ftshade}0.0 & \cellcolor{ftshade}0.0 \\
\bottomrule
\end{tabular}
\end{table}

\begin{table*}[t]
\centering\small
\caption{Agent frameworks on Qwen3.8-27B with a shared budget of 100 model calls per episode. \textbf{Bold}/\underline{underline}: best/second best.}
\label{tab:interface-results}
\setlength{\tabcolsep}{3.4pt}
\begin{tabular}{@{}l*{16}{c}@{}}
\toprule
& \multicolumn{4}{c}{\cellcolor{catA}A} & \multicolumn{4}{c}{\cellcolor{catB}B} & \multicolumn{4}{c}{\cellcolor{catC}C} & \multicolumn{4}{c}{\cellcolor{catAll}Overall} \\
\cmidrule(lr){2-5}\cmidrule(lr){6-9}\cmidrule(lr){10-13}\cmidrule(l){14-17}
Agent framework / interface & \cellcolor{catA}PS & \cellcolor{catA}PSR & \cellcolor{catA}TSR & \cellcolor{catA}LTSR & \cellcolor{catB}PS & \cellcolor{catB}PSR & \cellcolor{catB}TSR & \cellcolor{catB}LTSR & \cellcolor{catC}PS & \cellcolor{catC}PSR & \cellcolor{catC}TSR & \cellcolor{catC}LTSR & PS & PSR & TSR & \textbf{LTSR} \\
\midrule
PyRUA-Lean~\cite{si2026pyrualean} & 7.4 & \underline{5.3} & 0.0 & 0.0 & 5.4 & 2.9 & 2.5 & 0.0 & 0.0 & 0.0 & 0.0 & 0.0 & 4.5 & 2.8 & 1.0 & 0.0 \\
RPent~\cite{zhang2026harnessvla} & \cellcolor{zebra}12.1 & \cellcolor{zebra}\underline{5.3} & \cellcolor{zebra}\underline{3.7} & \cellcolor{zebra}0.0 & \cellcolor{zebra}7.1 & \cellcolor{zebra}4.2 & \cellcolor{zebra}2.5 & \cellcolor{zebra}0.0 & \cellcolor{zebra}4.7 & \cellcolor{zebra}0.0 & \cellcolor{zebra}0.0 & \cellcolor{zebra}0.0 & \cellcolor{zebra}8.0 & \cellcolor{zebra}3.3 & \cellcolor{zebra}2.2 & \cellcolor{zebra}0.0 \\
HexaShell~\cite{gao2026hexashell} & 11.1 & \underline{5.3} & 2.1 & 0.0 & 5.8 & 5.4 & \underline{2.9} & 0.0 & 2.9 & 0.0 & 0.0 & 0.0 & 6.7 & 3.8 & 1.8 & 0.0 \\
PhysicalRSI~\cite{mmlab2026physicalrsi} & \cellcolor{zebra}6.3 & \cellcolor{zebra}2.1 & \cellcolor{zebra}0.0 & \cellcolor{zebra}0.0 & \cellcolor{zebra}10.4 & \cellcolor{zebra}10.0 & \cellcolor{zebra}2.5 & \cellcolor{zebra}0.0 & \cellcolor{zebra}2.4 & \cellcolor{zebra}0.0 & \cellcolor{zebra}0.0 & \cellcolor{zebra}0.0 & \cellcolor{zebra}6.8 & \cellcolor{zebra}4.7 & \cellcolor{zebra}1.0 & \cellcolor{zebra}0.0 \\
uni-robot-agent~\cite{yuan2026unirobotagent} & 6.3 & \underline{5.3} & 0.0 & 0.0 & 7.9 & 5.8 & 2.5 & 0.0 & \textbf{6.5} & \textbf{2.4} & 0.0 & 0.0 & 7.0 & 4.7 & 1.0 & 0.0 \\
SenseNova-RoboRSI~\cite{sensenova2026roborsi} & \cellcolor{zebra}\underline{16.3} & \cellcolor{zebra}\underline{5.3} & \cellcolor{zebra}\underline{3.7} & \cellcolor{zebra}0.0 & \cellcolor{zebra}\underline{14.2} & \cellcolor{zebra}\underline{11.3} & \cellcolor{zebra}\underline{2.9} & \cellcolor{zebra}0.0 & \cellcolor{zebra}\underline{5.9} & \cellcolor{zebra}\textbf{2.4} & \cellcolor{zebra}0.0 & \cellcolor{zebra}0.0 & \cellcolor{zebra}\underline{12.5} & \cellcolor{zebra}\underline{6.8} & \cellcolor{zebra}\underline{2.3} & \cellcolor{zebra}0.0 \\

Agent-as-Policy & 11.1 & 2.6 & 2.1 & 0.0 & 10.8 & 10.4 & 0.0 & 0.0 & 1.2 & 0.0 & 0.0 & 0.0 & 8.2 & 5.0 & 0.7 & 0.0 \\
EEF & \cellcolor{zebra}4.2 & \cellcolor{zebra}3.2 & \cellcolor{zebra}2.6 & \cellcolor{zebra}0.0 & \cellcolor{zebra}0.8 & \cellcolor{zebra}0.0 & \cellcolor{zebra}0.0 & \cellcolor{zebra}0.0 & \cellcolor{zebra}0.0 & \cellcolor{zebra}0.0 & \cellcolor{zebra}0.0 & \cellcolor{zebra}0.0 & \cellcolor{zebra}1.7 & \cellcolor{zebra}1.0 & \cellcolor{zebra}0.8 & \cellcolor{zebra}0.0 \\
Tool-augmented EEF & \textbf{23.2} & \textbf{10.5} & \textbf{5.3} & 0.0 & \textbf{15.8} & \textbf{14.6} & \textbf{10.4} & \textbf{2.9} & 1.8 & 0.0 & 0.0 & 0.0 & \textbf{14.2} & \textbf{9.2} & \textbf{5.8} & \textbf{1.2} \\
\bottomrule
\end{tabular}
\end{table*}

\begin{table*}[t]
\centering\small
\caption{VLMs on all 60 tasks through Tool-augmented EEF; $^\dagger$open-weight; $^\S$text-only reference (GLM-5.3 receives no images). Overall is recomputed from displayed A/B/C means with 19/24/17 task weights. LTSR charges A100 latency with speculative decoding (open-weight; Tab.~\ref{tab:spec-decoding}) or API latency. \textbf{Bold}/\underline{underline}: best/second best.}
\label{tab:agent-results}
\setlength{\tabcolsep}{2.8pt}
\begin{tabular}{@{}ll*{16}{c}@{}}
\toprule
& & \multicolumn{4}{c}{\cellcolor{catA}A} & \multicolumn{4}{c}{\cellcolor{catB}B} & \multicolumn{4}{c}{\cellcolor{catC}C} & \multicolumn{4}{c}{\cellcolor{catAll}Overall} \\
\cmidrule(lr){3-6}\cmidrule(lr){7-10}\cmidrule(lr){11-14}\cmidrule(l){15-18}
Class & Model & \cellcolor{catA}PS & \cellcolor{catA}PSR & \cellcolor{catA}TSR & \cellcolor{catA}LTSR & \cellcolor{catB}PS & \cellcolor{catB}PSR & \cellcolor{catB}TSR & \cellcolor{catB}LTSR & \cellcolor{catC}PS & \cellcolor{catC}PSR & \cellcolor{catC}TSR & \cellcolor{catC}LTSR & PS & PSR & TSR & \textbf{LTSR} \\
\midrule
 & Gemini 4 Argon~\cite{google2026gemini4argon} & \underline{41.1} & 24.7 & 20.5 & \textbf{10.5} & \underline{40.4} & \textbf{36.3} & \underline{24.6} & \textbf{12.1} & \textbf{21.8} & \textbf{7.6} & \textbf{6.5} & 0.0 & \underline{35.3} & \textbf{24.5} & \underline{18.2} & \textbf{8.2} \\
 & Claude Opus 5.5~\cite{anthropic2026opus55} & \cellcolor{zebra}30.5 & \cellcolor{zebra}17.4 & \cellcolor{zebra}12.6 & \cellcolor{zebra}0.0 & \cellcolor{zebra}30.8 & \cellcolor{zebra}26.7 & \cellcolor{zebra}23.8 & \cellcolor{zebra}2.5 & \cellcolor{zebra}12.9 & \cellcolor{zebra}\underline{1.2} & \cellcolor{zebra}\underline{0.6} & \cellcolor{zebra}0.0 & \cellcolor{zebra}25.7 & \cellcolor{zebra}16.5 & \cellcolor{zebra}13.7 & \cellcolor{zebra}1.0 \\
 & GPT-6 Astra~\cite{openai2026gpt6astra} & \textbf{43.7} & \textbf{28.4} & \underline{22.1} & \underline{5.8} & \textbf{40.8} & \underline{34.6} & \textbf{30.0} & 2.9 & \underline{18.8} & 0.0 & 0.0 & 0.0 & \textbf{35.5} & \underline{22.8} & \textbf{19.0} & \underline{3.0} \\
 & GPT-6.1 Sol~\cite{openai2026gpt61sol} & \cellcolor{zebra}38.9 & \cellcolor{zebra}\underline{27.4} & \cellcolor{zebra}\textbf{24.2} & \cellcolor{zebra}1.6 & \cellcolor{zebra}37.1 & \cellcolor{zebra}32.9 & \cellcolor{zebra}24.2 & \cellcolor{zebra}\underline{4.6} & \cellcolor{zebra}15.3 & \cellcolor{zebra}0.0 & \cellcolor{zebra}0.0 & \cellcolor{zebra}0.0 & \cellcolor{zebra}31.5 & \cellcolor{zebra}21.8 & \cellcolor{zebra}17.3 & \cellcolor{zebra}2.3 \\
\cmidrule(l){2-18}
 & Qwen3.5-4B$^\dagger$~\cite{qwen2026qwen35} & 13.7 & 10.5 & 2.6 & 0.0 & 2.9 & 2.5 & 2.1 & 0.0 & 0.6 & 0.0 & 0.0 & 0.0 & 5.7 & 4.3 & 1.7 & 0.0 \\
 & Qwen3.5-9B$^\dagger$~\cite{qwen2026qwen35} & \cellcolor{zebra}13.7 & \cellcolor{zebra}10.5 & \cellcolor{zebra}0.0 & \cellcolor{zebra}0.0 & \cellcolor{zebra}1.3 & \cellcolor{zebra}0.4 & \cellcolor{zebra}0.0 & \cellcolor{zebra}0.0 & \cellcolor{zebra}0.0 & \cellcolor{zebra}0.0 & \cellcolor{zebra}0.0 & \cellcolor{zebra}0.0 & \cellcolor{zebra}4.8 & \cellcolor{zebra}3.5 & \cellcolor{zebra}0.0 & \cellcolor{zebra}0.0 \\
 & Gemma 4 26B-A4B$^\dagger$~\cite{google2026gemma4} & 17.9 & 10.5 & 5.3 & 0.0 & 9.2 & 8.3 & 6.3 & 2.1 & 0.6 & 0.0 & 0.0 & 0.0 & 9.5 & 6.7 & 4.2 & 0.8 \\
 & Qwen3.8-27B$^\dagger$~\cite{qwen2026qwen38} & \cellcolor{zebra}23.2 & \cellcolor{zebra}10.5 & \cellcolor{zebra}5.3 & \cellcolor{zebra}0.0 & \cellcolor{zebra}15.8 & \cellcolor{zebra}14.6 & \cellcolor{zebra}10.4 & \cellcolor{zebra}2.9 & \cellcolor{zebra}1.8 & \cellcolor{zebra}0.0 & \cellcolor{zebra}0.0 & \cellcolor{zebra}0.0 & \cellcolor{zebra}14.2 & \cellcolor{zebra}9.2 & \cellcolor{zebra}5.8 & \cellcolor{zebra}1.2 \\
 & Muse Glimmer 30B$^\dagger$~\cite{meta2026museglimmer} & 11.1 & 7.9 & 2.6 & 0.0 & 6.7 & 6.3 & 0.0 & 0.0 & 1.2 & 0.0 & 0.0 & 0.0 & 6.5 & 5.0 & 0.8 & 0.0 \\
 & Qwen3.8-Flash-Next$^\dagger$~\cite{qwen2026qwen38} & \cellcolor{zebra}17.9 & \cellcolor{zebra}10.5 & \cellcolor{zebra}2.6 & \cellcolor{zebra}0.0 & \cellcolor{zebra}19.6 & \cellcolor{zebra}18.8 & \cellcolor{zebra}12.5 & \cellcolor{zebra}2.1 & \cellcolor{zebra}1.2 & \cellcolor{zebra}0.0 & \cellcolor{zebra}0.0 & \cellcolor{zebra}0.0 & \cellcolor{zebra}13.8 & \cellcolor{zebra}10.8 & \cellcolor{zebra}5.8 & \cellcolor{zebra}0.8 \\
& DeepSeek V4.1 Flash$^\dagger$~\cite{deepseek2026v41} & 11.1 & 3.2 & 2.6 & 0.0 & 16.3 & 14.6 & 4.2 & 0.0 & 1.2 & 0.0 & 0.0 & 0.0 & 10.3 & 6.8 & 2.5 & 0.0 \\
\multirow{-12}{*}{\textcolor{realorange}{\textbf{VLM}}} & GLM-5.3$^{\dagger\S}$~\cite{zai2026glm53} & \cellcolor{zebra}5.3 & \cellcolor{zebra}0.0 & \cellcolor{zebra}0.0 & \cellcolor{zebra}0.0 & \cellcolor{zebra}2.9 & \cellcolor{zebra}2.5 & \cellcolor{zebra}2.1 & \cellcolor{zebra}0.0 & \cellcolor{zebra}0.0 & \cellcolor{zebra}0.0 & \cellcolor{zebra}0.0 & \cellcolor{zebra}0.0 & \cellcolor{zebra}2.8 & \cellcolor{zebra}1.0 & \cellcolor{zebra}0.8 & \cellcolor{zebra}0.0 \\
\bottomrule
\end{tabular}
\end{table*}

The main evaluation protocol uses the easy level of all 60 tasks, with 10 seeds per task for every model. We evaluate two groups of methods and report them separately (Tabs.~\ref{tab:main-results}--\ref{tab:agent-results}): end-to-end policies (VLAs and WAMs) and foundation-model agents (VLMs and agent frameworks).

\paragraph{End-to-end policies.}
An end-to-end policy maps the current observation and instruction to actions in a single model call. We cover two classes, VLAs and WAMs, taken from the RoboDojo leaderboard~\cite{chen2026robodojo}, since \name{} shares its embodiment: WALL-OSS-1.0-lite~\cite{zhai2025walloss}, Galaxea G0.5~\cite{liu2026g05}, $\pi_{0.5}$~\cite{black2025pi05}, and X-VLA~\cite{zheng2026xvla} as VLAs, and VPP2~\cite{guo2026vpp2}, Awomo-0.5~\cite{awomo2026awomo05}, InternW0-$\Delta$~\cite{miao2026internw0delta}, and OpenWAM-$\alpha$~\cite{wang2026openwam} as WAMs.

\paragraph{Fine-tuning.}
\label{sec:finetune}
We fine-tune two VLAs ($\pi_{0.5}$ and X-VLA) and two WAMs (OpenWAM-$\alpha$ and InternW0-$\Delta$) from their released RoboDojo checkpoints on oracle demonstrations from 45 tasks, holding out 15 tasks for generalization evaluation. Every model is trained for three epochs on the same demonstration set. Tab.~\ref{tab:main-results} compares released and fine-tuned checkpoints on all 60 tasks, and Tab.~\ref{tab:seen-unseen} separates seen and held-out tasks; settings are in App.~\ref{app:finetune}.

\paragraph{VLM agents.}
We compare agent configurations using Qwen3.8-27B~\cite{qwen2026qwen38} with a shared budget of 100 model calls per episode (Tab.~\ref{tab:interface-results}). These include six open frameworks---PyRUA-Lean~\cite{si2026pyrualean}, RPent~\cite{zhang2026harnessvla}, HexaShell~\cite{gao2026hexashell}, PhysicalRSI~\cite{mmlab2026physicalrsi}, uni-robot-agent~\cite{yuan2026unirobotagent}, and SenseNova-RoboRSI~\cite{sensenova2026roborsi}---as well as Agent-as-Policy~\cite{roboprobe2026}, RoboDojo EEF control~\cite{chen2026robodojo}, and Tool-augmented EEF. We use the best-performing Tool-augmented EEF for all VLM backbones. Its tools operate on visual observations and measured robot state, without access to privileged simulator ground truth (App.~\ref{app:tool-eef}).

\paragraph{VLM backbones.}
VLMs are queried in closed loop without robot-specific training, using Tool-augmented EEF. We evaluate four proprietary models, Gemini~4 Argon~\cite{google2026gemini4argon}, Claude Opus~5.5~\cite{anthropic2026opus55}, GPT-6 Astra~\cite{openai2026gpt6astra}, and GPT-6.1 Sol~\cite{openai2026gpt61sol}, and seven open-weight models: Qwen3.5-4B/9B~\cite{qwen2026qwen35}, Gemma~4 26B-A4B~\cite{google2026gemma4}, Qwen3.8-27B and Qwen3.8-Flash-Next~\cite{qwen2026qwen38}, Muse Glimmer 30B~\cite{meta2026museglimmer}, and DeepSeek V4.1 Flash~\cite{deepseek2026v41}. As a text-only reference we also run the official GLM-5.3~\cite{zai2026glm53}, a language model without image input: it receives the same requests, but the camera images are dropped, so it acts only from the instruction, the measured robot state, and tool feedback. All run with provider-recommended settings and low thinking mode. For LTSR, open-weight models run on eight A100 GPUs with their fastest speculative-decoding configuration; proprietary models are charged their API latency.

\section{Results and Ablation Study}
\label{sec:results}

\subsection{Main Results}
All policies and agents share the same low-level 16-D joint-position action interface (Tabs.~\ref{tab:main-results}--\ref{tab:agent-results}).

\paragraph{VLAs and WAMs: training improves success, but generalization remains limited.}
The eight released checkpoints achieve at most 2.3\% overall PSR and 0.5\% TSR, indicating little zero-shot transfer to \name{} (Tab.~\ref{tab:main-results}). Three epochs of task-specific training improve all four models: OpenWAM-$\alpha$ reaches the highest PSR, 14.8\% at 10.0\% TSR, and $\pi_{0.5}$ the highest TSR, 10.3\% at 12.8\% PSR, followed by InternW0-$\Delta$ (12.3\%, 8.0\%) and X-VLA (10.2\%, 6.7\%). Most gains come from Event-Reactive tasks, where three of the four exceed 17\% TSR, while Throughput TSR stays at or below 1.8\%. The gains remain on seen tasks: seen-task TSR rises to 8.2--13.8\%, but on the 15 held-out tasks only X-VLA reaches nonzero TSR (2.0\%; Tab.~\ref{tab:seen-unseen}). Demonstrations thus improve execution but barely transfer to unseen tasks.

Inference speed decides how much of these gains survives Latency-Aligned evaluation. The VLAs keep most of their TSR: $\pi_{0.5}$ retains 8.3\% LTSR against 10.3\% TSR, and X-VLA 4.8\% against 6.7\%. The slower WAMs lose most of it: OpenWAM-$\alpha$ falls from 10.0\% to 3.8\% and InternW0-$\Delta$ from 8.0\% to 2.2\%, so $\pi_{0.5}$ also has the highest LTSR.

\paragraph{VLM agents: zero-shot transfer is constrained by inference latency.}
Without robot-specific training, Gemini~4 Argon achieves 24.5\% PSR and 18.2\% TSR through Tool-augmented EEF, while GPT-6 Astra reaches 22.8\% and 19.0\% (Tab.~\ref{tab:agent-results}). For context, GPT-6 Astra's physical completion rate is close to its 22.48\% simulation SR on RoboDojo~\cite{robodojo2026leaderboard}; the different interfaces preclude a controlled comparison of task difficulty. With inference latency included, Gemini and GPT-6 Astra fall to 8.2\% and 3.0\% LTSR, GPT-6.1 Sol and Claude Opus~5.5 to 2.3\% and 1.0\%, and no open-weight VLM exceeds 1.2\%. Gemini keeps more of its TSR because it needs fewer turns (a median of 3 against 6 for GPT-6 Astra per successful episode), which brings it close to the fine-tuned $\pi_{0.5}$ (8.3\% LTSR); these rates depend on the API serving speed (App.~\ref{app:api-speed}). \label{sec:interface-comparison}%

\begin{table}[t]
\centering\footnotesize
\caption{\textbf{Single-factor perturbations of Lab training episodes.} Before/after: $\pi_{0.5}$ TSR (\%), zero latency; $\Delta$ in percentage points. $^{**}p{<}.01$ (exact McNemar). Oracle: 488/490 perturbed episodes solved. With measured latency, later events again reduce TSR ($-10.5$ points).}
\label{tab:pi05-perturb}
\setlength{\tabcolsep}{3.5pt}
\begin{tabular}{@{}llrccc@{}}
\toprule
Perturbation & Tasks & $n$ & Before & After & $\Delta$ \\
\midrule
Lighting & B1--B7 & 70 & 45.7 & 45.7 & $+0.0$ \\
Mass, friction & B1--B7 & 70 & 45.7 & 44.3 & $-1.4$ \\
Object pose & B1, B4 & 20 & 80.0 & 55.0 & $-25.0$ \\
\midrule
Event earlier & B1--B7 & 32 & 40.6 & 46.9 & $+6.3$ \\
Event later & B1--B7 & 38 & 50.0 & 10.5 & $-39.5^{**}$ \\
Stated duration & A5 & 20 & 60.0 & 0.0 & $-60.0^{**}$ \\
\bottomrule
\end{tabular}
\end{table}

\begin{table}[t]
\centering\footnotesize
\caption{\textbf{A5: lifts remain near demonstration time (16\,s).} Times: episode seconds; \emph{none}: no lift by deadline. $^{\dagger}$Immersion 0.6\,s later, measured-latency rerun.}
\label{tab:pi05-example}
\setlength{\tabcolsep}{4pt}
\begin{tabular}{@{}lccc@{}}
\toprule
Stated duration & Required lift & Actual lift & Success \\
\midrule
12\,s (trained) & 16.0 & 16.1 & 6/10 \\
9\,s & 13.1 & \emph{none} & 0/10 \\
15\,s & 19.0 & 16.1 & 0/10 \\
12\,s, delayed entry$^{\dagger}$ & 16.6 & 16.3 & 0/10 \\
\bottomrule
\end{tabular}
\end{table}

\paragraph{Fine-tuning relies on the training clock.}
We fine-tune $\pi_{0.5}$ on 224 Lab demonstrations with elapsed time in the prompt, then perturb its training episodes (Tab.~\ref{tab:pi05-perturb}). Lighting and physics barely change TSR; object pose reduces it nonsignificantly by 25 points. Later events reduce TSR from 50.0\% to 10.5\%, with 66\% premature responses; earlier events preserve success. Changing A5's duration yields zero success: a 15\,s instruction still produces a lift at 16.1\,s, about 3\,s early (Tab.~\ref{tab:pi05-example}). Delayed immersion also causes premature lifting, suggesting reliance on demonstration timing rather than the required waiting interval.

\subsection{Ablation: VLM Decode Speed}
\label{sec:ablation-tps}
\begin{figure}[tbp]
  \centering
  \captionsetup[subfigure]{labelformat=simple}
  \renewcommand{\thesubfigure}{(\Alph{subfigure})}
  \begin{subfigure}[t]{0.49\linewidth}
    \centering
    \includegraphics[width=\linewidth]{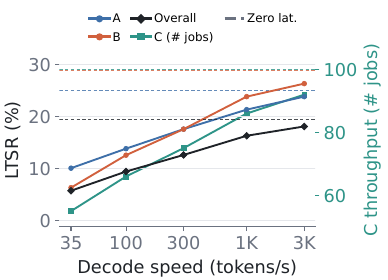}
    \caption{GPT-6 Astra}
    \label{fig:ablation-tps-gpt6}
  \end{subfigure}\hfill
  \begin{subfigure}[t]{0.49\linewidth}
    \centering
    \includegraphics[width=\linewidth]{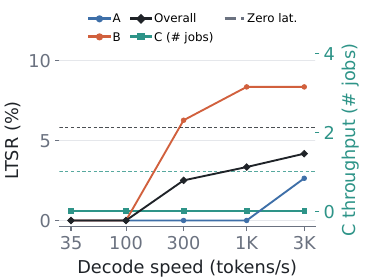}
    \caption{Qwen3.8-Flash-Next}
    \label{fig:ablation-tps-qwen}
  \end{subfigure}
  \caption{\textbf{Effect of decode speed on LTSR.} Left axis: LTSR; right axis: completed Throughput jobs; dashed: zero-latency reference.}
  \label{fig:ablation-tps}
\end{figure}
To isolate decoding latency, we assume speeds of 35--3K tokens/s and compute each query's delay as its output token count divided by the assumed speed. We add this delay to environment time in closed-loop reruns of GPT-6 Astra and Qwen3.8-Flash-Next (Fig.~\ref{fig:ablation-tps}). For GPT-6 Astra, success rises monotonically with speed: overall LTSR grows from 5.7\% to 18.0\% (Event-Reactive 6.2\% to 26.3\%, Temporal Window 10.0\% to 23.8\%) and completed Throughput jobs from 55 to 92. Event-Reactive tasks benefit most, as each call consumes part of their response deadline, and gains diminish beyond 1K tokens/s as LTSR approaches the inference-paused reference. Qwen3.8-Flash-Next (Fig.~\ref{fig:ablation-tps-qwen}) stays near zero at every speed; its zero-latency reference is already low, so its loss is capability-limited. These results show how our benchmark can guide targeted inference acceleration for VLM agents by identifying where faster decoding improves task success.

\subsection{Analysis: Latency Cost Across Policy Families}
\label{sec:ablation-latency}
\begin{figure}[tbp]
  \centering
  \includegraphics[width=\linewidth]{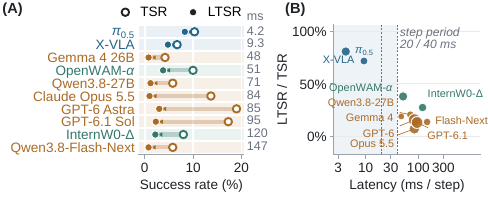}
  \caption{\textbf{Latency cost across policy families.} (A)~TSR and LTSR by inference time per action step. (B)~LTSR/TSR ratio; area $\propto$ TSR. Blue: VLA, green: WAM (both fine-tuned), orange: VLM.}
  \label{fig:ablation-latency}
\end{figure}
Per action step, a 25\,Hz policy step for VLAs and WAMs and a 50\,Hz control step for VLM tool calls (Fig.~\ref{fig:ablation-latency}), only the two VLAs stay within their 40\,ms step period, at 4.2 and 9.3\,ms, and keep 81\% and 72\% of their TSR. The other plotted models need 48--147\,ms and keep at most 38\% (OpenWAM-$\alpha$ at 51\,ms); the proprietary VLMs keep at most 16\% despite the highest TSRs.

\subsection{Failure Case Study}
\label{sec:failure-analysis}
\begin{figure}[t]
  \centering
  \includegraphics[width=\linewidth]{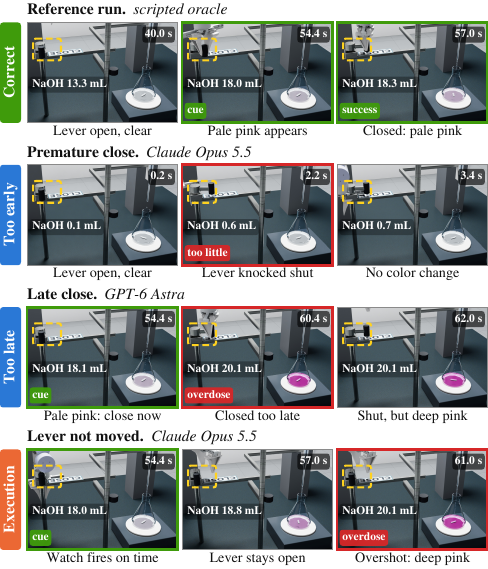}
  \caption{\textbf{Titration task.} Close the stopcock (dashed box) when the solution turns pale pink, after 18--20\,mL of NaOH.}
  \label{fig:failure-casestudy}
\end{figure}
Fig.~\ref{fig:failure-casestudy} shows one titration task failed in three ways: Claude Opus~5.5 knocks the lever shut after 0.7 of the required 18\,mL; GPT-6 Astra closes it at 60.4\,s, after the flask has turned strongly pink; and in another Opus episode, the trigger fires on time but the grasp misses the lever. Our evaluation distinguishes these outcomes by relating physical completion to the cue and valid time window: the first two close the lever outside the window, whereas the third fails to execute a timely response. These traces suggest different interventions---better cue-conditioned action selection, shorter response delays, and more reliable manipulation---showing how \name{} can guide improvements beyond a single success score.

\flushbottom
\section{Discussion and Limitations}
\label{sec:discussion}
Fine-tuned VLAs offer fast execution but limited transfer, while VLM agents generalize better but lose most temporal success to inference delay. These complementary strengths motivate coupling a VLM planner with a learned VLA/WAM controller. Through latency sweeps and failure analysis, \name{} can guide targeted improvements in inference speed, timing decisions, and manipulation reliability. Our evaluation uses simulation with simplified processes and some compressed time scales, so transfer to physical robots remains to be validated. LTSR also depends on inference hardware and serving conditions, which should be considered when comparing systems.

\section{Conclusion}
We introduced \name{}, a benchmark of 60 manipulation tasks with Temporal Window, Event-Reactive, and Throughput requirements, evaluated under a Latency-Aligned protocol that reports physical completion and temporal compliance with and without inference latency. Fine-tuned policies improve on familiar tasks but rarely transfer, while VLM agents transfer but struggle with latency, motivating generalizable agents with temporal reasoning and fast execution.

\clearpage
\section*{LLM Usage Statement}
We used OpenAI Codex and Anthropic Claude Code to assist with code implementation and debugging, and OpenAI ChatGPT and Google Gemini for language editing to improve the clarity and readability of the manuscript. All experimental results were obtained from actual experimental runs and were not generated or fabricated by LLMs. The authors reviewed and verified the implementations, experimental results, and scientific conclusions, and take full responsibility for the content of this work.
{\small
\bibliographystyle{ieeenat_fullname}
\bibliography{roboclock}}
\clearpage
\appendix
\raggedbottom
\makeatletter\setlength{\@fptop}{0pt}\setlength{\@dblfptop}{0pt}\setlength{\@fpsep}{14pt plus 4pt}\setlength{\@dblfpsep}{14pt plus 4pt}\@secpenalty=0\makeatother
\setcounter{dbltopnumber}{2}\renewcommand{\dbltopfraction}{0.85}\setcounter{topnumber}{3}\renewcommand{\topfraction}{0.85}\setcounter{totalnumber}{5}\renewcommand{\textfraction}{0.1}
\begin{center}
  {\Large\bfseries Appendix}
\end{center}
\noindent Apps.~\ref{app:setup}--\ref{app:task-catalog} give the evaluation setup, the Tool-augmented EEF interface, and the task catalog; App.~\ref{app:timing} the time limits; Apps.~\ref{app:emh}--\ref{app:agent-behavior} further VLM analyses; Apps.~\ref{app:manifest}--\ref{app:reproducibility} the task manifest, monitor checks, study map, logs, and release; and App.~\ref{app:real-robot} a real-robot pilot.

\section{Simulation and Evaluation Details}
\label{app:setup}
Tab.~\ref{tab:setup} compares our settings with RoboDojo~\cite{chen2026robodojo}; Tab.~\ref{tab:tasks} gives task counts per domain.
\begin{table*}[t]
\centering
\caption{Simulation and sensor settings. Our rig, cameras, and rates follow RoboDojo~\cite{chen2026robodojo}; rows marked $^\dagger$ differ.}
\label{tab:setup}
\footnotesize
\begin{tabular}{@{}p{2.6cm}p{6.6cm}p{4.6cm}@{}}
\toprule
Setting & Ours & RoboDojo \\
\midrule
Simulator$^\dagger$            & Isaac Sim 6.0.1, PhysX (CPU)          & Isaac Sim 5.1.0, Isaac Lab \\
Physics rate                   & 250\,Hz ($\Delta t_{\mathrm{phys}}=4$\,ms) & 250\,Hz ($\Delta t=4$\,ms) \\
Control rate$^\dagger$         & 50\,Hz ($\Delta t_{\mathrm{ctrl}}=20$\,ms, 5 substeps) & -- \\
Policy / data rate             & 25\,Hz nominal; Latency-Aligned at evaluation & 25\,Hz \\
Rendering                      & once per policy observation           & every 10 physics steps \\
Renderer                       & RTX RaytracedLighting, DLAA, \texttt{quality} & same \\
Robot                          & dual ARX X5, 6-DoF + parallel gripper & same \\
Base separation                & 0.60\,m, same height and yaw          & same \\
Rig placement$^\dagger$        & per task (origin, yaw)                & one fixed world pose \\
Head camera pose               & $(0.04, 0, 0.543)$\,m from base midpoint, $60^\circ$ down & same \\
Head camera                    & Gemini 345Lg, $640\times480$, $96.0^\circ\times79.6^\circ$ & Gemini 345Lg, $640\times480$ \\
Wrist cameras                  & D435 on \texttt{link6}, $640\times480$, $77.7^\circ\times62.3^\circ$ & same \\
Observation$^\dagger$          & 3$\times$RGB (+depth), joints, EE poses, gripper, env.\ time, instruction & 3$\times$RGB-D, joints, EE poses, gripper \\
Action                         & 16-D absolute joint-position targets  & joint positions \\
Inference timing$^\dagger$     & block-and-replay, $L_{\mathrm{infer}}+\delta$ charged to env.\ time ($\delta=20$\,ms) & -- \\
Episode horizon$^\dagger$      & task time budget                      & $1.2\times$ P90 demo length \\
Episodes per task$^\dagger$    & easy; 10 seeds per model            & 50 \\
Metrics$^\dagger$              & PS / PSR / TSR / LTSR                       & success rate, score \\
\bottomrule
\end{tabular}
\end{table*}

\begin{table}[t]
\centering
\caption{Task suite by domain and primary temporal category. W: Temporal Window, R: Event-Reactive, T: Throughput.}
\label{tab:tasks}
\small
\setlength{\tabcolsep}{4pt}
\begin{tabular}{@{}lccccl@{}}
\toprule
Domain & W & R & T & Tasks & Levels \\
\midrule
Laboratory      & 6  & 7  & 3  & 16 & E / M / H \\
Household       & 4  & 3  & 3  & 10 & E / M / H \\
Healthcare      & 4  & 6  & 4  & 14 & E / M / H \\
Sports \& games & 2  & 5  & 3  & 10 & E / M / H \\
Industry        & 3  & 3  & 4  & 10 & E / M / H \\
\midrule
Total           & 19 & 24 & 17 & 60 & \\
\bottomrule
\end{tabular}
\end{table}

\paragraph{Simulation and rig.}
PhysX runs on the CPU at 250\,Hz, five substeps per 20\,ms control step, as in RoboDojo. Images are rendered once per policy observation with the RTX real-time renderer, which does not change the dynamics. The two arms stand side by side with bases 0.60\,m apart; the head camera is 0.543\,m above and 0.04\,m in front of their midpoint, pitched $60^\circ$ down, and each wrist camera sits on the flange. Unlike RoboDojo, the rig is placed per task within each scene, and its geometry is checked at every reset. Every cue lies inside the head-camera image. Simulation and rendering run on RTX GPUs, with one episode per simulator process. The eight NVIDIA A100 80\,GB GPUs serve model inference and policy fine-tuning.

\paragraph{Observations, actions, and episodes.}
Proprioception comprises 16 joint positions, both EEF poses, and gripper widths. All methods output the same 16-D joint-position action through model-specific adapters. The main protocol uses the easy level with 10 seeds per task for every model, and the horizon is the task's time budget.

\paragraph{Foundation-model settings.}
VLMs and agent-framework backbones use provider-recommended settings with low thinking mode; all agent frameworks use Qwen3.8-27B. LTSR charges API-served models their measured API latency and locally served models their latency on the A100 node. DeepSeek V4.1 Flash and the text-only GLM-5.3 reference are accessed through APIs. Table~\ref{tab:spec-decoding} lists serving configurations for six open-weight models; their mean billed latencies range from 2.3 to 5.9\,s per query.
\begin{table*}[t]
\centering
\caption{Inference configurations for LTSR on eight A100s. Each listed open-weight model uses its fastest speculative decoding: a built-in MTP head or a DFlash drafter. $K$: draft tokens per step (default of each model's or drafter's released serving configuration); speedup vs.\ no drafter; billed latency: mean per-query time charged to environment time. Qwen3.8-27B backs every agent framework; API-served models, including DeepSeek V4.1 Flash and GLM-5.3, are charged their measured API latency.}
\label{tab:spec-decoding}
\footnotesize
\begin{tabular}{@{}llccrccc@{}}
\toprule
Model & Decoding & $K$ & Accept.\ (\%) & tok/s & Speedup & TTFT (s) & Billed latency (s) \\
\midrule
Qwen3.5-4B          & built-in MTP          & 3  & 70.0 & 562.1 & 2.20$\times$ & 0.22 & 2.27 \\
Qwen3.5-9B          & built-in MTP          & 3  & 72.1 & 500.4 & 1.82$\times$ & 0.23 & 2.49 \\
Gemma 4 26B-A4B     & z-lab DFlash          & 15 & 21.8 & 289.5 & 1.50$\times$ & 0.44 & 4.28 \\
Qwen3.8-27B         & DFlash2 drafter       & 7  & 42.6 & 274.1 & 2.44$\times$ & 0.32 & 4.59 \\
Muse Glimmer 30B    & Meta DFlash assistant & 15 & 20.7 & 259.8 & 2.86$\times$ & 0.24 & 4.72 \\
Qwen3.8-Flash-Next  & built-in MTP          & 3  & 54.9 & 212.0 & 1.82$\times$ & 0.48 & 5.94 \\
\bottomrule
\end{tabular}
\end{table*}

\paragraph{Task validation.}
\label{app:validation}
The scripted oracle meets all clauses in 99.2\% of episodes at the released limits. Human teleoperation achieves 96\% TSR under the same temporal constraints.

\subsection{Fine-tuning VLAs and WAMs}
\label{app:finetune}
We hold out 15 tasks, one per category in each domain. The other 45 provide 610 oracle demonstrations (12--14 per task, easy and medium; 610{,}098 frames at 25\,Hz) with head and wrist images, proprioception, environment time, and joint targets; evaluation seeds are disjoint from demonstration seeds. Elapsed environment time is appended to the instruction at 0.1\,s resolution. All four policies start from their released RoboDojo checkpoints and are fine-tuned for three epochs on eight A100 GPUs with the defaults of their released training interfaces (Tab.~\ref{tab:finetune}); only the epoch count, the text limits, and normalization ranges that the demonstrations exceed are changed.
\begin{table*}[t]
\centering\footnotesize
\caption{Fine-tuning configuration (three epochs for every model). Except for the epoch count, text limits, and normalization ranges, all values are the defaults of each model's released training interface. Batch: per-GPU batch $\times$ gradient accumulation $\times$ GPUs. Text limits in tokens; $\pi_{0.5}$ includes discretized state tokens.}
\label{tab:finetune}
\setlength{\tabcolsep}{4pt}
\renewcommand{\arraystretch}{1.12}
\begin{tabularx}{\textwidth}{@{}p{2.0cm}YYYY@{}}
\toprule
Setting & $\pi_{0.5}$ & X-VLA & OpenWAM-$\alpha$ & InternW0-$\Delta$ \\
\midrule
Initialization & RoboDojo sim & RoboDojo sim & RoboDojo sim & RoboDojo sim \\
Epochs & 3 & 3 & 3 & 3 \\
Action space & Joint delta, 14-D & EEF, 20-D (xyz, rot6d, gripper) & EEF, 20-D & Joint, 14-D \\
Action chunk & 50 (2.0\,s) & 30 (1.2\,s) & 32 (1.28\,s) & 32 (1.28\,s) \\
Camera input & Head + two wrists; $224^2$ each & Head only; $224^2$ & Three views tiled; $384\times320$ & Three views tiled; $384\times256$ \\
Text limit & $200\to288$ & $50\to224$ & 512 & $128\to256$ \\
Batch & $32\times1\times8=256$ & $32\times1\times8=256$ & $4\times1\times8=32$ & $8\times32\times8=2048$ \\
Learning rate & $2.5\times10^{-5}$; 500-step warm-up; cosine to $2.5\times10^{-6}$ & $10^{-4}$, constant; VLM multiplier 0.1; first 1K steps update only soft prompts and action heads & $10^{-4}$; 5\% warm-up; cosine to $10^{-6}$ & $5\times10^{-5}$; 5\% warm-up; cosine to $5\times10^{-7}$ \\
Optimizer & AdamW $(0.9,0.95)$; EMA 0.99; FSDP & AdamW $(0.9,0.95)$ & AdamW $(0.9,0.95)$; weight decay 0.01; bf16; ZeRO-2 & AdamW $(0.9,0.95)$; weight decay 0.01; bf16; ZeRO-1 \\
Normalization & Quantile; state range expanded & None; fixed loss weights & Min--max; EEF position range expanded & Z-score; joint-1 std expanded \\
\bottomrule
\end{tabularx}
\end{table*}

\section{Tool-augmented EEF Interface}
\label{app:tool-eef}
Every VLM in Tab.~\ref{tab:agent-results} uses the same eleven tools, settings, and prompt on all 60 tasks.

\paragraph{Observations and tools.}
For vision-capable agents, each turn carries the head and wrist images, the instruction, the task clock, and each arm's measured tool point and gripper opening. The text-only GLM-5.3 reference receives no images. Head depth only lifts a chosen pixel to a world point; the agent never reads object poses, task state, or event times, and tool results are measurements only. The model answers with one to four tool calls (Tab.~\ref{tab:tool-eef}). A target is a head-image pixel or a world point, and a \texttt{when} trigger holds a motion until the image patch at a pixel changes. Invalid calls are refused with the valid range.

\begin{table*}[t]
\centering\footnotesize
\caption{The eleven tools in Tool-augmented EEF. Numeric arguments are model-chosen within fixed ranges; targets are pixels or world points.}
\label{tab:tool-eef}
\begin{tabular}{@{}l p{0.56\linewidth} p{0.3\linewidth}@{}}
\toprule
Tool & Behavior & Arguments \\
\midrule
\texttt{move} & Straight-line move of the tool point to a target raised by \texttt{height}. & arm, to, height, approach, yaw, tilt, gripper, opening, speed, when \\
\texttt{move\_by} & Straight-line move by an offset from the measured tool point; with \texttt{guard} it stops on contact and keeps pressing. & arm, by, guard, gripper, opening, speed, when \\
\texttt{home} & Return to the arm's initial pose. & arm, gripper, opening, speed, when \\
\texttt{grasp} & Open, approach from the stand-off, go \texttt{depth} past the target under guard, close with \texttt{squeeze}, lift. & arm, object, approach, yaw, tilt, depth, opening, squeeze, lift, when \\
\texttt{place} & Rise to the carry clearance, move over the release point, lower (guarded for \texttt{on}/\texttt{into}), open, back off. & arm, target, mode, height, yaw, opening, when \\
\texttt{press} & Closed fingers from the press stand-off, \texttt{depth} past the point, hold, back off. & arm, target, approach, yaw, tilt, depth, hold, when \\
\texttt{push} & Closed fingers come down behind the point and slide by an offset. & arm, by, target, approach, yaw, tilt, when \\
\texttt{pull} & Close on the point as in \texttt{grasp}, drag by an offset, release. & arm, by, target, approach, yaw, tilt, depth, opening, squeeze, when \\
\texttt{turn} & Close on the point and rotate the hand by \texttt{angle}; with \texttt{hinge}, swing on an arc about a world axis (lever, crank, door). & arm, angle, target, hinge, approach, yaw, tilt, depth, opening, squeeze, when \\
\texttt{wait} & Hold both arms for a duration, until a clock time, or until a watched pixel changes. & seconds, until\_t, watch, timeout \\
\texttt{configure} & Change episode settings (speeds, stand-offs, contact and arrival thresholds, watch parameters); returns all of them. & the 19 settings \\
\bottomrule
\end{tabular}
\end{table*}

\paragraph{Execution.}
Tools execute by inverse kinematics as 16-D joint targets, in closed loop on proprioception and at the rig's joint speeds. Guarded motions stop on contact, and unreachable or colliding motions fail without moving. The interface thus adds no privileged information and no per-task tuning, and every query is charged its latency (Sec.~\ref{sec:time_model}).

\paragraph{Other interfaces in Tab.~\ref{tab:interface-results}.}
Agent-as-Policy~\cite{roboprobe2026} emits low-level arm commands at every query; RoboDojo EEF~\cite{chen2026robodojo} sets absolute tool poses, planned with CuRobo.

\section{Task Catalog and Demonstrations}
\label{app:task-catalog}
\begin{table*}[t]
\centering
\small
\setlength{\tabcolsep}{4pt}
\renewcommand{\arraystretch}{1.08}
\caption{Task catalog: Laboratory and Healthcare. IDs are domain-local. A: Temporal Window; B: Event-Reactive; C: Throughput.}
\label{tab:task-catalog}
\begin{tabularx}{\textwidth}{@{}p{1.05cm}>{\raggedright\arraybackslash}p{3.3cm}>{\centering\arraybackslash}p{0.6cm}Y@{}}
\toprule
ID & Task & Cat. & Physical goal and temporal obligation \\
\midrule
\rowcolor{black!5}
\multicolumn{4}{@{}l@{}}{\textbf{Laboratory} (16 tasks: 6 A, 7 B, 3 C)} \\
\addlinespace[2pt]
A1 & Water collection & A & Close the faucet while the collected volume is within the target band. \\
A2 & Phase separation & A & Close the stopcock at the phase boundary before the upper phase enters the receiver. \\
A3 & Chromatography fractions & A & Exchange receiving tubes to collect each colored fraction during its elution interval. \\
A4 & Titration endpoint & A & Reduce or stop titrant flow at the indicator endpoint, before excess titrant is added. \\
A5 & Film development & A & Retain the reel in the stop bath for the specified interval, then transfer it to the fixer. \\
A6 & Warm-to-ice transfer & A & Pick the tube up from the bench stand, keep it in the warm bath for the specified duration, then transfer it to ice. \\
B1 & Hotplate isolation & B & After the alarm, lift the beaker clear of the hotplate and place it upright in the safety tray. \\
B2 & Electrical isolation & B & Open the knife switch after the load indicator turns red, before the response deadline. \\
B3 & Motor-stall response & B & Detect the stopped wheel and disconnect its supply before the response deadline. \\
B4 & Test-tube rescue & B & Grasp and stabilize the tilting tube after the trigger, before it falls or loses sample. \\
B5 & Freezer-door closure & B & Close the released freezer door after its alarm and before the response deadline. \\
B6 & Vacuum relief & B & Press the vent plunger after the pressure warning, before the bumping deadline. \\
B7 & Waste-inlet closure & B & Close the inlet pinch clamp after the level alarm, before the carboy overflows. \\
C1 & Rack organization & C & Sort upright tubes into color-matched racks within the horizon and accuracy requirement. \\
C2 & Rack loading & C & Load upright samples into color-matched slots while meeting the rate and accuracy targets. \\
C3 & Calibration placement & C & Align samples with their color markers and axes while meeting the rate and accuracy targets. \\
\addlinespace[4pt]
\rowcolor{black!5}
\multicolumn{4}{@{}l@{}}{\textbf{Healthcare} (14 tasks: 4 A, 6 B, 4 C)} \\
\addlinespace[2pt]
A1 & Drip-chamber level & A & Open the roller clamp so that the bag drips into the chamber, then close it while the chamber level is inside the target band. \\
A2 & Sequential bag switch & A & Switch the selector after the first bag runs empty but before its drip chamber runs dry. \\
A3 & Bag-valve ventilation & A & Squeeze the resuscitator bag once per 6\,s breath cycle, each breath inside the window around its beat, with no extra breaths. \\
A4 & Infusion startup & A & After the request, seat the bag on the trainer spike and hook, then press the pump START key before the deadline. \\
B1 & Oxygen desaturation & B & After the low-SpO$_2$ alarm, turn the wall flowmeter until its float settles in the ordered flow band. \\
B2 & AED response & B & After the alarm, place both trainer pads and trigger the simulated shock before its deadline. \\
B3 & Hazardous-drug spill & B & After the drug bottle breaks, cover the spreading spill with the absorbent towel, then put both glass pieces in the sharps bin. \\
B4 & Ventilator reconnection & B & After the circuit comes off the tracheostomy tube, reseat the elbow on the connector and release it. \\
B5 & Anaphylaxis auto-injector & B & After the low-blood-pressure alarm, press the adrenaline auto-injector's tip on the thigh pad until it fires and hold it for 3\,s, before the response deadline. \\
B6 & Haemorrhage tourniquet & B & After the thigh wound starts to bleed, turn the tourniquet's windlass rod half a turn into its retaining clip and release it before the response deadline. \\
C1 & Pharmacy order bins & C & Fill prescription bins with the requested vials while meeting the rate and accuracy targets. \\
C2 & Instrument-tray sorting & C & Place instruments flat in their matching tray slots within the shift and accuracy limits. \\
C3 & Crash-cart restocking & C & Restock upright vials in the matching drawer lanes within the shift and accuracy limits. \\
C4 & Blood-tube triage & C & Route upright tubes to the analyzer racks matching their stopper colors within the shift. \\
\bottomrule
\end{tabularx}
\end{table*}

\begin{figure*}[t]
\centering
\includegraphics[width=\textwidth]{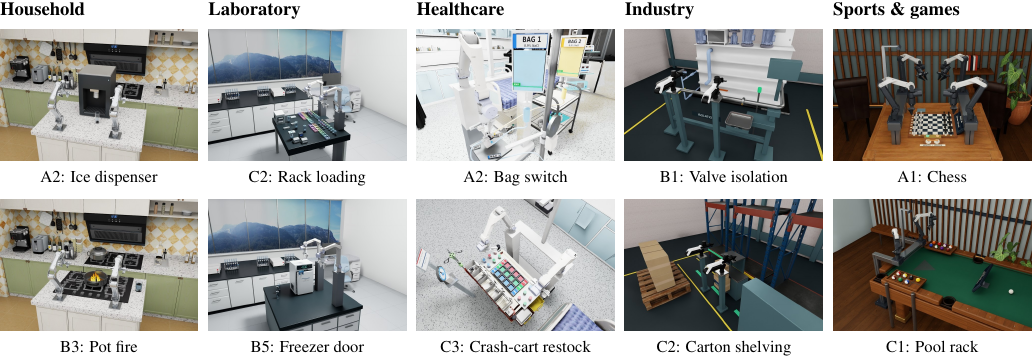}
\caption{\textbf{Scenes and assets across the five domains,} two task scenes per column. Excluding robots, room shells, variants, and procedural props, the domains use 27 (Household), 14 (Laboratory), 26 (Healthcare), 23 (Industry), and 21 (Sports \& games) object and fixture models, from Lightwheel Kitchen, RoboCasa, YCB-Video, HANDAL, IndustReal, Sketchfab, Poly Haven, HOT3D, CAD libraries, and NVIDIA Isaac.}
\label{fig:scenes-assets}
\end{figure*}

Tab.~\ref{tab:task-catalog} lists all 60 tasks; A, B, and C denote Temporal Window, Event-Reactive, and Throughput (W, R, T in Tab.~\ref{tab:tasks}), and levels and seeds instantiate them without adding tasks. Fig.~\ref{fig:scenes-assets} shows scenes and asset sources, and Figs.~\ref{fig:task-demo-a}--\ref{fig:task-demo-c} show oracle demonstrations at the medium level (seed 50); frame labels are environment times.

\paragraph{Windows that open at a known start.}
In four Temporal Window tasks the interval opens at the start or at a rule-defined cue, so in practice only its end binds. Healthcare A4 (from the instruction) and Household A4 (from the episode start) set a stated time limit on a multi-step procedure. Sports \& games A1 and A2 follow game clocks: the move and clock press are due after the opponent's press, and the shot must be released between READY and the shot-clock expiry, so a release before READY fails. We class them as Temporal Window because the interval is fixed in advance by the instruction or the rules and the robot paces a procedure inside it, instead of responding to a fault or hazard. Their limits also satisfy the Event-Reactive inequality, so they can equally be read as deadline tasks; all category results count them under A.

\paragraph{Secondary constraints.}
Besides its primary requirement, a task may carry further timed clauses, listed per task in Tabs.~\ref{tab:timing-lab}--\ref{tab:timing-sports}. For example, Laboratory A5 must also release the reel in the fixer within 5\,s of the lift, and Healthcare B3 must cover the spill within 0.4 of its response limit. Temporal success requires every clause.

\subsection{Example Demonstrations}
\label{app:demo-examples}
We describe one oracle run per category (Sec.~\ref{sec:construction}), in environment time.

\paragraph{Temporal Window: Film development (Laboratory A5, Fig.~\ref{fig:task-demo-a}b).}
The reel must stay in the stop bath for 12\,s from its measured immersion, be lifted at most 4, 2, or 1\,s later (easy, medium, hard), and rest released in the fixer within 5\,s of the lift. The oracle times the lift from the measured immersion; its rise lands 0.05\,s after the countdown ends.

\paragraph{Event-Reactive: Motor-fault emergency stop (Industry B2, Fig.~\ref{fig:task-demo-b}a).}
At a seeded onset (14--18\,s) the fault panel switches from RUN to MOTOR FAULT; the robot must actuate the emergency stop within 3\,s at the easy level, and a press before the onset fails. The oracle holds its start pose until the onset and actuates the stop about 2.0\,s after it.

\paragraph{Throughput: Ball service (Sports \& games C3, Fig.~\ref{fig:task-demo-c}b).}
Six ball demands follow a schedule (the first at 21.65, 13.20, or 8.05\,s, then every 12.00, 7.32, or 4.46\,s); a demand that finds no stocked ball is served late by the next ball stocked, and more than four balls in the hopper fail as overflow. The oracle runs one grasp-and-place cycle per arm to keep the station stocked ahead of each demand.
\renewcommand{\dblfloatpagefraction}{0.5}
\begin{table*}[p]
\centering
\small
\setlength{\tabcolsep}{4pt}
\renewcommand{\arraystretch}{1.08}
\ContinuedFloat
\caption[]{Task catalog (continued): Household, Sports \& games, and Industry. Parenthesized Industry numbers are recording aliases.}
\begin{tabularx}{\textwidth}{@{}p{1.05cm}>{\raggedright\arraybackslash}p{3.3cm}>{\centering\arraybackslash}p{0.6cm}Y@{}}
\toprule
ID & Task & Cat. & Physical goal and temporal obligation \\
\midrule
\rowcolor{black!5}
\multicolumn{4}{@{}l@{}}{\textbf{Household} (10 tasks: 4 A, 3 B, 3 C)} \\
\addlinespace[2pt]
A1 & Sugar pouring and stirring & A & Tilt the open jar to pour sugar into the tea, set it down while the sugar heap is within the gram band, then stir and return the spoon. \\
A2 & Ice dispensing & A & Hold the glass under the dispenser and release the paddle when the ice reaches the target band. \\
A3 & Pancake flip & A & Flip the pancake when its underside is golden, then leave it cooking in the pan. \\
A4 & Kettle pouring & A & Pour the target mass of water into the jug and set the kettle down before the deadline. \\
B1 & Faucet shutoff & B & Stop the running faucet before the filling bowl overflows. \\
B2 & Rolling-object catch & B & Catch and retain the rolling object before it falls from the counter. \\
B3 & Burner shutoff & B & Turn off the burner after the pot catches fire. \\
C1 & Meal service & C & Serve the ordered meals in order, relaying bowls via the staging mat to the single serving position before each waiter pickup. \\
C2 & Cola service & C & Fill the glasses to their marked levels without spilling, then return both bottles upright. \\
C3 & Cutlery organization & C & Sort the mixed utensils into the matching drawer compartments as quickly as possible. \\
\addlinespace[4pt]
\rowcolor{black!5}
\multicolumn{4}{@{}l@{}}{\textbf{Sports \& games} (10 tasks: 2 A, 5 B, 3 C)} \\
\addlinespace[2pt]
A1 & Chess exchange and clock & A & After the opponent presses the clock, execute the scheduled move and press the clock in time. \\
A2 & Basketball shot clock & A & Release a shot after READY and before the shot clock expires. \\
B1 & Whack-a-mole & B & Hit green moles with held mallets before they retract; avoid red decoys. \\
B2 & Reactive boxing & B & Strike the indicated pad after GO, then return to guard within the response limits. \\
B3 & Bench-press assistance & B & After the lifter stalls, catch the falling bar with both hands and return it to the rack. \\
B4 & Red-light freeze & B & Push on green; stop hands and tools after red and remain still until green returns. \\
B5 & Two-stick taiko & B & Strike each cued drum within its note window; paired notes require simultaneous hits. \\
C1 & Pool-ball racking & C & Place the required balls into the rack as quickly as possible, with stable released placements. \\
C2 & Speed cups & C & Build each displayed cup pattern, release stable cups and press the bell to advance. \\
C3 & Ball-machine service & C & Replenish the feeder for scheduled demand without exceeding hopper capacity. \\
\addlinespace[4pt]
\rowcolor{black!5}
\multicolumn{4}{@{}l@{}}{\textbf{Industry} (10 tasks: 3 A, 3 B, 4 C)} \\
\addlinespace[2pt]
A1 & Zincate dip and rinse & A & Pick the part from the loading pad and immerse it in the zincate bath; lift it out within the recipe window after full immersion, then release it in the rinse bath. \\
A2 & Rotary-carrier loading & A & Release the pump housing upright in the designated carrier while it crosses the loading sector. \\
A3 & Adhesive cover seating & A & Seat the housing cover within the allowed interval after adhesive application. \\
B1 & Leak isolation & B & Fully close the isolation valve within the response limit after the pipe starts leaking. \\
B2 & Motor emergency stop & B & Press the stop button within the response limit after the motor-test warning. \\
B3 & Bearing arrest and recovery & B & Stop the rolling bearing inside the safety area after zone entry, then recover it to the tray. \\
C1 & Maintenance-kit packing & C & Pack a bearing, bracket and pin into each case, release all items and move completed kits to output. \\
C2 & Carton shelving & C & Match carton SKUs to shelf identities, release cartons upright and withdraw before counting them. \\
C3 & Weigh-and-sort & C & Weigh each part on the bench scale, released and at rest, then sort it by the settled reading into a PASS or REJECT (underweight) tray. \\
C4 & Eyeshadow assembly & C & Place the four colors in their designated case positions, then move the completed palette to output. \\
\midrule
\multicolumn{4}{@{}l@{}}{\textbf{Total: 60 tasks; 19 A, 24 B and 17 C.}} \\
\bottomrule
\end{tabularx}
\end{table*}

\begingroup
\newcommand{\rcFrame}[2]{%
  \begin{minipage}[t]{0.192\textwidth}
    \centering
    \includegraphics[width=\linewidth]{figures/seq-#1}%
    \par\vspace{1pt}{\footnotesize $t=#2$\,s\par}
  \end{minipage}}
\newcommand{\rcSequenceTitle}[1]{%
  {\small\textbf{#1}\hfill\footnotesize Time $\longrightarrow$\par}\vspace{2pt}}

\begin{figure*}[p]
\centering
\rcSequenceTitle{(a) Household A3: Pancake flip}
\rcFrame{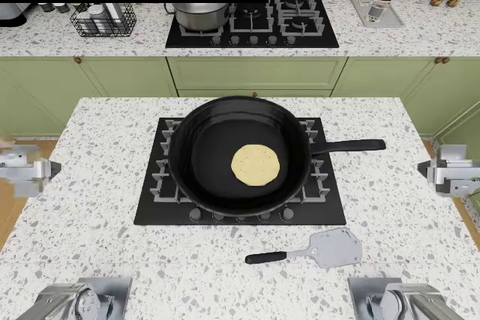}{0.00}\hfill
\rcFrame{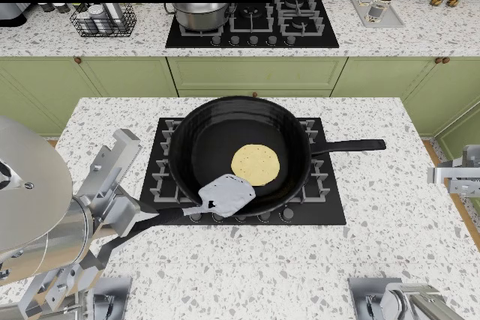}{30.00}\hfill
\rcFrame{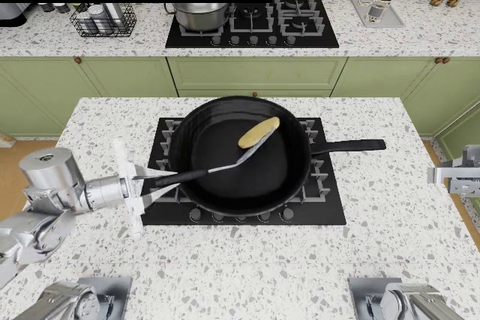}{41.00}\hfill
\rcFrame{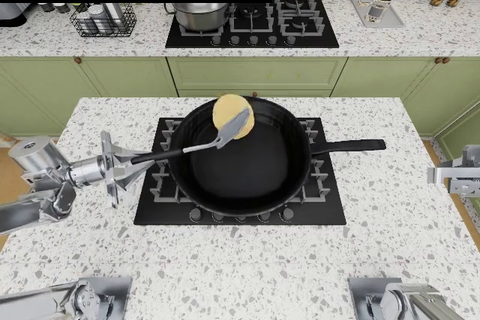}{42.60}\hfill
\rcFrame{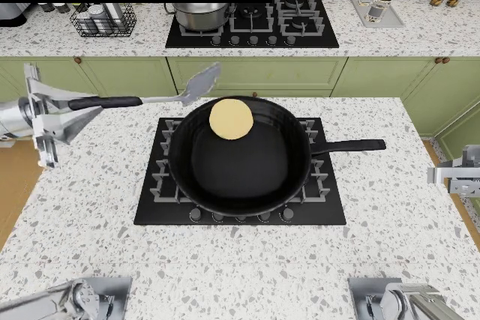}{45.00}
\par\vspace{5pt}
\rcSequenceTitle{(b) Laboratory A5: Film development}
\rcFrame{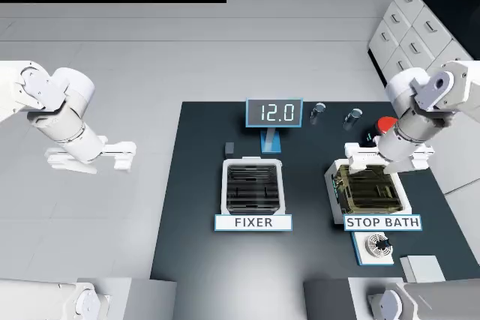}{0.00}\hfill
\rcFrame{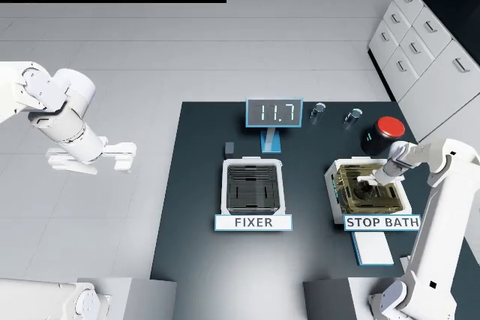}{4.48}\hfill
\rcFrame{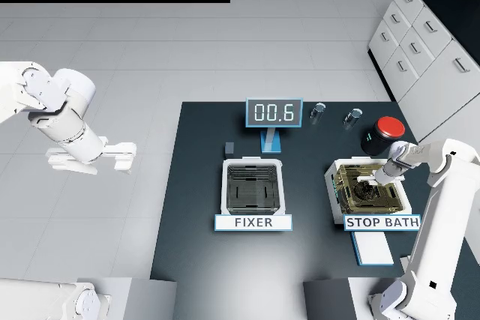}{15.52}\hfill
\rcFrame{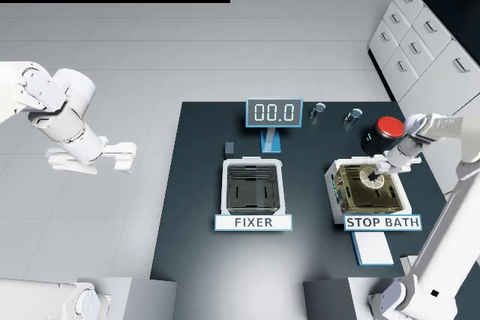}{16.60}\hfill
\rcFrame{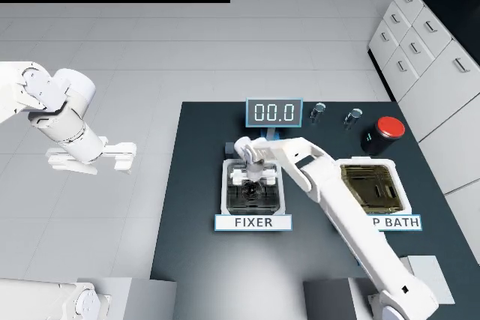}{18.60}
\caption{\textbf{Category A: Temporal Window.} (a) Wait until the underside is golden, then flip the pancake. (b) Immerse the reel in the stop bath, lift it once the 12\,s countdown ends, and release it in the fixer.}
\label{fig:task-demo-a}
\end{figure*}

\begin{figure*}[p]
\centering
\rcSequenceTitle{(a) Industry B2: Motor emergency stop}
\rcFrame{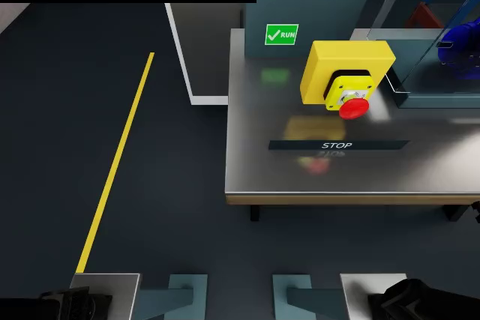}{0.00}\hfill
\rcFrame{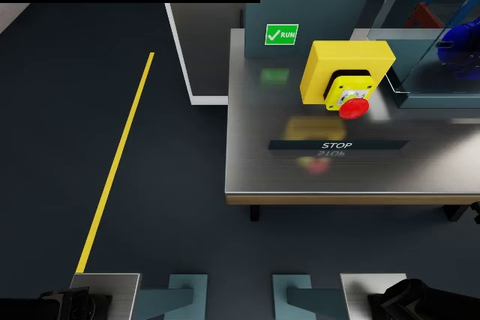}{14.00}\hfill
\rcFrame{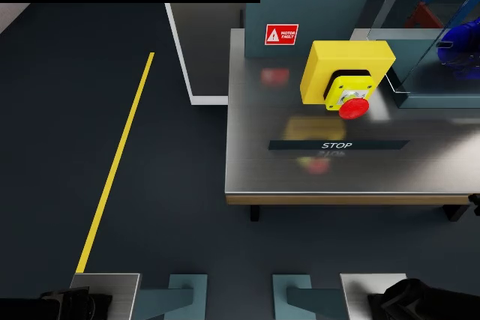}{14.68}\hfill
\rcFrame{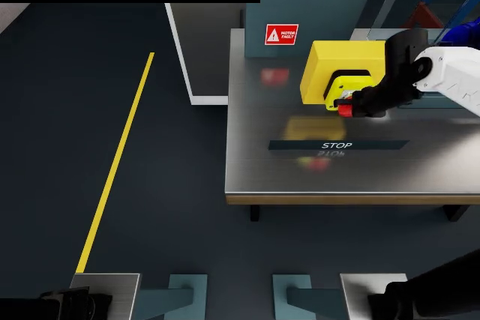}{16.60}\hfill
\rcFrame{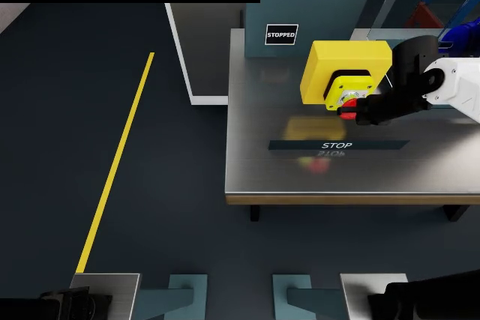}{19.00}
\par\vspace{5pt}
\rcSequenceTitle{(b) Healthcare B5: Anaphylaxis auto-injector}
\rcFrame{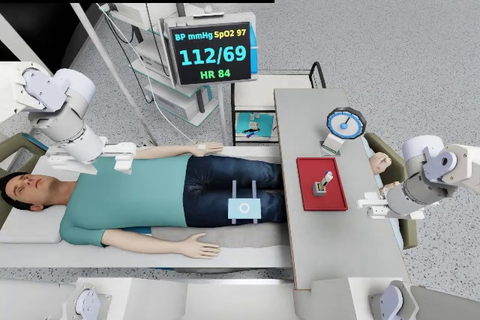}{4.00}\hfill
\rcFrame{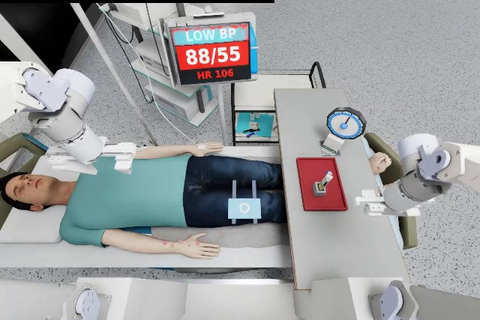}{7.96}\hfill
\rcFrame{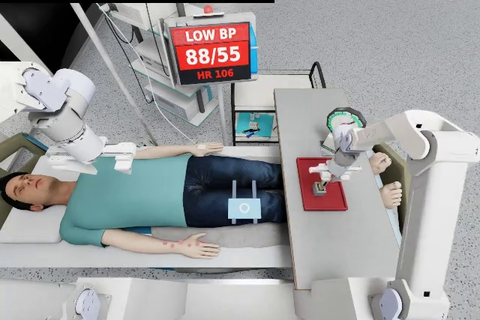}{9.32}\hfill
\rcFrame{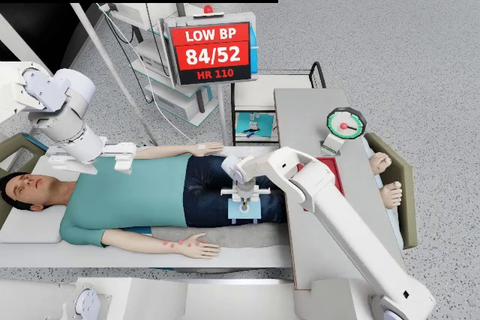}{11.60}\hfill
\rcFrame{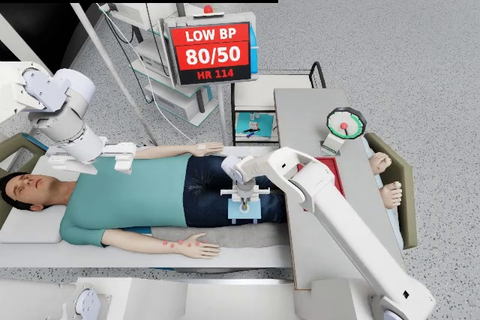}{14.56}
\caption{\textbf{Category B: Event-Reactive.} (a) The fault panel turns from RUN to MOTOR FAULT, and the robot actuates the emergency stop. (b) The LOW BP alarm triggers the auto-injector: grasp it, press its tip on the thigh pad, and hold for 3\,s.}
\label{fig:task-demo-b}
\end{figure*}

\begin{figure*}[p]
\centering
\rcSequenceTitle{(a) Laboratory C2: Color-matched rack loading}
\rcFrame{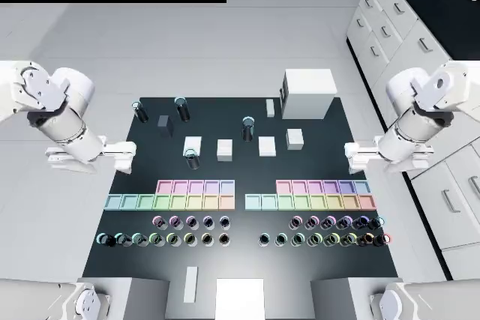}{0.00}\hfill
\rcFrame{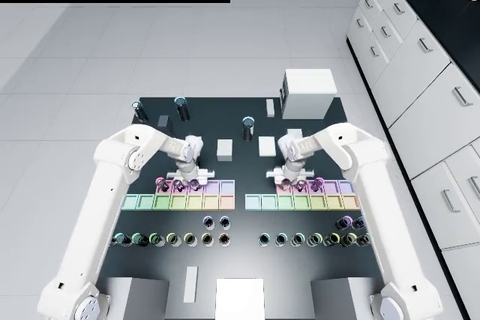}{15.00}\hfill
\rcFrame{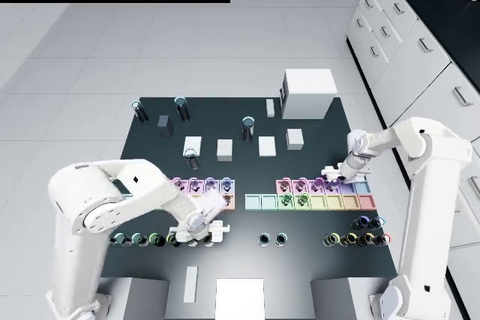}{32.00}\hfill
\rcFrame{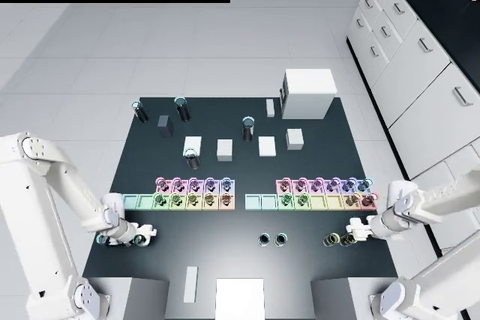}{54.00}\hfill
\rcFrame{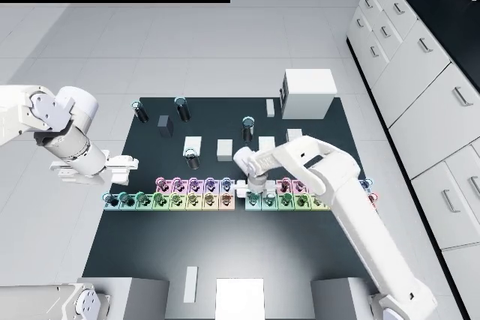}{74.00}
\par\vspace{5pt}
\rcSequenceTitle{(b) Sports \& games C3: Ball-machine service}
\rcFrame{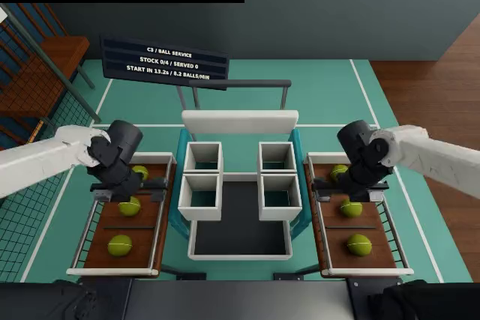}{0.00}\hfill
\rcFrame{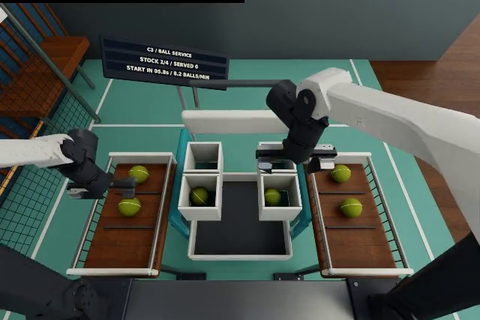}{8.00}\hfill
\rcFrame{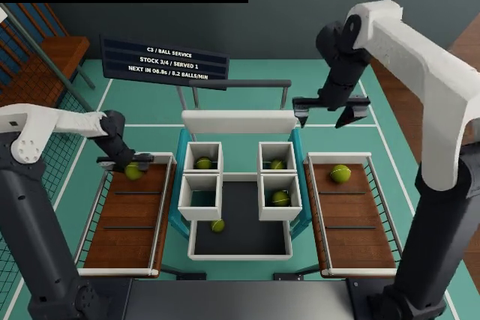}{16.00}\hfill
\rcFrame{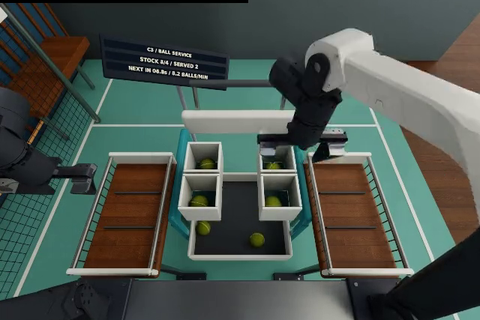}{24.00}\hfill
\rcFrame{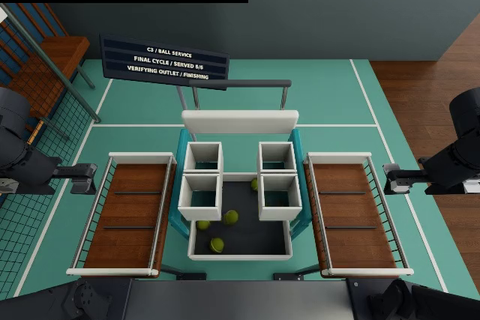}{51.52}
\caption{\textbf{Category C: Throughput.} (a) Repeated placements fill the sample racks. (b) Repeated feeder loading serves successive ball demands.}
\label{fig:task-demo-c}
\end{figure*}

\endgroup

\renewcommand{\dblfloatpagefraction}{0.99}

\section{Time Limits: Sources and Calibration}
\label{app:timing}\label{app:timing-cal}\label{app:timing-oracle}
Each level's time is set either by a physical parameter of the scene or by a time limit (Tabs.~\ref{tab:timing-process} and~\ref{tab:timing-limits}); Tabs.~\ref{tab:timing-lab}--\ref{tab:timing-sports} give the source of every value.

\paragraph{Windows set by a physical parameter (18 tasks, Tab.~\ref{tab:timing-process}).}
Here the window follows from a quantity in the scene (a flow rate, a band, a slope, a rotating sector, or the tolerance around a prescribed duration), taken from a real-world document where one exists. Medium and hard change the quantity itself, and no window is shorter than 1\,s.

\begin{table*}[t]
\centering\footnotesize
\setlength{\tabcolsep}{4pt}
\caption{Process-set windows: the quantity each level changes and the resulting time. E/M/H: easy/medium/hard. Lab.: Laboratory; Hlth.: Healthcare; House.: Household; Ind.: Industry.}
\label{tab:timing-process}
\begin{tabularx}{\textwidth}{@{}l Y l l@{}}
\toprule
Task & Quantity changed by the level & E / M / H & Resulting time (E / M / H) \\
\midrule
Lab.\ A1 & Accepted volume band below the 500\,mL mark, at a fixed 10\,mL/s & 440 / 460 / 480--500\,mL & 6 / 4 / 2\,s to close the faucet \\
Lab.\ A2 & Lower phase that may remain when the stopcock closes, at 5\,mL/s & 20 / 12.5 / 7.5\,mL & 4 / 2.5 / 1.5\,s \\
Lab.\ A3 & Gap between two eluting bands & 20 / 12 / 5\,s & the same, to exchange the tube \\
Lab.\ A4 & Titrant flow with the stopcock open (2\,mL endpoint band) & 0.33 / 0.50 / 1.0\,mL/s & 6 / 4 / 2\,s \\
Lab.\ A5 & Lateness allowed after the 12\,s stop bath & $+$4 / 2 / 1\,s & the same \\
Lab.\ A6 & Tolerance on the 30\,s warm bath & $\pm$4 / 2 / 1\,s & the same \\
Hlth.\ A1 & Time the chamber level stays in the 40--55\,\% band & 6 / 4.5 / 3\,s & the same \\
Hlth.\ A2 & Time from bag 1 empty until its drip chamber runs dry & 8 / 5.5 / 3\,s & the same \\
Hlth.\ A3 & Window around each breath, one every 6\,s & 3 / 2 / 1\,s & the same \\
House.\ A1 & Accepted sugar mass in the glass & 10--30 / 10--25 / 10--15\,g & at least 1.09\,s at hard at the fastest pour \\
House.\ A2 & Interval between ice cubes while the paddle is held & 2.0 / 1.5 / 1.3\,s & release window of at least 1\,s \\
House.\ A3 & Pan heat: browning rate relative to easy, which follows a recipe (first side about 2\,min) & $\times$1 / 1.33 / 2 & first side golden after 69--120 / 52--90 / 35--60\,s \\
House.\ B1 & Faucet flow & 4.5 / 6.0 / 8.3\,L/min & overflow after 10.7--18.7 / 8--14 / 5.8--10.1\,s \\
House.\ B2 & Slope of the board the object rolls down & 4.5 / 6 / 9$^\circ$ & 1.2--2.3\,s to the counter edge (about 1.2\,s at hard) \\
House.\ B3 & Growth rate of the pan fire & $\times$0.75 / 1 / 1.5 & flame spreads after 48--90 / 36--68 / 24--45\,s \\
Ind.\ A2 & Loading sector of the table turning at 5$^\circ$/s & 60 / 50 / 40$^\circ$ & 12 / 10 / 8\,s \\
Sports B4 & Time allowed to come to rest in each red phase & 8 / 4 / 2\,s & the same \\
Sports B5 & Window around each drum note, notes 5\,s apart & $\pm$2 / 1.41 / 1\,s & the same \\
\bottomrule
\end{tabularx}
\end{table*}

\paragraph{Time limits (42 tasks, Tab.~\ref{tab:timing-limits}).}
In the other tasks the requirement is a deadline, a budget, or a throughput tier, set against $t$, the slowest of ten oracle runs (seeds 0--9). Hard is $\max(1.25t,\ t+0.30\,\text{s},\ 1\,\text{s})$, so a policy as fast as the oracle keeps a fifth of the limit to perceive and decide. Easy is a documented real-world time where one exists (\eg, a ventilator's 20\,s apnea interval or the 14\,s shot clock) and $4t$ otherwise; medium is the geometric mean. Throughput tiers use $4$, $\sqrt5$, and $1.25\times t_o$ by default; Sports C3 uses the process-capped schedule in Tab.~\ref{tab:timing-sports}.

\begin{table*}[t]
\centering\footnotesize
\setlength{\tabcolsep}{4pt}
\caption{Time limits. $t$: slowest reference time over seeds 0--9. Easy basis: \emph{doc}, a documented real-world time; $4t$ where none exists; \emph{cap}, the time the scene's process allows where $4t$ would exceed it. Hard is $\max(1.25t,\ t+0.30\,\text{s},\ 1\,\text{s})$ and medium is $\sqrt{\text{easy}\times\text{hard}}$. The 17 Throughput tasks use $4/\sqrt5/1.25\times t_o$ by default, with the process-capped exception of Sports C3 (Tabs.~\ref{tab:timing-lab}--\ref{tab:timing-sports}). Abbreviations as in Tab.~\ref{tab:timing-process}.}
\label{tab:timing-limits}
\begin{tabularx}{\textwidth}{@{}l Y r l l@{}}
\toprule
Task & Limit & $t$ (s) & Easy basis & E / M / H (s) \\
\midrule
Lab.\ B1 & Alarm to beaker lifted & 1.17 & $4t$ & 4.69 / 2.63 / 1.48 \\
Lab.\ B1 & Alarm to beaker placed in the tray & 5.08 & $4t$ & 20.32 / 11.36 / 6.35 \\
Lab.\ B2 & Red indicator to knife switch open & 1.50 & $4t$ & 6.02 / 3.36 / 1.88 \\
Lab.\ B3 & Motor stall to power cut & 1.59 & $4t$ & 6.37 / 3.56 / 1.99 \\
Lab.\ B4 & Tube tilt to tube caught & 1.69 & $4t$ & 6.76 / 3.79 / 2.12 \\
Lab.\ B5 & Door alarm to freezer door closed & 1.82 & $4t$ & 7.28 / 4.07 / 2.28 \\
Lab.\ B6, B7 & Warning to plunger pressed; alarm to inlet clamped & 0.32 & $4t$ & 1.3 / 1.14 / 1.0 \\
Hlth.\ A4 & Instruction to pump START pressed & 6.66 & $4t$ & 26.64 / 14.9 / 8.33 \\
Hlth.\ B1 & Alarm to oxygen float settled in the band & 4.15 & doc: nurses' median time to reach an alarm & 55 / 16.9 / 5.19 \\
Hlth.\ B2 & Arrest to shock (low end of a seeded range) & 23.13 & $4t$ & 92.52 / 51.73 / 28.92 \\
Hlth.\ B3 & Break to both glass pieces binned (low end) & 14.61 & $4t$ & 58.44 / 32.68 / 18.27 \\
Hlth.\ B4 & Disconnection to circuit reseated & 3.14 & doc: ventilator apnea interval & 20 / 8.87 / 3.93 \\
Hlth.\ B5 & Alarm to end of the 3\,s injector hold & 6.90 & doc: untrained adults' mean time & 33.7 / 17.05 / 8.63 \\
Hlth.\ B6 & Bleeding onset to windlass secured & 4.35 & doc: trained providers' time & 24 / 11.43 / 5.44 \\
House.\ A4 & Start to kettle set down & 28.69 & $4t$ & 115 / 64 / 36 \\
Ind.\ A1 & Width of the exit window after the minimum dip & 8.11 & doc: 50\,s longest first dip & 28 / 16.85 / 10.14 \\
Ind.\ A3 & Bead end to cover seated & 13.17 & doc: 40\,s open time & 40 / 25.67 / 16.47 \\
Ind.\ B1 & Leak onset to valve closed & 5.02 & cap: spill tray full & 14 / 9.38 / 6.28 \\
Ind.\ B2 & Warning to stop button actuated & 2.02 & 3\,s, $3.3\times$ the slowest measured press & 3 / 2.75 / 2.53 \\
Ind.\ B3 & Zone entry to stable arrest & 1.19 & $4t$ & 4.77 / 2.67 / 1.50 \\
Sports A1 & Opponent's press to own clock press & 7.90 & doc: 30\,s increment & 30 / 17 / 10 \\
Sports A2 & READY to shot released & 4.59 & doc: 14\,s shot clock & 14 / 9 / 6 \\
Sports B1 & Mole rise to hit & 1.76 & cap: 4\,s between moles & 4.0 / 3.0 / 2.2 \\
Sports B2 & GO cue to pad hit; hit to guard & 0.68; 0.72 & cap: next cue & 1.6 / 1.25 / 1.0; 1.4 / 1.2 / 1.05 \\
Sports B3 & Bar descent to catch & 0.51 & $4t$ & 2.05 / 1.43 / 1.0 \\
\bottomrule
\end{tabularx}
\end{table*}

\begin{table*}[p]
\centering
\footnotesize
\setlength{\tabcolsep}{4pt}
\renewcommand{\arraystretch}{1.0}
\caption{Time-parameter provenance: Laboratory. (a) Tasks whose window is set by a physical parameter of the scene (Tab.~\ref{tab:timing-process}); their values are not sourced. (b) Tasks with a time limit (Tab.~\ref{tab:timing-limits}) and Throughput tasks; each value is compared with a documented real-world time.}
\label{tab:timing-lab}
\begin{tabularx}{\textwidth}{@{}l p{3.2cm} p{2.6cm} Y@{}}
\toprule
\multicolumn{4}{@{}l}{\textit{(a) Window set by a physical parameter (Tab.~\ref{tab:timing-process})}} \\
Task & Parameter & Value (E/M/H) & Note \\
\midrule
A1 & Faucet flow, handle fully open & 10\,mL/s at every tier & 500\,mL fill in 50\,s; design setting for a low-flow metering outlet. \\
A1 & Accepted band below the 500\,mL mark & 440/460/480--500\,mL & 6/4/2\,s of flow before the mark. Oracle margin (hard): 1.04\,s early side, 0.96\,s late side. \\
A2 & Drain rate; lower phase & 5\,mL/s; 120\,mL & One constant-flow pass; the interface reaches the stopcock 24\,s after opening. \\
A2 & Lower phase that may remain when the stopcock closes & 20/12.5/7.5\,mL & 4/2.5/1.5\,s of drainage before contamination. Oracle closes 0.76\,s before contamination. \\
A3 & Eluent flow & 1.5\,mL/s (90\,mL/min) & Flash-chromatography solvent descent in a 47\,mm column (88\,mL/min). \\
A3 & Band volume & 13.5\,mL per band & 9\,s per band at the eluent flow; a run lasts 63--108\,s. \\
A3 & Gap between bands (tube switch) & 20/12/5\,s & Time to exchange the tube between two bands; the oracle needs about 1.9\,s per switch. \\
A4 & Titrant flow, stopcock open & 0.33/0.50/1.0\,mL/s & 6/4/2\,s to pass the 2\,mL endpoint band. \\
A4 & Endpoint band (first pink to excess) & 18--20\,mL & 6/4/2\,s at the tier's full flow. Oracle hard leaves 0.46\,mL. \\
A5 & Stop-bath time before the fixer & 12\,s & Prescribed bath time; the lateness tolerance below runs from its end. \\
A5 & Lift lateness tolerance (late side only) & $+$4/2/1\,s & Hard is the 1\,s floor. Oracle lateness 0.052\,s. \\
A5, A6 & Loading limit (A5: reel into the stop bath; A6: tube picked from the bench stand) & 15\,s (untiered) & Design setting; oracle immersed after 4.08\,s. \\
A5 & Release in the fixer after the lift & 5\,s (untiered) & Design setting; oracle releases at most 2.40\,s after the lift. \\
A6 & Warm-bath (heat-shock) duration & 30\,s & Prescribed bath time; the exit tolerance below is centred on it. \\
A6 & Bath-exit tolerance & $\pm$4/2/1\,s & Hard is the 1\,s floor. Oracle error 0.000\,s at every tier. \\
A6 & Transfer to ice after bath exit & 5\,s (untiered) & Design setting; oracle 3.14\,s. \\
\midrule
\multicolumn{4}{@{}l}{\textit{(b) Time limit or throughput tier (Tab.~\ref{tab:timing-limits})}} \\
Task & Parameter & Value (E/M/H) & Source and relation to our value \\
\midrule
B1--B7 & Event onset after the episode start & 4--10\,s, uniform (all tiers) & Constant foreperiods let subjects predict the onset, randomly intermixed ones do not~\cite{tllab_correa2006,tlhospital_forgaard2016}; we draw it per seed. \\
B1 & Lift deadline after the alarm & 4.69/2.63/1.48\,s & $t=1.17$\,s; easy $4t$. Drivers' perception--brake time is 1.25\,s for unexpected and 1.5\,s for surprise events~\cite{tllab_green2000}; hard is 0.99$\times$, easy 3.1$\times$ the surprise time. \\
B1 & Placement deadline in the tray & 20.32/11.36/6.35\,s & $t=5.08$\,s; easy $4t$. Analogy only: surprise reaction 1.5\,s~\cite{tllab_green2000}, fast 30\,cm reach-to-grasp 0.42\,s~\cite{tllab_runnarong2019}, block transfer 0.78\,s~\cite{tllab_mathiowetz1985}: 2.7\,s; hard is 2.4$\times$ that. \\
B2 & Knife-switch isolation deadline & 6.02/3.36/1.88\,s & $t=1.50$\,s (easy $4t$), incl.\ 0.6\,s oracle cue delay. Subjects stopped an unexpected robot motion by releasing a held button in 0.53\,s on average, 1.42\,s at mean${}+2$\,SD~\cite{tllab_helander1987}; hard is 1.3$\times$ the latter. \\
B3 & Power cut after motor stall & 6.37/3.56/1.99\,s & $t=1.59$\,s; easy $4t$. Hard is 1.4$\times$ the 1.42\,s (mean${}+2$\,SD) robot-stop response~\cite{tllab_helander1987}, within the ``order of seconds'' to stall damage of small DC motors~\cite{tllab_pololufaq}. \\
B4 & Catch the tube after 32$^\circ$ tilt & 6.76/3.79/2.12\,s & $t=1.69$\,s with the rig centred on the tube; easy $4t$. Simple visual reaction (0.21--0.23\,s~\cite{tl_woods2015srt}) plus a fast 30\,cm reach-to-grasp (0.42\,s~\cite{tllab_runnarong2019}) is 0.63--0.65\,s; hard is 3.3$\times$ that. \\
B5 & Freezer-door closing deadline & 7.28/4.07/2.28\,s & $t=1.82$\,s with the freezer facing the rig, incl.\ 0.3\,s cue delay; easy $4t$. Shortest documented opening 10--15\,s~\cite{tllab_thermoult}: ours 1.4--6.6$\times$ shorter; real door alarms wait (default 2\,min~\cite{tllab_phcbidu700}). \\
B6 & Vent-plunger deadline & 1.3/1.14/1.0\,s & $t=0.32$\,s; easy $4t$; hard is the 1\,s floor. Simple visual reaction (0.21--0.23\,s~\cite{tl_woods2015srt}) plus a fast reach (0.42\,s~\cite{tllab_runnarong2019}) is 0.63--0.65\,s; ours is 1.5--2.1$\times$ that. \\
B7 & Inlet-clamp deadline & 1.3/1.14/1.0\,s & As B6 (human reaction plus reach 0.63--0.65\,s~\cite{tl_woods2015srt,tllab_runnarong2019}; ours 1.5--2.1$\times$). Inflow is 10\,mL of headroom divided by the deadline. \\
\midrule
C1 & Time per tube, $N$=20--30 tubes & 15.908/8.893/\allowbreak 4.971\,s per item, $\times N$ total & $t_o=3.977$\,s/tube. Hard = 724 tubes/h; commercial sorters do 1{,}150--1{,}200/h~\cite{tllab_yaskawa,tllab_atellica}, 1.6--1.7$\times$ faster. \\
C2 & Time per sample, $N$=20--30 & 11.228/6.277/\allowbreak 3.509\,s per item, $\times N$ total & $t_o=2.807$\,s/item. Hard = 1{,}026/h, 86--89\,\% of the sorter rates~\cite{tllab_yaskawa,tllab_atellica} (different machine). \\
C3 & Time per sample, $N$=20--30 & 12.100/6.764/\allowbreak 3.781\,s per item, $\times N$ total & $t_o=3.025$\,s/item. Adults move one block over a partition in 0.72--0.78\,s~\cite{tllab_mathiowetz1985}; hard is 4.8--5.2$\times$ that, as each sample is also aligned to a marker axis. Order of magnitude only. \\
\bottomrule
\end{tabularx}
\end{table*}

\begin{table*}[p]
\centering
\footnotesize
\setlength{\tabcolsep}{4pt}
\renewcommand{\arraystretch}{1.08}
\caption{Time-parameter provenance: Healthcare. (a) Tasks whose window is set by a physical parameter of the scene (Tab.~\ref{tab:timing-process}); their values are not sourced. (b) Tasks with a time limit (Tab.~\ref{tab:timing-limits}) and Throughput tasks; each value is compared with a documented real-world time.}
\label{tab:timing-hospital}
\begin{tabularx}{\textwidth}{@{}l p{3.2cm} p{2.6cm} Y@{}}
\toprule
\multicolumn{4}{@{}l}{\textit{(a) Window set by a physical parameter (Tab.~\ref{tab:timing-process})}} \\
Task & Parameter & Value (E/M/H) & Note \\
\midrule
A1 & Drip-chamber level band & 40--55\,\%, all levels & Design setting near the usual one-third to one-half fill; the level rate sets the window below. \\
A1 & Window: level inside the band, counted from the robot's own opening of the clamp & 6 / 4.5 / 3\,s & The level is still until the clamp is opened and reaches the band 3--5\,s later (seeded). \\
A2 & Window: bag~1 empty until its drip chamber runs dry (two 100\,mL bags, gravity, wide open) & 8 / 5.5 / 3\,s & Set by the drawn chamber (about 79\,mL) and the gravity flow; bag~1 empties 5--7.5\,s into the episode (seeded). \\
A3 & Breath period (4 breaths) & 6\,s (10/min), all levels & One beat every 6\,s on the on-screen timer; the first beat comes 7--9\,s into the episode (seeded). \\
A3 & Window around each beat & 3 / 2 / 1\,s & Hard keeps every interval within 5--7\,s; easy allows 3--9\,s. Our squeeze takes 0.65\,s. \\
\midrule
\multicolumn{4}{@{}l}{\textit{(b) Time limit or throughput tier (Tab.~\ref{tab:timing-limits})}} \\
Task & Parameter & Value (E/M/H) & Source and relation to our value \\
\midrule
A4 & Limit, instruction to end of START press (bag spiked and hung first) & 26.64 / 14.9 / 8.33\,s & Easy ($4t$, oracle 6.66\,s) is close to the 26.5\,s mean critical-care nurses took, after training, to program a secondary antibiotic infusion on the fastest of three pumps (36.6 and 58.8\,s on the others, up to 101\,s before training) \cite{tlhospital_giuliano2015}; ours also spikes and hangs the bag. \\
B1--B6 & Cue onset (alarm, arrest, break, disconnection or bleeding) & 4--10\,s (seeded) & Randomized per seed, as reaction-time studies vary the foreperiod ``to prevent anticipation'' \cite{tlhospital_forgaard2016}; real alarms and arrests are likewise unannounced. \\
\addlinespace[2pt]
B1 & Ordered oxygen flow (simple face mask), float held 0.5\,s & 5, 6 or 7\,L/min $\pm$1 (by seed), all levels & Inside the 5--10\,L/min initial flow for a simple face mask in the BTS guideline~\cite{tlhospital_bts2017}; the alarm fires when SpO$_2$ crosses 90\,\%, below the guideline's 94--98\,\% target range. \\
B1 & SpO$_2$ fall before and after the alarm; recovery & 0.12 and 0.6\,\%/s; 2.5\,\%/s after 0.8\,s & Before the alarm the fall runs at the real rate: an apnoeic patient on room air desaturates in 45--60\,s \cite{tlhospital_weingart2012}, about 6\,\% in 50\,s, and the episode starts mid-fall, 0.5--1.2\,\% above the alarm line. The rate after the alarm and the recovery are design settings: saturation is to be observed for at least 5\,min after an increase \cite{tlhospital_bts2017}, while ours recovers in about 4\,s, after the scored response. \\
B1 & Response limit, alarm to float settled in the band & 55 / 16.9 / 5.19\,s & Easy is the median 55\,s after which nurses outside the room reached critical alarms, over half of them desaturations (26\,\% were answered within 90\,s) \cite{tlhospital_joshi2017}. Hard is from the oracle (4.15\,s, about 2.45\,s of it the reach). \\
B2 & AED analysis and charge, pads on to SHOCK ready & 8\,s & Matches ``typically in 8 seconds'' \cite{tlhospital_frx}; one other AED is ready within at most 12\,s \cite{tlhospital_zollaedplus}. \\
B2 & Response limit, arrest to shock, incl.\ the 8\,s analysis (seeded range) & 92.52--97.52 / 51.73--56.66 / 28.92--32.92\,s & In-situ in-hospital simulations: AED arrival to first shock about 122\,s (difference of medians) \cite{tlhospital_staerk2022}; ours is 1.3--4.2$\times$ shorter (AED at hand, no power-on). Oracle 23.13\,s; easy $4t$ at the low end of the range. \\
B3 & Response limit, break to both glass pieces in the sharps bin (seeded range) & 58.44--64.44 / 32.68--37.88 / 18.27--22.27\,s & Spills ``must be contained and cleaned immediately''~\cite{tlhospital_usp800}, no time given. In surprise drills with a broken vial, trained pharmacy staff took a mean 11.9\,min (17.3\,min before training) for the whole procedure \cite{tlhospital_garnier2025}; our two steps are limited to 11--39$\times$ less (oracle $t=14.61$\,s; easy $4t$ at the low end of the range). \\
B3 & Containment deadline (towel on the puddle), $0.4\times$ the response limit & 23.38--25.78 / 13.07--15.15 / 7.31--8.91\,s & Spills ``must be contained and cleaned immediately'' \cite{tlhospital_usp800}, with no time given; trained staff took a mean 11.9\,min for a whole broken-vial spill drill \cite{tlhospital_garnier2025}, 28--98$\times$ our containment step (order of magnitude only). The puddle reaches its full 0.10\,m radius at the deadline; the oracle lays the towel by 4.35\,s. \\
\addlinespace[2pt]
B4 & Response limit, disconnection to elbow reseated, released and retained & 20 / 8.87 / 3.93\,s & Easy is the adult default apnoea interval of an ICU ventilator (20\,s) \cite{tlhospital_pb840}, close to the 23.3\,s that logged alarms for unintended disconnections lasted on average (10--55\,s) \cite{tlhospital_evans2005}; medium and hard are below the shortest logged alarm. All end before room-air apnoea desaturates (45--60\,s) \cite{tlhospital_weingart2012}. Hard from the oracle (3.14\,s). \\
\bottomrule
\end{tabularx}
\end{table*}

\begin{table*}[p]
\ContinuedFloat
\centering
\footnotesize
\setlength{\tabcolsep}{4pt}
\renewcommand{\arraystretch}{1.12}
\caption{Time-parameter provenance: Healthcare (continued).}
\begin{tabularx}{\textwidth}{@{}l p{3.2cm} p{2.6cm} Y@{}}
\toprule
\multicolumn{4}{@{}l}{\textit{(b) Time limit or throughput tier (Tab.~\ref{tab:timing-limits}; continued)}} \\
Task & Parameter & Value (E/M/H) & Source and relation to our value \\
\midrule
B5 & Hold after the auto-injector fires & 3\,s, all levels & The device's own instruction: ``firmly push the orange tip \ldots\ against the outer thigh and hold in place for 3 seconds'', at 90$^\circ$ to the thigh, through clothing if necessary \cite{tlhospital_epipen2026}. \\
B5 & Response limit, low-blood-pressure alarm to end of the 3\,s hold & 33.7 / 17.05 / 8.63\,s & Easy is the mean 33.72\,s untrained adults took to give an EpiPen in a simulated emergency (Jext 29.21\,s, Emerade 14.73\,s) \cite{tlhospital_knibb2015}; clinical teams took a median 95\,s (IQR 77--252) with an auto-injector in in-situ simulations \cite{tlhospital_maa2020}. Guidance gives no time: adrenaline is first-line and given early \cite{tlhospital_rcuk2021}. Hard is from the oracle (6.90\,s at the farthest tray-to-pad placement, 3\,s of it the hold). \\
B5 & Systolic fall before and after the alarm; recovery after the dose & 0.1 and 1\,mmHg/s; 12\,mmHg/s after 1\,s & Before the alarm the pressure falls on the scale of the sources, minutes from baseline to the alarm: fatal reactions to injected drugs reached arrest after a median 5\,min \cite{tlhospital_pumphrey2000}; the episode starts mid-fall, within 1\,mmHg of the alarm line. The fall after the alarm and the recovery are design settings; the recovery, shown after the scored response, is faster than the 8$\pm$2\,min to peak plasma adrenaline \cite{tlhospital_simons1998}. \\
B6 & Windlass travel to the retaining clip (seeded); bleeding stopped & 164--196$^\circ$; 30$^\circ$ before the clip, all levels & One 180$^\circ$ windlass turn occludes a correctly strapped Combat Application Tourniquet, and the rod is secured in the clip ``at a rotation beyond that at which bleeding is stopped'' \cite{tlhospital_wall2019}. The $\pm$16$^\circ$ spread (rod start and clip angles) and the 30$^\circ$ are design settings. \\
B6 & Bleeding rate with the windlass untwisted & 4\,mL/s, all levels & A thigh bleeding trainer lost 99$\pm$17\,mL of simulated blood during a 24$\pm$4\,s application, 4.1\,mL/s \cite{tlhospital_lechner2024}. Falls linearly to zero 30$^\circ$ before the clip (design). \\
B6 & Response limit, bleeding onset to rod released in the clip (0.3\,s dwell) & 24 / 11.43 / 5.44\,s & Easy is the 24$\pm$4\,s experienced providers took to apply the tourniquet to a thigh trainer, timed until the windlass was secured \cite{tlhospital_lechner2024}; ours starts with the band placed. Hard from the oracle (4.35\,s). Severe bleeding can kill ``in as little as five minutes'' \cite{tlhospital_acsstb2025}. \\
\addlinespace[2pt]
C1 & Budget, 8 vials into 4 order bins; $t_o=16.09$\,s & 64.4 / 36.0 / 20.1\,s (8.0 / 4.5 / 2.5\,s per vial) & Pick-to-light reaches 300--1200 lines/h, 3--12\,s per line \cite{tlhospital_ciscoeagle2018}: easy, medium inside, hard $1.2\times$ faster. Pharmacists: median 60\,s per prescription \cite{tlhospital_takase2022}. \\
C2 & Budget, 6 instruments into tray slots; $t_o=18.27$\,s & 73.1 / 40.9 / 22.8\,s (12.2 / 6.8 / 3.8\,s per item) & Central-processing records give a mean 17.7\,s to pack one instrument into a surgical tray (7.6--31.6\,s) \cite{tlhospital_chin2014}; ours are 1.5 / 2.6 / 4.7$\times$ faster (no inspection). A vendor reports 15--28\,min per whole tray \cite{tlhospital_steris}. \\
C3 & Budget, 12 vials into drawer lanes; $t_o=28.44$\,s & 113.8 / 63.6 / 35.5\,s (9.5 / 5.3 / 3.0\,s per vial) & Pharmacy technicians restocked an automated dispensing cabinet in 11.5\,s per restock (21.3\,s after a layout change) \cite{tlhospital_kelm2018}; ours are 1.2 / 2.2 / 3.8$\times$ faster per vial. \\
C4 & Budget, 12 tubes into 3 racks by stopper color; $t_o=26.34$\,s & 105.4 / 58.9 / 32.9\,s (410 / 733 / 1313 tubes/h) & Automated sorters are rated up to 1150 \cite{tlhospital_atellica} and 1400 tubes/h \cite{tlhospital_cobasp512}; hard lies between them, easy and medium are 0.3--0.6$\times$ those rates. \\
\bottomrule
\end{tabularx}
\end{table*}

\begin{table*}[p]
\centering\footnotesize
\setlength{\tabcolsep}{4pt}
\renewcommand{\arraystretch}{1.05}
\caption{Time-parameter provenance: Household. (a) Tasks whose window is set by a physical parameter of the scene (Tab.~\ref{tab:timing-process}); their values are not sourced. (b) Tasks with a time limit (Tab.~\ref{tab:timing-limits}) and Throughput tasks; each value is compared with a documented real-world time.}
\label{tab:timing-household}
\begin{tabularx}{\textwidth}{@{}l p{3.2cm} p{2.6cm} Y@{}}
\toprule
\multicolumn{4}{@{}l}{\textit{(a) Window set by a physical parameter (Tab.~\ref{tab:timing-process})}} \\
Task & Parameter & Value (E/M/H) & Note \\
\midrule
A1 & Sugar in the glass when the jar is set down (gram band) & 10--30 / 10--25 / 10--15\,g & About 2.4--7.1 / 2.4--6.0 / 2.4--3.6 teaspoons; the only cue is the visible sugar heap. \\
A1 & Sugar flow once the jar tilts past its onset (40--60$^\circ$, later when emptier)\newline\textit{oracle: pours at 2.5\,g/s, stops at the band middle} & $\le$4\,g/s at full tilt, $\times$0.85--1.15 per seed, same at every tier & At the fastest pour the hard band lasts $\ge$1.09\,s (5\,g at 4.6\,g/s), the 1\,s floor. \\
A1 & Spill limit & 2\,g & Design setting. \\
\addlinespace[2pt]
A2 & Cube interval while the paddle is held\newline\textit{oracle: releases 0.6\,s after the last cube} & 2.0 / 1.5 / 1.3\,s nominal; per seed $\times$0.93--1.13, $\pm$5\,\% jitter & H is capped at 1.15$\times$ the M rate so that the release window (first cube in the band to the next cube) is at least 1\,s (1.61 / 1.21 / 1.08\,s worst case). \\
A2 & Spin-up to first cube; run-on after release & 0.6--1.3\,s (M, $\div$ tier scale); 0.12\,s (M) & Simulation settings of the dispenser. \\
A2 & Ice per cube; mass rate & 8.5\,g; 4.3 / 5.7 / 6.5\,g/s & Mass rate = cube mass / cube interval above. \\
\addlinespace[2pt]
A3 & Pan heat: underside browning 0.55--0.75 (hidden $\tau$ 50--80\,s, start age 15--25\,s), browning rate $\times$1 / 1.33 / 2\newline\textit{oracle: flips $\le$9.8\,s after band entry (H)} & flip window 42.5--93.5 / 31.9--70.1 / 21.3--46.7\,s nominal & First side golden after 69--120 / 52--90 / 35--60\,s; easy is set to a common recipe time (about 2\,min), medium and hard are a hotter pan. \\
\addlinespace[2pt]
B1 & Faucet flow, lever fully open & 4.5 / 6.0 / 8.3\,L/min & H equals the US maximum kitchen-faucet flow (2.2\,gpm). \\
B1 & Room left in the bowl; task fails at 0.5\,L spilled\newline\textit{oracle: flow stopped at 2.42\,s} & 0.8--1.4\,L of a 1.62\,L bowl & Overflow after 10.7--18.7 / 8--14 / 5.8--10.1\,s; earliest overflow in oracle runs 6.57\,s (H). \\
\addlinespace[2pt]
B2 & Board slope\newline\textit{oracle: stops the object $\ge$4.4\,cm from the edge} & 4.5 / 6 / 9$^\circ$ & Rolling under gravity with damping: 1.2--2.3\,s from release to the counter edge over objects and tiers (H ${\approx}$1.2\,s). Real counters are level; the slope starts the roll. \\
B2 & Release onset & U[1,\,3]\,s & Randomized so the release cannot be anticipated. \\
\addlinespace[2pt]
B3 & Pan-fire growth rate while the burner is on\newline\textit{oracle: burner off at 5.6\,s} & $\times$0.75 / 1 / 1.5 & Flame reaches 0.8 after 48--90 / 36--68 / 24--45\,s (nominal 64 / 48 / 32\,s). \\
B3 & Flame decay after burner off & $\tau=1$\,s & Abstraction: the flame dies out within a few seconds once the burner is off. \\
B3 & Flame out for (success) & 3\,s & Confirms the fire stays out; design setting. \\
\addlinespace[2pt]
All & A2--B3 process-rate scale (rates $\times$, deadlines $\div$) & 0.75 / 1 / 1.5 & Design setting; deadlines shrink accordingly. A1 is tiered by its band only. \\
\midrule
\multicolumn{4}{@{}l}{\textit{(b) Time limit or throughput tier (Tab.~\ref{tab:timing-limits})}} \\
Task & Parameter & Value (E/M/H) & Source and relation to our value \\
\midrule
A4 & Kettle pour flow at full tilt & $\le$40\,mL/s, $\pm$15\,\% per seed (300\,g in $\ge$7.5\,s) & Human volume-targeted pours take 3.2--8.7\,s~\cite{tlhousehold_huang}, the scale of our minimum pour; one measured gooseneck kettle peaks at 24\,g/s~\cite{tlhousehold_gagne}, 1.7$\times$ below ours. \\
A4 & Budget: kettle set down and released, from $t=0$ & 115 / 64 / 36\,s & A measured gooseneck kettle peaks at 24\,g/s~\cite{tlhousehold_gagne}, so 300\,g takes at least 12.5\,s to pour; H allows 2.9$\times$ that for reach, pour and set-down. Human volume-targeted pours take 3.2--8.7\,s~\cite{tlhousehold_huang}. \\
A4 & Dose tolerance; read delay; scale display lag & $\pm$10\,g; 3\,s; 0.3\,s & A bench scale settles in ``2 seconds typical''~\cite{tlhousehold_adamcbc}: our read delay is 1.5$\times$ that, the display lag well inside it. The $\pm$10\,g tolerance is a design setting (oracle error $\le$1.7\,g). \\
\addlinespace[2pt]
C1 & Waiter pickups, 4 orders (stream); reference schedule $10.53+12.96\,k$\,s & first 42.12 / 23.55 / 13.16\,s, every 51.84 / 28.98 / 16.20\,s; last 197.64 / 110.48 / 61.76\,s & Reference = slowest oracle set-downs, $\times$4 / $\sqrt5$ / 1.25. Analogy only: a fast-food drive-thru visit, line entry to leaving with food, averaged 5\,min 43\,s over ten US chains in 2023~\cite{tlhousehold_nacs2024}; our dishes are pre-made and the first pickup comes 8--26$\times$ sooner. \\
C2 & Six glasses, $t_o=43.21$\,s & 172.8 / 96.6 / 54.0\,s & $4/\sqrt5/1.25\,t_o$. H leaves 9.0\,s per glass (two arms); a balanced draught tap fills a pint in about 8\,s~\cite{tlhousehold_ba2011}, and human volume-targeted pours take 3.2--8.7\,s~\cite{tlhousehold_huang}. \\
C3 & Six utensils, $t_o=25.97$\,s & 103.9 / 58.1 / 32.5\,s & $4/\sqrt5/1.25\,t_o$. H leaves 5.4\,s per utensil (two arms); one timed home load of 75 items (30 of them utensils) was dried and put away in 6\,min 45\,s, 5.4\,s per item~\cite{tlhousehold_cockeyed}; human shelf-stocking cycles take 1--4\,s per item~\cite{tlhousehold_roy}. \\
\addlinespace[2pt]
All & Release/settle dwells (jar and spoon in A1, glass, kettle, objects) & 1\,s & Half the ``2 seconds typical'' a bench scale needs to settle on a placed load~\cite{tlhousehold_adamcbc}; ours confirms the object rests. Analogy only. \\
\bottomrule
\end{tabularx}
\end{table*}

\begin{table*}[p]
\centering
\footnotesize
\renewcommand{\arraystretch}{1.1}
\caption{Time-parameter provenance: Industry. (a) Tasks whose window is set by a physical parameter of the scene (Tab.~\ref{tab:timing-process}); their values are not sourced. (b) Tasks with a time limit (Tab.~\ref{tab:timing-limits}) and Throughput tasks; each value is compared with a documented real-world time.}
\label{tab:timing-industry}
\begin{tabularx}{\textwidth}{@{}l p{3.2cm} p{2.6cm} Y@{}}
\toprule
\multicolumn{4}{@{}l}{\textit{(a) Window set by a physical parameter (Tab.~\ref{tab:timing-process})}} \\
Task & Parameter & Value (E/M/H) & Note \\
\midrule
A2\,(42) & Rotary accumulation table speed & 5$^\circ$/s (0.83\,rpm), all levels & A part takes 72\,s per turn; with the sector below it sets the loading window. \\
A2\,(42) & Loading-sector width & 60 / 50 / 40$^\circ$ & 12 / 10 / 8\,s to load the part before the sector passes. \\
A2\,(42) & Table start angle (seed-drawn) & $\pm$10$^\circ$ & Shifts the loading window by $\pm$2\,s so it cannot be anticipated. \\
\midrule
\multicolumn{4}{@{}l}{\textit{(b) Time limit or throughput tier (Tab.~\ref{tab:timing-limits})}} \\
Task & Parameter & Value (E/M/H) & Source and relation to our value \\
\midrule
A1\,(41) & Minimum zincate-dip immersion (\texttt{process\_min\_s}), from full immersion of the part picked from the loading pad & 18--22\,s (seed), all levels & Zincate dips last 20--120\,s~\cite{tlindustry_rudy2024}, 20--50\,s for the first dip of a double-zincate cycle~\cite{tlindustry_us5182006}; our 18--20\,s draws are up to 10\% shorter. \\
A1\,(41) & Exit-window width after the minimum & 28 / 16.85 / 10.14\,s & The patent's first-dip range of 20--50\,s spans 30\,s~\cite{tlindustry_us5182006}; easy (28\,s) is 0.93$\times$ that span and closes at most 50\,s after immersion, hard is a third of it. Oracle exit 22.10--25.64\,s after immersion. \\
A3\,(43) & Seating window opens after bead end & 2\,s & Corrugated-packaging hot melts are specified with open times in classes of 1--3, 3--5 and 5--15\,s, the time within which the flaps are mated~\cite{tlindustry_bostik2017}; our earliest seating at 2\,s lies inside the 1--3\,s class. Analogy only: it gives the dispenser time to clear. \\
A3\,(43) & Seating window closes after bead end & 40 / 25.67 / 16.47\,s & Easy is the longest documented hot-melt open time, 40\,s~\cite{tlindustry_3m3764} (shortest ${<}$15\,s~\cite{tlindustry_e2190}); data-sheet open time runs from application, ours from bead end. Oracle $t{=}13.17$\,s. \\
B1\,(44) & Coolant leak flow & 0.14\,L/s & Matches a 1/8\,in hole at 60\,psi (2.28\,gpm $=$ 0.144\,L/s)~\cite{tlindustry_bdvwa}. \\
B1\,(44) & Spill-tray limit / capacity & 2.0 / 2.6\,L & Smallest commercial tray opened holds 11.5\,L~\cite{tlindustry_kingfisher}, 4.4$\times$ ours. The limit is reached ${\approx}$14\,s after onset, which caps the easy deadline. \\
B1\,(44) & Valve fully closed after leak onset & 14 / 9.38 / 6.28\,s & Oracle $t{=}5.02$\,s; easy is capped by the tray limit ($4t$ would be 20.1\,s). Actuated quarter-turn valves stroke in 0.5--6\,s~\cite{tlindustry_omega,tlindustry_gfea}; ours is 1.05--2.3$\times$ the 6\,s maximum and adds detection and reach. LOPA credits a manual response only if $\ge$20\,min are available~\cite{tlindustry_bridges2010}. \\
B2\,(45) & Stop button actuated after warning & 3 / 2.75 / 2.53\,s & Measured e-stop presses: means 0.2--0.9\,s, hand near the button~\cite{tlindustry_fukaya1993}; ours is 2.8--3.3$\times$ the slowest mean (0.91\,s). Oracle $t{=}2.02$\,s. \\
B3\,(46) & Outfeed belt speed & 0.35\,m/s (0.33--0.37) & Standard unit-handling conveyor speed 65\,ft/min $=$ 0.33\,m/s~\cite{tlindustry_ciscoeagle}; nominal 6\,\% faster. \\
B3\,(46) & Slope of the gravity lane through the 0.3\,m safety zone & 8.2$^\circ$ & An un-braked bearing crosses the zone in 0.74--0.79\,s, so the brake must be in the lane by then at every level. A gravity roller conveyor is specified with a 7--10$^\circ$ decline for products under 5\,lb~\cite{tlindustry_dornergravity}; ours lies inside. \\
B3\,(46) & Stable arrest after zone entry & 4.77 / 2.67 / 1.50\,s & A conveyor drum-motor brake closes 13--500\,ms after power is cut~\cite{tlindustry_interroll2025}; ours is 3--9.5$\times$ the slowest closing and includes detection, reach and settling (oracle $t{=}1.19$\,s; easy $4t$). Analogy only. \\
A3, B1--B3 & Event onsets (seed-drawn) & bead end 16--20\,s; leak 20--24\,s; warning 14--18\,s; gate 7--9\,s & A variable foreperiod prevents anticipation: with a fixed one, a simple response can be fully prepared in advance~\cite{tlindustry_bausenhart2026}. Our onsets vary over 2--4\,s. \\
C1\,(47) & Two maintenance kits (6 parts) & $t_o{=}64.50$\,s; 258.0 / 144.2 / 80.6\,s & Hard allows 13.4\,s per part (two arms), 2.6$\times$ the 5.06--5.16\,s per component modelled for manual and cobot-supported kit preparation~\cite{tlindustry_fager2020}; human warehouse pickers take 5--10\,s per item~\cite{tlindustry_correll2018}. \\
C2\,(48) & Three cartons shelved & $t_o{=}45.32$\,s; 181.3 / 101.3 / 56.6\,s & 18.9 / 33.8 / 60.4\,s per carton, inside the 6.9--124\,s per case of manual full-case picking (525 to 29 cases per man-hour); man-up order picking takes 26.9\,s (134 cases/h)~\cite{tlindustry_stone2014}. \\
C3\,(49) & Three parts weighed and sorted & $t_o{=}69.36$\,s; 277.4 / 155.1 / 86.7\,s & Hard allows 28.9\,s per part, 4.7--4.8$\times$ a kitting move (5.06--5.16\,s per component~\cite{tlindustry_fager2020}) plus a 1\,s scale stabilisation~\cite{tlindustry_ohausranger3000}. Analogy only: no weigh-and-sort cycle time found. \\
C3\,(49) & Part at rest on the pan before a reading counts; display settling & 0.5\,s; within 2\,g over 0.4\,s & Checkweighers should stabilise in under 0.5\,s, one model in 0.43\,s~\cite{tlindustry_awt2022}; a bench scale states 1\,s~\cite{tlindustry_ohausranger3000}. Our 0.4\,s settling matches the former, rest plus settling (0.9\,s) the latter. Underweight parts weigh 70\,\% of nominal (105 vs.\ 150\,g, PASS at $\ge$125\,g). \\
C4\,(50) & Two four-pan palettes & $t_o{=}81.46$\,s; 325.8 / 182.2 / 101.8\,s & Hard allows 10.2\,s per placement (8 pans, 2 cases), 2.0$\times$ the 5.06--5.16\,s per kitted component~\cite{tlindustry_fager2020} and at the upper end of 5--10\,s per picked item~\cite{tlindustry_correll2018}; analogy only. \\
\bottomrule
\end{tabularx}
\end{table*}

\begin{table*}[p]
\centering
\footnotesize
\setlength{\tabcolsep}{4pt}
\renewcommand{\arraystretch}{1.08}
\caption{Time-parameter provenance: Sports \& games. (a) Tasks whose window is set by a physical parameter of the scene (Tab.~\ref{tab:timing-process}); their values are not sourced. (b) Tasks with a time limit (Tab.~\ref{tab:timing-limits}) and Throughput tasks; each value is compared with a documented real-world time.}
\label{tab:timing-sports}
\begin{tabularx}{\textwidth}{@{}l p{3.2cm} p{2.6cm} Y@{}}
\toprule
\multicolumn{4}{@{}l}{\textit{(a) Window set by a physical parameter (Tab.~\ref{tab:timing-process})}} \\
Task & Parameter & Value (E/M/H) & Note \\
\midrule
B4 & RED to confirmed rest of hands, tools and puck & 8 / 4 / 2\,s & Time allowed to come to rest in each red phase; oracle $t\le0.136$\,s. Each red lasts the limit plus a seeded 2.2--3.2\,s hold. \\
B5 & Note window, symmetric about the note & $\pm$2.0 / 1.41 / 1.0\,s & Hard is the 1\,s floor. Oracle strikes 0.104\,s early: hard margin 0.896\,s. \\
B5 & Note interval; first note (8 notes) & 5\,s; 12\,s & Wide enough that windows never overlap. \\
\midrule
\multicolumn{4}{@{}l}{\textit{(b) Time limit or throughput tier (Tab.~\ref{tab:timing-limits})}} \\
Task & Parameter & Value (E/M/H) & Source and relation to our value \\
\midrule
A1 & Clock-press deadline after the opponent's press & 30 / 17 / 10\,s & Oracle $t=7.90$\,s. Easy equals the common 30\,s Fischer increment~\cite{tlsports_fide2023}; hard equals the 10\,s increment of the World Rapid Championship and is 5$\times$ the 2\,s blitz increment~\cite{tlsports_fide2025}. An increment is added thinking time, not a deadline. \\
A2 & Shot clock, READY to release & 14 / 9 / 6\,s & Oracle $t=4.59$\,s, grasp included. Easy equals the FIBA 14\,s shot-clock reset~\cite{tlsports_fiba2026}; FIBA's 5\,s free-throw limit (ball already in hand) is below our hard. \\
A2 & Flight grace after release & 4\,s & Successful free throws of 20 league players took 1.11\,s (0.91--1.55\,s) from release to the basket~\cite{tlsports_kelmendi2025}; ours is 3.6$\times$ the mean. Our 0.3\,m-apex lob flies ${\approx}0.5$\,s (ballistics, our estimate). \\
A2 & READY onset after the start & U[2,\,6)\,s & Randomized so the onset cannot be anticipated: with a variable foreperiod, the moment of the signal is uncertain, whereas a fixed one can be anticipated~\cite{tlsports_han2022,tlhospital_forgaard2016}. \\
B1 & Mole time-up, from its rise & 4.0 / 3.0 / 2.2\,s & Oracle $t=1.756$\,s. Easy is capped at the 4\,s between moles ($4t$ would be 7.0\,s). Lab whack-a-mole tasks show targets for 1.8\,s~\cite{tlsports_inquisit} and 1.25\,s~\cite{tlsports_naber2013}; ours are 1.2--3.2$\times$ longer. Arcade manual states no unit~\cite{tlsports_bsrfec}. \\
B1 & Mole spacing; moles per episode & 4.0\,s; 9 (${\approx}1/3$ decoys) & Between the 2.3\,s trial of the Inquisit go/no-go task and the 1.5\,s stimulus plus 5\,s interval of the study it reproduces~\cite{tlsports_inquisit}. \\
B2 & GO cue to pad hit & 1.6 / 1.25 / 1.0\,s & Oracle $t=0.680$\,s. Easy is capped by the cue schedule: the next cue of a combination follows 2.8--3.5\,s later, so hit and return share about 3\,s. Boxers' punch delivery takes 0.36--0.66\,s, cue reaction excluded~\cite{tlsports_piorkowski2011,tlsports_finlay2023}; hard is 1.5--2.8$\times$ that. \\
B2 & Hit to glove back in guard & 1.4 / 1.2 / 1.05\,s & Oracle $t=0.716$\,s. In four-punch combinations, boxers went from the first to the fourth contact in 1.04--1.35\,s, about 0.35--0.45\,s per punch-and-recovery cycle~\cite{tlsports_piorkowski2011}; our hard is 2.3--3$\times$ that. Analogy only: no isolated return-to-guard time found. \\
B3 & Catch deadline, from bar descent & 2.05 / 1.43 / 1.0\,s & Oracle $t=0.512$\,s; easy $4t$; hard is the 1\,s floor. The fatigued lifter lowers the bar at 5\,cm/s, so it stays above the danger height for 3.2--5.3\,s. IPF rules let spotters take a failing bar but set no time~\cite{tlsports_ipf2026}. Hands-in-place catch reactions take 0.18--0.25\,s~\cite{tlsports_eckner2011}; our hard is 4--5.6$\times$ that. \\
B3 & Lifter stall before the drop & 0.65--1.1\,s (seeded) & Brackets the ${\approx}0.8$\,s bench-press sticking region quoted in~\cite{tlsports_vandentillaar2014}; that is a slowdown at near-maximal load, not a stall. \\
C1 & Time tier to rack 15 balls & 510.0 / 285.1 / 159.4\,s & $t_o=127.49$\,s; tiers 4, $\sqrt5$, 1.25$\times t_o$. Adults 40 and older move 61--83 blocks a minute one at a time in the Box and Block Test~\cite{tlsports_sralab}, so 15 transfers take 11--15\,s; hard allows 11--15$\times$ that. Analogy only: no racking time found. \\
C2 & Time tier for 2 cards, after 2\,s preparation & 125.6 / 70.2 / 39.3\,s & $t_o=31.41$\,s. Speed Cups is an untimed race~\cite{tlsports_amigo}; a 12-cup sport-stacking record cycle takes 4.7\,s~\cite{tlsports_guinness}, while hard allows 19.6\,s per card of 6 cups. At Box and Block rates~\cite{tlsports_sralab} 6 transfers take 4.3--5.9\,s. \\
C3 & Ball demands: first demand; firing interval (6 demands) & 21.65 / 13.20 / 8.05\,s; 12.00 / 7.32 / 4.46\,s & Reference 6.44\,s $+$ 3.57\,s per demand, $\times$3.361 / 2.050 / 1.25. Easy is capped at a ball machine's slowest feed, 12\,s (range 2--12\,s)~\cite{tlsports_lobster}. \\
\bottomrule
\end{tabularx}
\end{table*}

\FloatBarrier
\section{Temporal Difficulty Study}
\label{app:emh}
We ran Gemini~4 Argon through Tool-augmented EEF on all 60 tasks at the medium and hard levels, on seeds 0--9 for every task (600 episodes per level; rates are averaged within each task first). PSR and TSR are measured with inference paused, and LTSR in Latency-Aligned mode with the measured API latency and $\delta=20$\,ms; the easy row repeats Tab.~\ref{tab:api-latest} (Tab.~\ref{tab:emh-full}). Because the levels tighten only the temporal parameters, PSR stays between 22.5 and 28.0\%, while overall TSR falls from 18.2\% at easy and 18.8\% at medium to 11.3\% at hard, mostly in Temporal Window (20.5 to 16.8\%) and Event-Reactive tasks (24.6 to 14.2\%). LTSR falls from 8.2\% at easy to 3.3\% and 1.5\%. The API serving speed may differ between the easy run and the later medium and hard runs, so the LTSR comparison across levels is indicative only.

\begin{table*}[t]
\centering\small
\caption{Gemini~4 Argon through Tool-augmented EEF on all 60 tasks at the three difficulty levels (10 seeds per task, 0--9). PSR and TSR with inference paused; LTSR with measured API latency. Easy: Tab.~\ref{tab:api-latest}. A/B/C: Temporal Window, Event-Reactive, Throughput; metrics in \%, macro-averaged over tasks; Overall is weighted by the 19/24/17 tasks of A/B/C.}
\label{tab:emh-full}
\setlength{\tabcolsep}{5pt}
\begin{tabular}{@{}l*{12}{c}@{}}
\toprule
& \multicolumn{3}{c}{\cellcolor{catA}A} & \multicolumn{3}{c}{\cellcolor{catB}B} & \multicolumn{3}{c}{\cellcolor{catC}C} & \multicolumn{3}{c}{\cellcolor{catAll}Overall} \\
\cmidrule(lr){2-4}\cmidrule(lr){5-7}\cmidrule(lr){8-10}\cmidrule(l){11-13}
Difficulty & PSR & TSR & LTSR & PSR & TSR & LTSR & PSR & TSR & LTSR & PSR & TSR & LTSR \\
\midrule
Easy   & 24.7 & 20.5 & 10.5 & 36.3 & 24.6 & 12.1 & 7.6  & 6.5 & 0.0 & 24.5 & 18.2 & 8.2 \\
Medium & 34.7 & 27.4 & 6.3  & 40.0 & 24.2 & 3.3  & 3.5  & 1.8 & 0.0 & 28.0 & 18.8 & 3.3 \\
Hard   & 27.4 & 16.8 & 1.1  & 30.0 & 14.2 & 2.9  & 6.5  & 1.2 & 0.0 & 22.5 & 11.3 & 1.5 \\
\bottomrule
\end{tabular}
\end{table*}

\section{Agent Behavior, Cost, and Per-Task Variation}
\label{app:agent-behavior}
Tab.~\ref{tab:agent-behavior} reports interaction costs for a six-model diagnostic subset under zero latency: model turns per episode, output tokens per turn, and API cost per episode.
\begin{table}[t]
\centering\footnotesize
\caption{\textbf{Interaction cost of VLM agents} (six-model diagnostic subset under zero latency, 10 seeds per task). Turns: model calls per episode. Tok: output tokens per turn, reasoning included. \$: API cost per episode (USD). $^\dagger$open-weight; $^\S$text-only.}
\label{tab:agent-behavior}
\setlength{\tabcolsep}{6pt}
\begin{tabular}{@{}lccc@{}}
\toprule
Model & Turns & Tok & \$ \\
\midrule
Gemini 4 Argon & 16.1 & 681 & 0.35 \\
Claude Opus 5.5 & 23.0 & 770 & 0.57 \\
GPT-6 Astra & 51.2 & 160 & 1.58 \\
GPT-6.1 Sol & 43.0 & 163 & 0.24 \\
DeepSeek V4.1 Flash$^\dagger$ & 66.3 & 2821 & 0.16 \\
GLM-5.3$^{\dagger\S}$ & 28.7 & 416 & 0.13 \\
\bottomrule
\end{tabular}
\end{table}

\subsection{Effect of API Serving Speed on LTSR}
\label{app:api-speed}
Latency-Aligned evaluation charges each query's round trip, so the LTSR of an API-served agent depends on the serving speed. Table~\ref{tab:api-speed} compares measured serving rates of 18 and 26 tokens/s for GPT-6 Astra (Microsoft Azure), and 22 and 36 tokens/s for Gemini~4 Argon (Google Cloud).

\begin{table}[t]
\centering\small
\caption{LTSR (\%) of API-served VLM agents at different measured serving rates. Each model appears at two decoding rates; tok/s: average decoding rate of the API during the evaluation. GPT-6 Astra is served through Microsoft Azure and Gemini~4 Argon through Google Cloud. Overall is weighted by the 19/24/17 tasks of categories A/B/C.}
\label{tab:api-speed}
\setlength{\tabcolsep}{4pt}
\begin{tabular}{@{}lccccc@{}}
\toprule
Model & tok/s & A & B & C & Overall \\
\midrule
\multirow{2}{*}{GPT-6 Astra} & 18 & 0.0 & 2.1 & 0.0 & 0.8 \\
 & 26 & 5.8 & 2.9 & 0.0 & 3.0 \\
\addlinespace[2pt]
\multirow{2}{*}{Gemini~4 Argon} & 22 & 0.0 & 2.5 & 0.0 & 1.0 \\
 & 36 & 10.5 & 12.1 & 0.0 & 8.2 \\
\bottomrule
\end{tabular}
\end{table}

Faster serving raises LTSR from 0.8\% to 3.0\% and from 1.0\% to 8.2\%, through tasks whose windows leave more than one model turn; tasks that must be answered within a few seconds stay at zero. Gemini gains more because it needs fewer turns (a median of 3 against 6 per successful episode). LTSR of API-served agents should therefore be reported with the provider and serving speed; Tab.~\ref{tab:agent-results} reports the faster serving.

\paragraph{Results with faster API serving.}
Tab.~\ref{tab:api-latest} repeats the rows of Tab.~\ref{tab:agent-results} for the four API-served models, which are the simulated values used in the comparison with the real robot in App.~\ref{app:real-robot}.

\begin{table*}[t]
\centering\small
\caption{Results of four API-served VLM agents through Tool-augmented EEF, with faster API serving and latency charged in closed loop (easy level). A/B/C: Temporal Window, Event-Reactive, Throughput; metrics in \%, macro-averaged over tasks; all values are those of Tab.~\ref{tab:agent-results}, and Overall is weighted by the 19/24/17 tasks of A/B/C. \textbf{Bold}: best.}
\label{tab:api-latest}
\setlength{\tabcolsep}{4pt}
\begin{tabular}{@{}l*{16}{c}@{}}
\toprule
& \multicolumn{4}{c}{\cellcolor{catA}A} & \multicolumn{4}{c}{\cellcolor{catB}B} & \multicolumn{4}{c}{\cellcolor{catC}C} & \multicolumn{4}{c}{\cellcolor{catAll}Overall} \\
\cmidrule(lr){2-5}\cmidrule(lr){6-9}\cmidrule(lr){10-13}\cmidrule(l){14-17}
Model & PS & PSR & TSR & LTSR & PS & PSR & TSR & LTSR & PS & PSR & TSR & LTSR & PS & PSR & TSR & \textbf{LTSR} \\
\midrule
Gemini 4 Argon & 41.1 & 24.7 & 20.5 & \textbf{10.5} & 40.4 & \textbf{36.3} & 24.6 & \textbf{12.1} & \textbf{21.8} & \textbf{7.6} & \textbf{6.5} & 0.0 & 35.3 & \textbf{24.5} & 18.2 & \textbf{8.2} \\
GPT-6 Astra & \textbf{43.7} & \textbf{28.4} & 22.1 & 5.8 & \textbf{40.8} & 34.6 & \textbf{30.0} & 2.9 & 18.8 & 0.0 & 0.0 & 0.0 & \textbf{35.5} & 22.8 & \textbf{19.0} & 3.0 \\
GPT-6.1 Sol & 38.9 & 27.4 & \textbf{24.2} & 1.6 & 37.1 & 32.9 & 24.2 & 4.6 & 15.3 & 0.0 & 0.0 & 0.0 & 31.5 & 21.8 & 17.3 & 2.3 \\
Claude Opus 5.5 & 30.5 & 17.4 & 12.6 & 0.0 & 30.8 & 26.7 & 23.8 & 2.5 & 12.9 & 1.2 & 0.6 & 0.0 & 25.7 & 16.5 & 13.7 & 1.0 \\
\bottomrule
\end{tabular}
\end{table*}

\paragraph{Decode-speed model behind Fig.~\ref{fig:ablation-tps}.}
The ablation of Sec.~\ref{sec:ablation-tps} replaces the measured latency of Eq.~\eqref{eq:latency} with one derived from an assumed decoding rate:
\begin{equation}
  \hat{L}_{\mathrm{infer},k}
  = \frac{N_{\mathrm{out},k}}{r_{\mathrm{dec}}},
  \label{eq:latency-decode}
\end{equation}
where $N_{\mathrm{out},k}$ is the number of output tokens of query $k$, reasoning included, and $r_{\mathrm{dec}}$ the assumed rate (35 to 3K tokens/s). It omits prefill, network, and queueing time and keeps token counts fixed, so Fig.~\ref{fig:ablation-tps} shows the effect of decoding speed, not the LTSR of a particular deployment.

\section{Task Manifest and Event Semantics}
\label{app:manifest}
Each task is specified by a manifest that fixes the scene, the robot, the observations, the temporal contract, and the physical goal. Every timed event is a predicate on physical state with fixed thresholds and is logged with its environment timestamp; temporal clauses are evaluated on these timestamps. Each window is anchored at its specified measured event: policy-triggered intervals move with the corresponding action, whereas external-event deadlines remain tied to the external trigger.

\section{Temporal Monitor Acceptance Checks}
\label{app:monitor}
Before model evaluation, the temporal monitors are checked on reference and deliberately invalid traces: an action inside, before, or after its window, a missing or unanswered event, an interrupted contact, a unit counted twice, and a rolling interval below the required rate must each yield the intended outcome.

\section{Study Matrix}
\label{app:studies}
Physical versus temporal success and the temporal taxonomy are reported in Tabs.~\ref{tab:main-results} and~\ref{tab:agent-results}; fine-tuning and transfer in Tabs.~\ref{tab:main-results} and~\ref{tab:seen-unseen}; the action interface in Tab.~\ref{tab:interface-results}; inference speed and latency across policy families in Secs.~\ref{sec:ablation-tps} and~\ref{sec:ablation-latency}; and agent behavior in App.~\ref{app:agent-behavior}.

\section{Dataset and Runtime Reporting}
\label{app:reporting}
Each episode record stores the task, level, seed, instruction, timestamped events, score stages, violations, and termination reason, and per query the environment time, the measured and charged latency, and the output tokens. All methods use the same available sensor rig and robot-state streams, with model-specific input selection and preprocessing (Tab.~\ref{tab:finetune} for the fine-tuned policies); the text-only GLM-5.3 reference receives no images. VLM agents read the task clock as text, fine-tuned policies receive it in the instruction, and only the oracle reads ground-truth state.

\section{Reproducibility}
\label{app:reproducibility}
We will release the task manifests, assets, splits, seeds, monitors, oracles, demonstrations, baseline adapters, prompts, run configurations, and aggregation code. Healthcare and laboratory parameters are not operational safety limits.

\FloatBarrier
\section{Real-Robot Pilot}
\label{app:real-robot}
This pilot carries the three temporal categories to a physical arm with six tasks (Tab.~\ref{tab:real-tasks}), scores six VLMs on them (App.~\ref{app:real-results}), and compares the resulting ranking with simulation.

\paragraph{Hardware.}
One UFactory xArm~6 with a soft-finger gripper stands at a table. A RealSense D435 head camera across the table gives a $640\times480$ color image and a wrist camera an infrared image, both calibrated to the arm's base frame. Unlike the simulated rig, there is a single arm and no depth image.

\paragraph{Control loop and interface.}
The model and the arm exchange one observation and one command per turn. The arm side enforces inverse kinematics, joint and workspace limits, a table guard, and a watchdog; the tool moves at 30\,mm/s, and an operator holds the emergency stop. An episode ends on \texttt{finish}, at the turn cap or 600\,s, on a controller error, or on operator interruption.

The single-arm Tool-augmented EEF interface (App.~\ref{app:tool-eef}) has ten tools and one call per turn; a target is a base-frame point or a head-image pixel with a height above the table, and tool results are measurements only. Direct pose commands were used in all C2 episodes and most A2 and B2 episodes; the remaining episodes used the tool interface.

There is no paused clock on hardware, so success corresponds to LTSR. The median model reply takes 5.1\,s, from 3.4\,s for Qwen3.8~27B to 16.8\,s for DeepSeek~V4.1~Flash.

Six VLMs were run: GPT-6~Astra, GPT-6.1~Sol, Gemini~4~Argon, Claude~Opus~5.5, DeepSeek~V4.1~Flash, and Qwen3.8~27B. Logs store every model input and command, the robot state per turn, call times and costs, and 10\,fps videos.

\begin{table*}[t]
\centering\scriptsize
\setlength{\tabcolsep}{3pt}
\caption{The six real-robot tasks, grouped as in simulation. A: Temporal Window; B: Event-Reactive; C: Throughput. Top row: the scene, photographed from the side; in the B2 and C2 photographs the head camera is the one on the boom above the arm, and the board of fiducial markers at the left edge is used only for camera calibration. Second row: the first head-camera image of an episode, as the model sees it, with the arm's base at the top. Third row: what the agent must do. Last two rows: the time requirement, which is $4\times$ the human standard time for A2, B1, C1, and C2 and the ringing duration of the alarm for B2; ``--'': none applies. $^{\dagger}$A1 encodes its temporal stopping condition through the physical volume limit: pouring must stop before the amount exceeds 100\,mL; no fixed duration is prescribed.}
\label{tab:real-tasks}
\newcommand{\rc}[1]{\multicolumn{1}{c}{#1}}
\newcommand{\rrimg}[1]{\includegraphics[width=\linewidth]{figures/real_#1.jpg}}
\begin{tabular}{@{}*{6}{p{0.153\linewidth}}@{}}
\toprule
\multicolumn{2}{c}{\textbf{A: Temporal Window}} & \multicolumn{2}{c}{\textbf{B: Event-Reactive}} & \multicolumn{2}{c}{\textbf{C: Throughput}} \\
\cmidrule(r){1-2}\cmidrule(lr){3-4}\cmidrule(l){5-6}
\rrimg{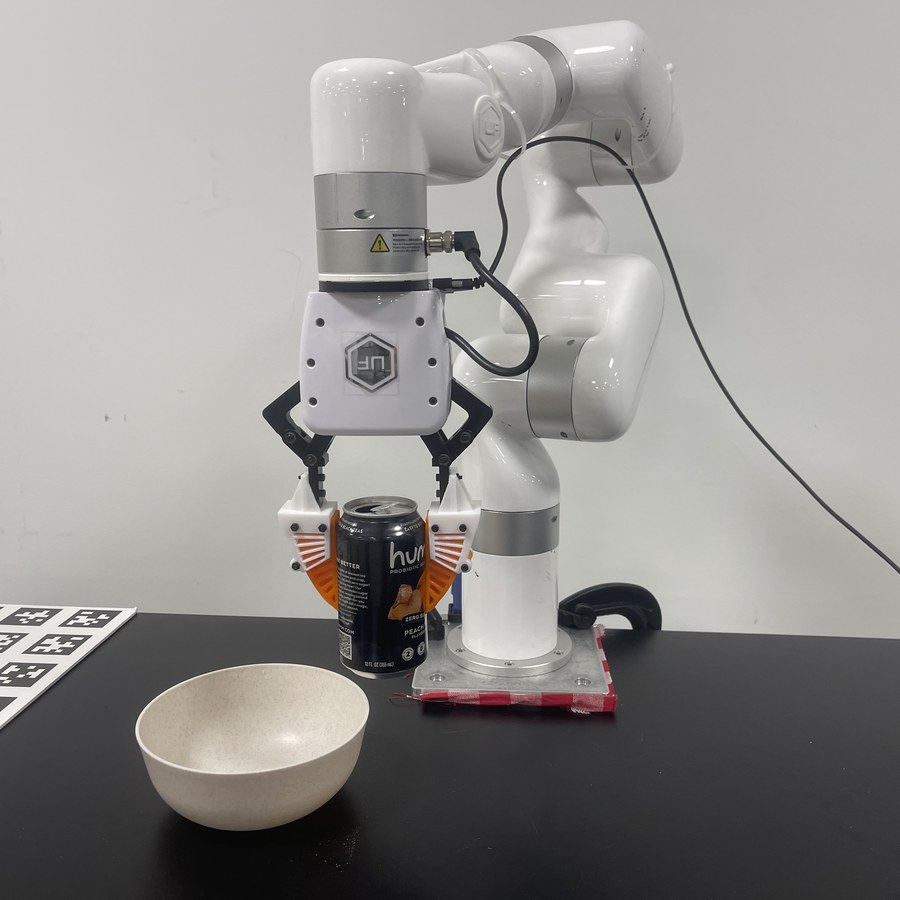} & \rrimg{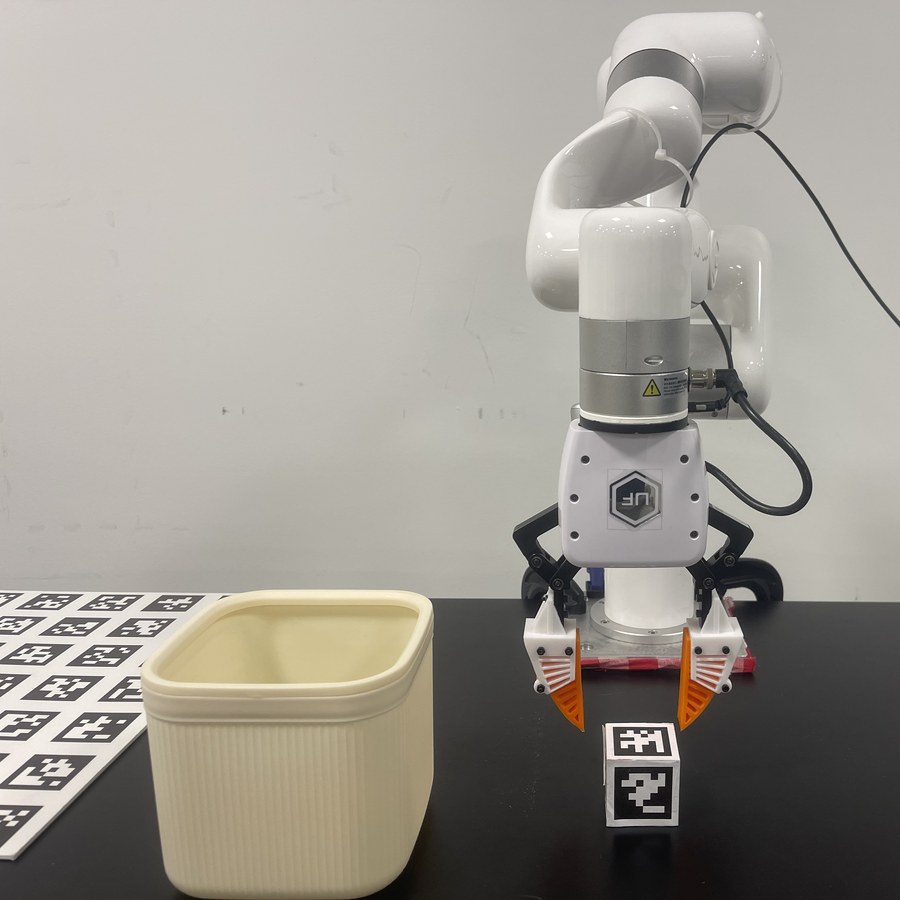} & \rrimg{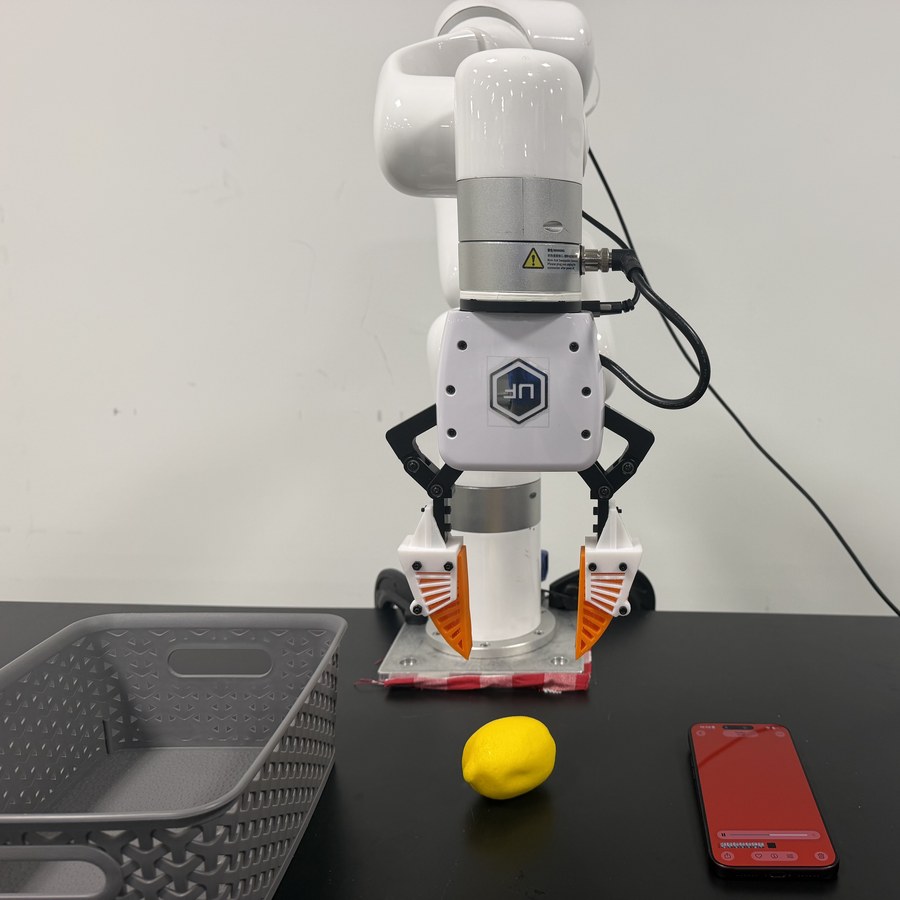} & \rrimg{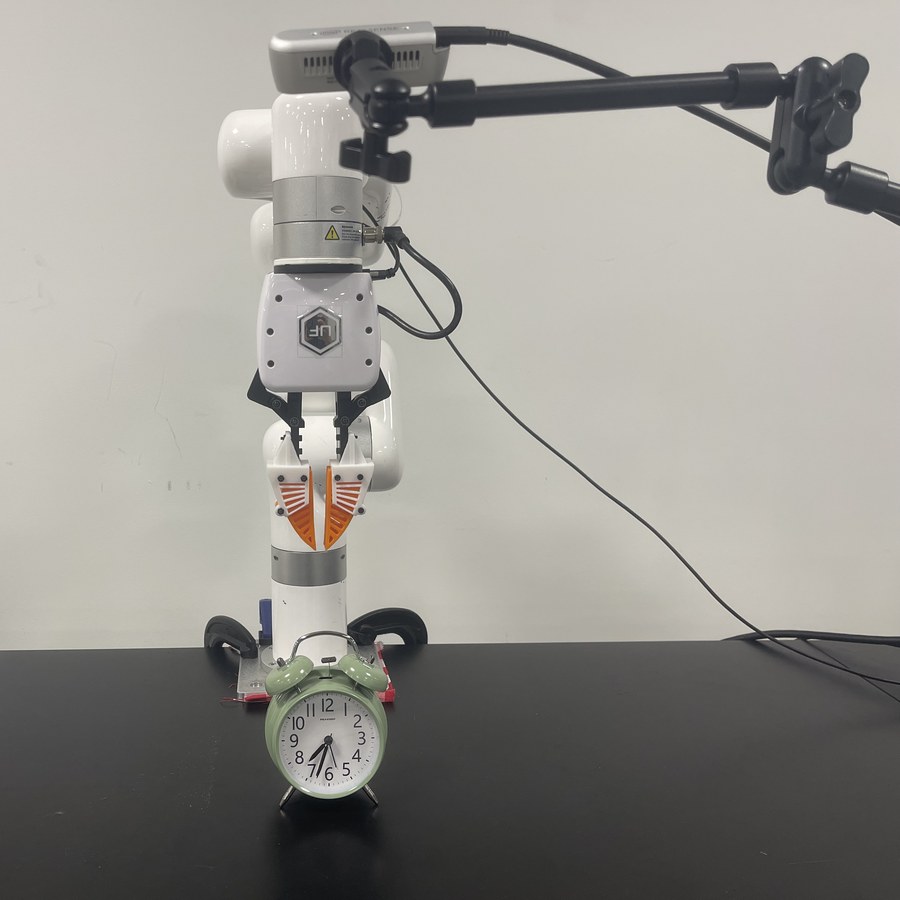} & \rrimg{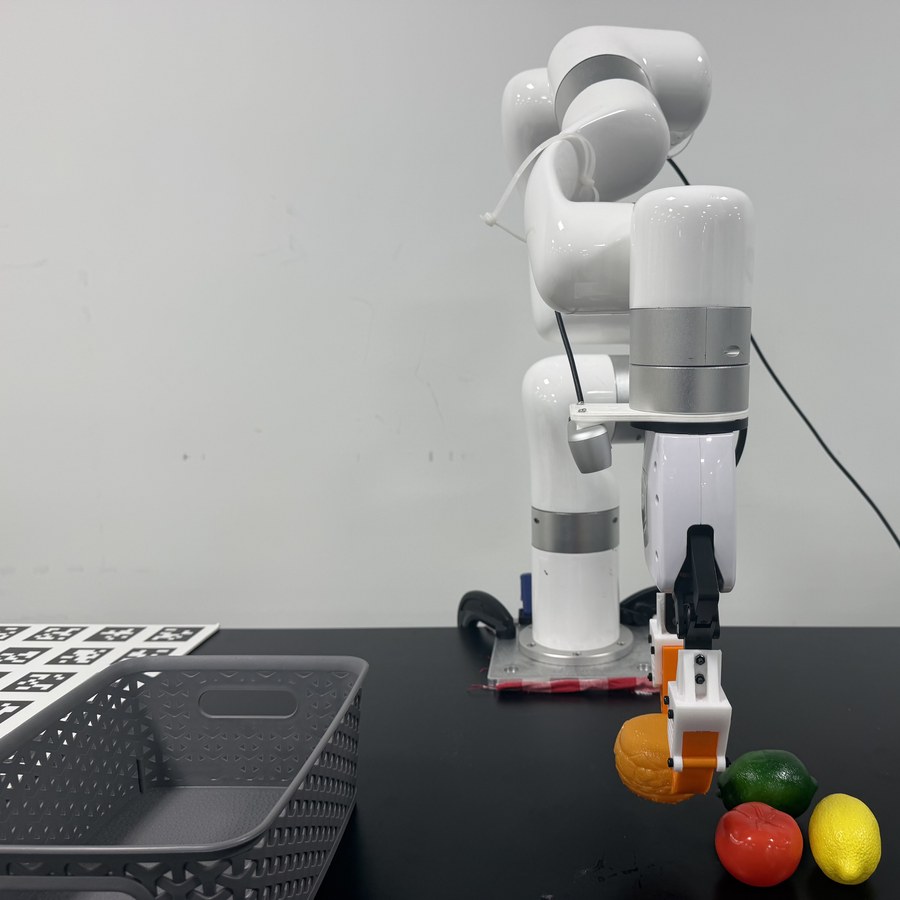} & \rrimg{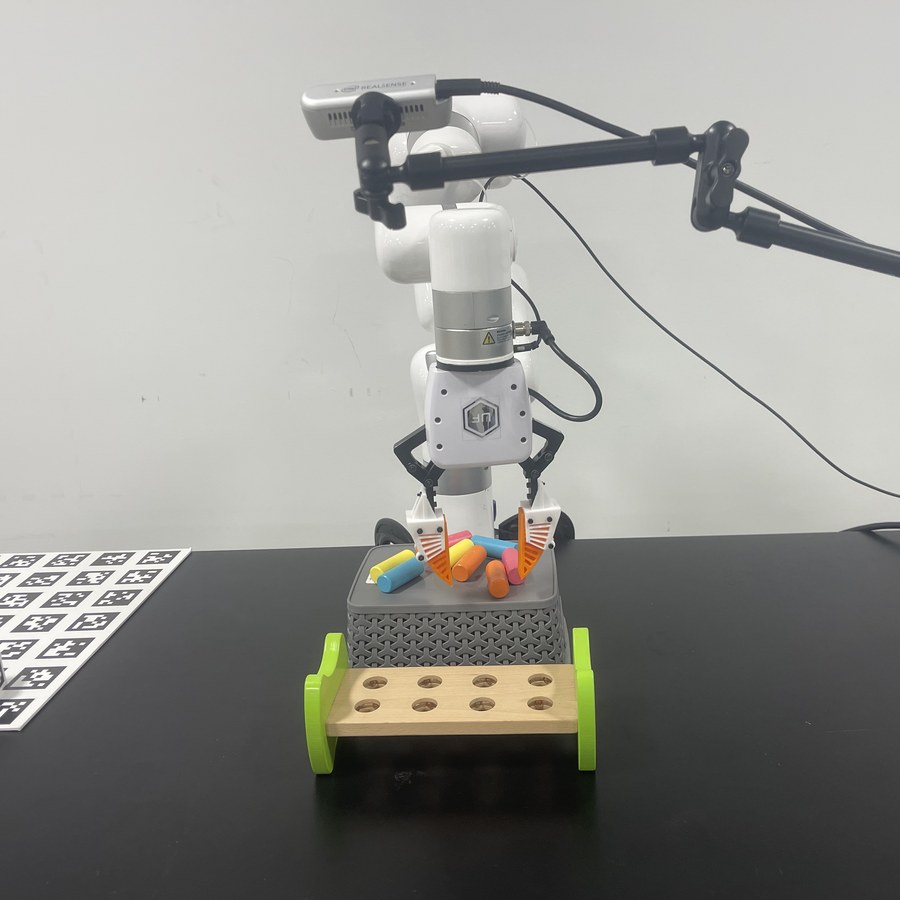} \\
\rrimg{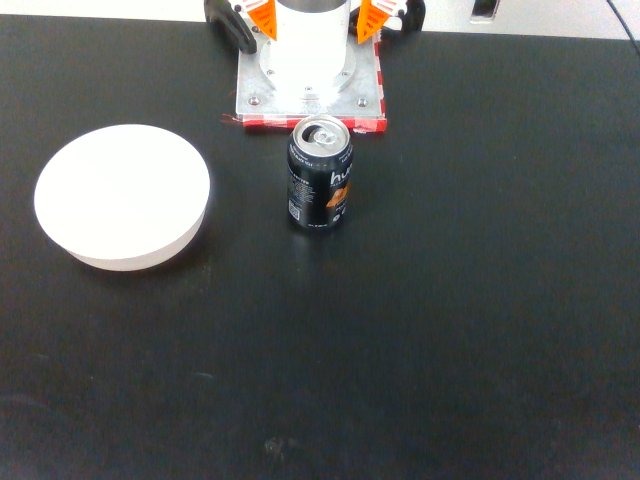} & \rrimg{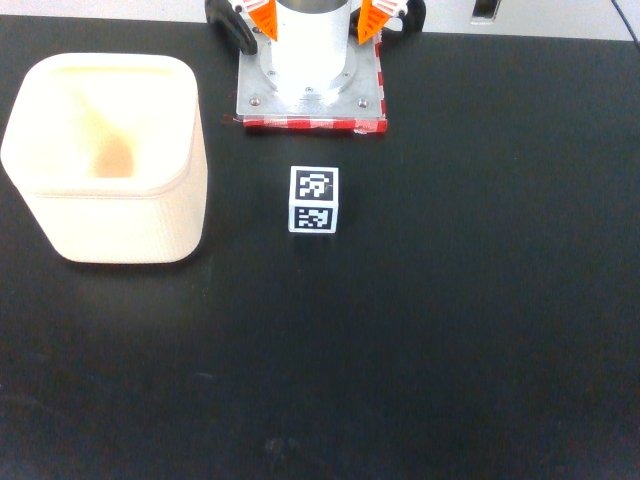} & \rrimg{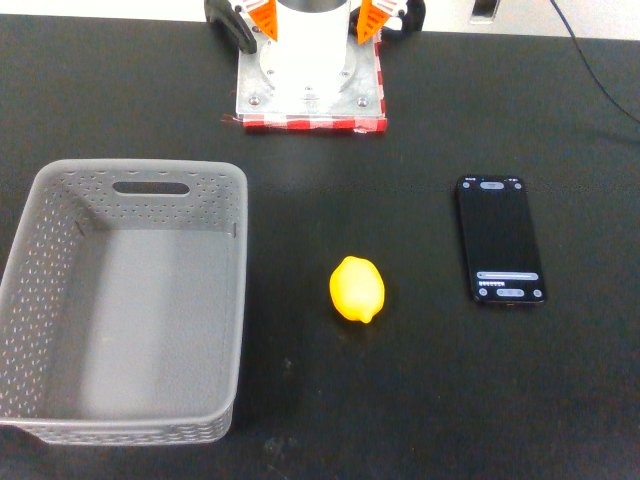} & \rrimg{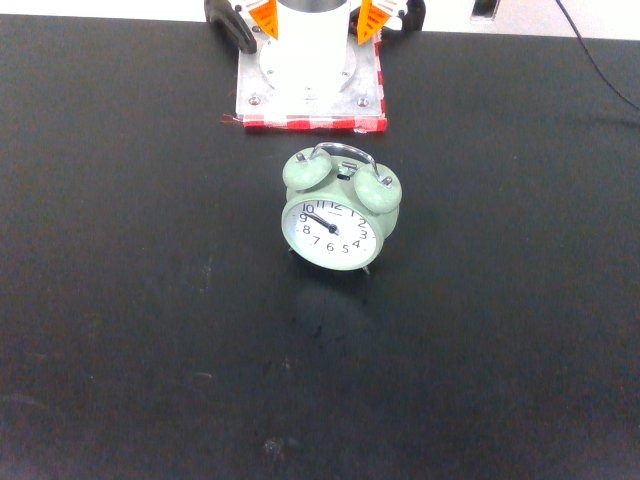} & \rrimg{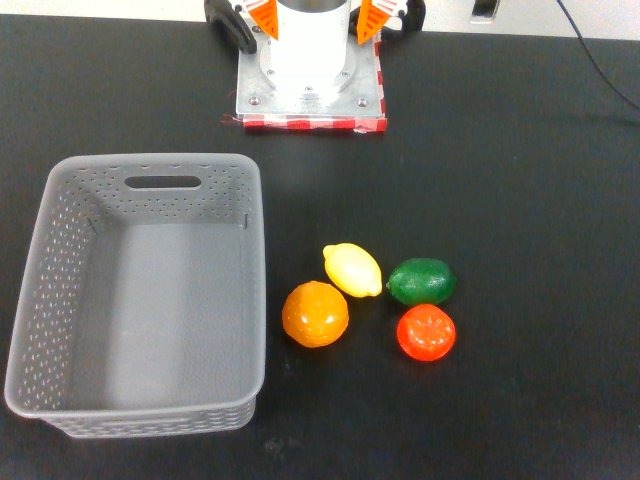} & \rrimg{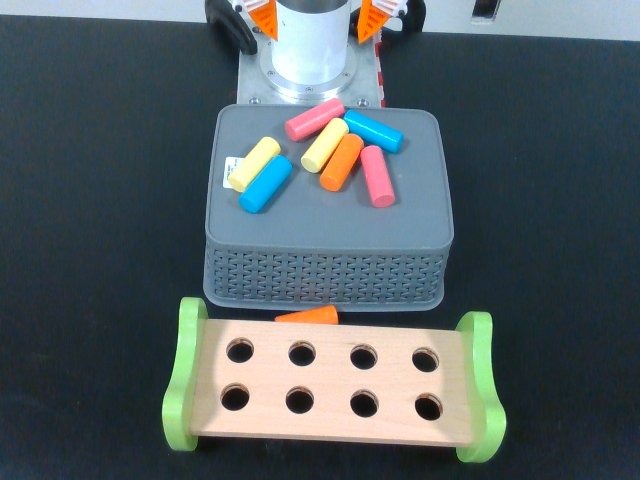} \\
\textbf{A1 Pour from a can.} Grasp the open drinks can, carry it over the white bowl, tilt it toward the bowl, set it upright, and put it back where it stood. &
\textbf{A2 Cube into box.} Grasp the 50\,mm tagged cube, lift it above the rim of the open storage box, and release it over the opening without touching the box. &
\textbf{B1 Lemon on red.} A phone on the table shows colors at unpredictable moments. When it turns red, put the toy lemon into the basket. Any motion before red fails; other colors are distractors. &
\textbf{B2 Alarm press.} When the twin-bell alarm clock rings, press its top between the two bells with the closed fingertips, then retreat. &
\textbf{C1 Fruits into basket.} Put the four toy fruits (about 5\,cm) into the basket with 10\,cm walls, as many and as fast as possible. &
\textbf{C2 Peg sorting.} Move eight pegs in four colors from a raised stand into the $2\times4$ pegboard so that each column holds one color. \\
\midrule
\multicolumn{6}{c}{\textbf{Temporal requirement}} \\
\rc{$\le100$\,mL$^{\dagger}$} & \rc{12.8\,s} & \rc{8.4\,s} & \rc{3\,min (alarm rings)} & \rc{21.6\,s} & \rc{72.8\,s} \\
\addlinespace[2pt]
\multicolumn{6}{c}{\textbf{Human standard time}} \\
\rc{--} & \rc{3.2\,s} & \rc{2.1\,s} & \rc{--} & \rc{5.4\,s} & \rc{18.2\,s} \\
\bottomrule
\end{tabular}
\end{table*}

\paragraph{Tasks.}
\emph{A1} pours from an open can into a bowl, stops before the poured amount exceeds 100\,mL, and returns the can to its starting position. \emph{A2} requires the cube to be released into the box within 12.8\,s. In \emph{B1} a phone screen shows colors at unpredictable moments; after red, and only then, the lemon must go into the basket within 8.4\,s. In \emph{B2} the agent presses the top of a ringing alarm clock within its 3\,min ring. In \emph{C1} four rolling toy fruits go into a basket (21.6\,s); in \emph{C2} eight pegs are sorted by color into a $2\times4$ pegboard (72.8\,s). A2, B1, C1, and C2 use four times the human standard time. B2 uses the alarm ringing duration. A1 implements its temporal stopping condition through the physical volume limit rather than a fixed duration.

\paragraph{Case study: B1 across six models.}
We ran B1 once per model and timed it from the red onset (Tab.~\ref{tab:real-b1-case}). Every model waited for red, but the first motion came 5.0--11.5\,s later, and the three models that delivered the lemon needed 45.8--78.1\,s, far beyond the simulated Event-Reactive limits. Claude~Opus~5.5 and Qwen3.8~27B grasped but did not deliver it, and DeepSeek~V4.1~Flash never grasped it. With one episode per model, these runs illustrate behavior and do not rank the models.

\begin{table*}[t]
\centering\footnotesize
\setlength{\tabcolsep}{2.5pt}
\caption{Case study on B1 (lemon on red), one episode per model. Start, Grasp, and Placed are seconds after the screen turned red, from the operator's record; ``--'' means the step was not reached. PS: 50 for holding the lemon, 100 for the lemon in the basket. Frames are from the head camera: 0.5\,s after the red onset, about 2\,s after the first motion, about 2\,s after the grasp (for DeepSeek~V4.1~Flash, which never grasped, 30\,s after the onset), and the end of the episode.}
\label{tab:real-b1-case}
\newcommand{\bcimg}[1]{\raisebox{-0.5\height}{\includegraphics[width=0.128\linewidth]{figures/b1case_#1.jpg}}}
\renewcommand{\arraystretch}{1.15}
\begin{tabular}{@{}l r r r r c c c c@{}}
\toprule
Model & Start (s) & Grasp (s) & Placed (s) & PS & Red onset & First motion & Grasp & End \\
\midrule
DeepSeek V4.1 Flash & 4.97 & -- & -- & 0 & \bcimg{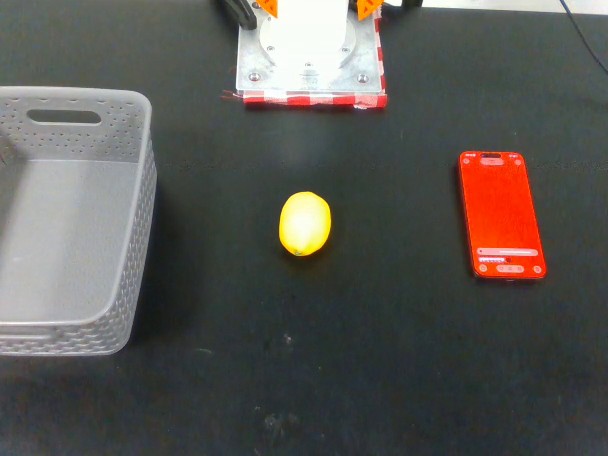} & \bcimg{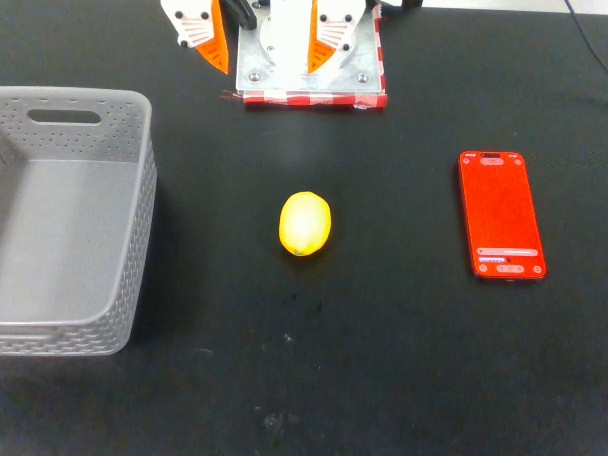} & \bcimg{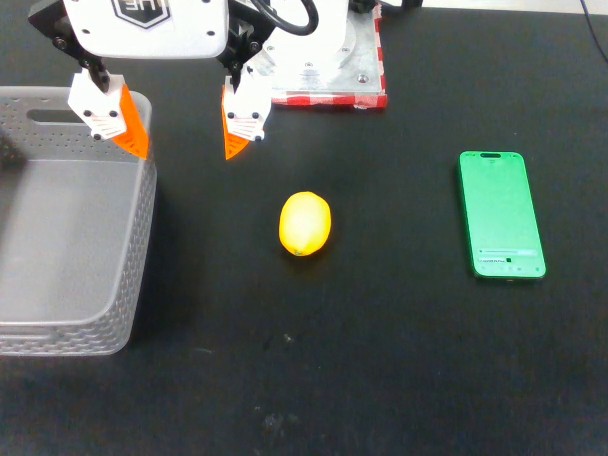} & \bcimg{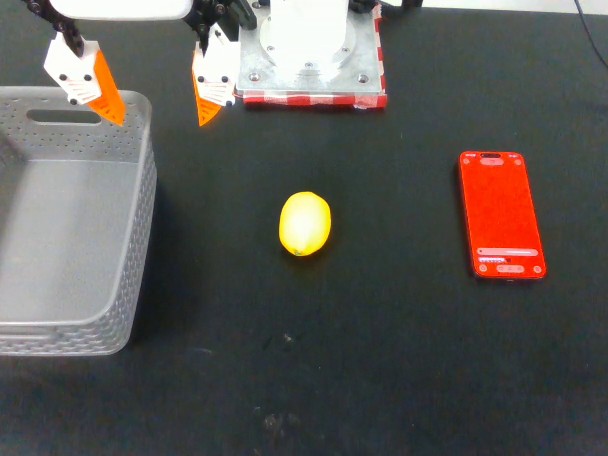} \\
GPT-6 Astra & 6.61 & 16.86 & 56.40 & 100 & \bcimg{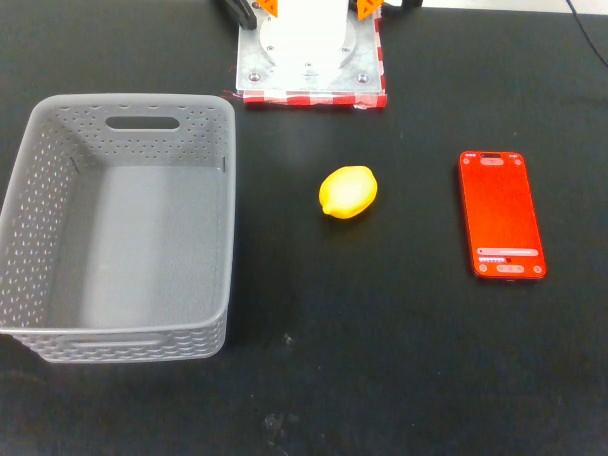} & \bcimg{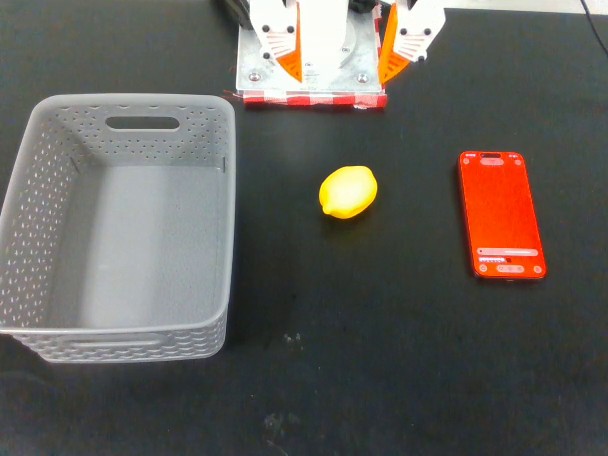} & \bcimg{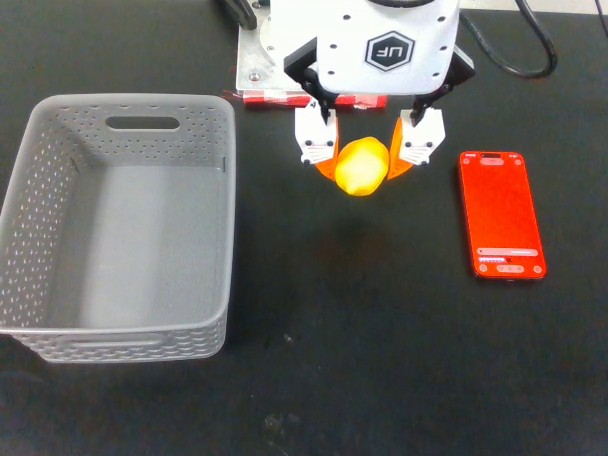} & \bcimg{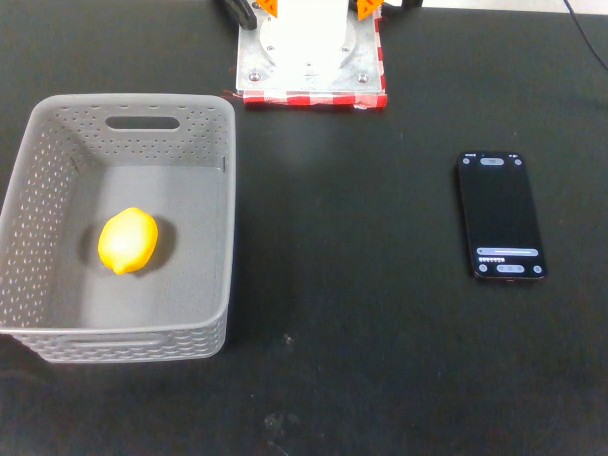} \\
Gemini 4 Argon & 11.52 & 43.41 & 78.07 & 100 & \bcimg{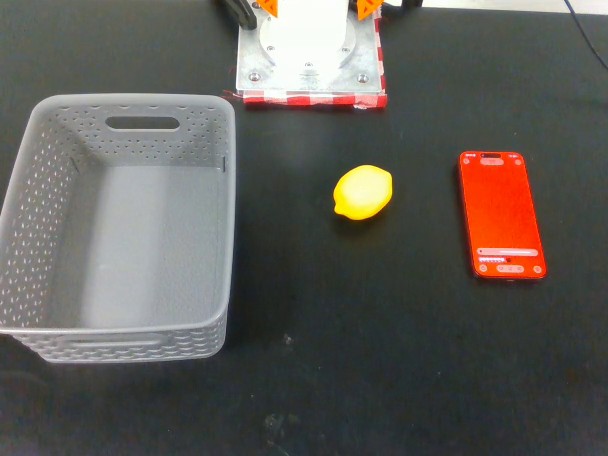} & \bcimg{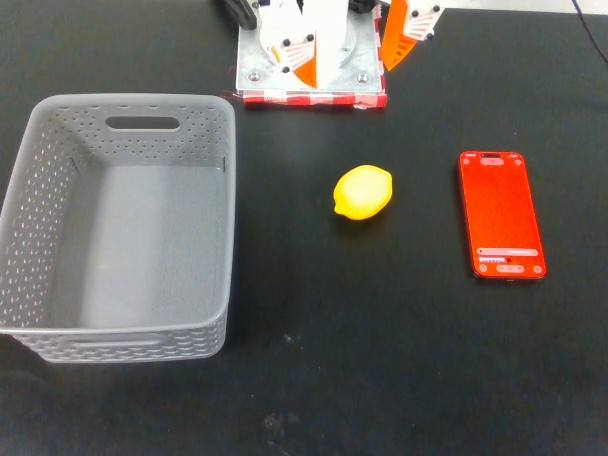} & \bcimg{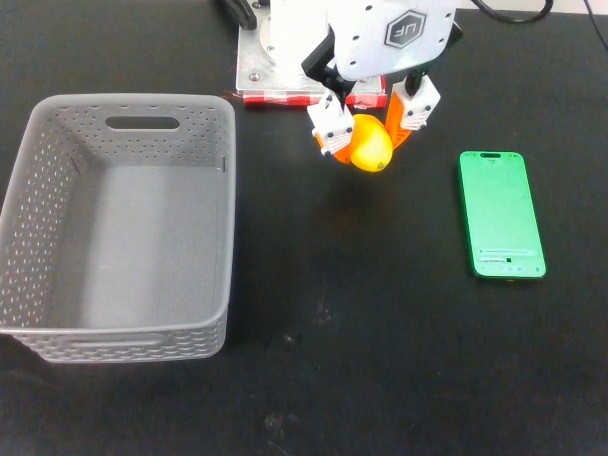} & \bcimg{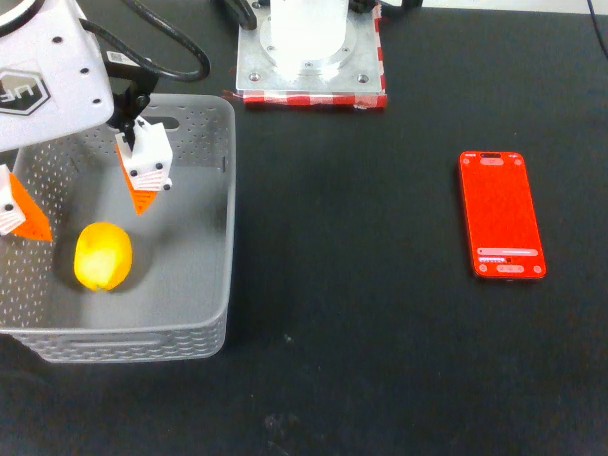} \\
Claude Opus 5.5 & 9.38 & 23.64 & -- & 50 & \bcimg{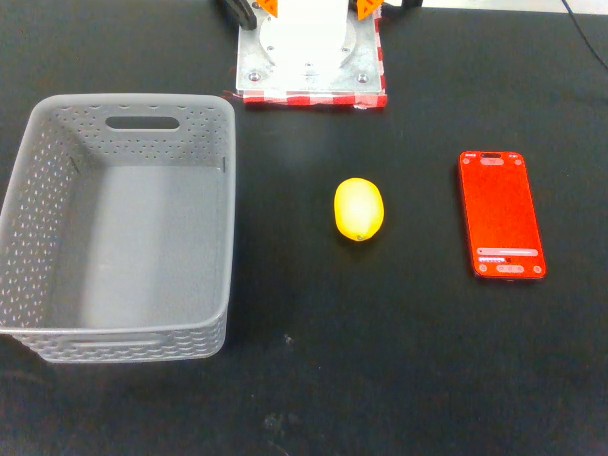} & \bcimg{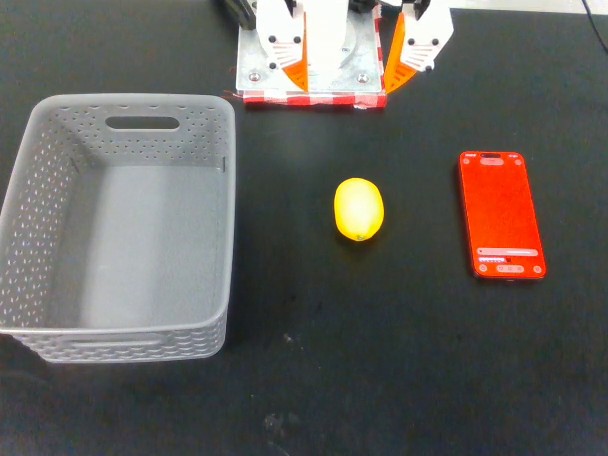} & \bcimg{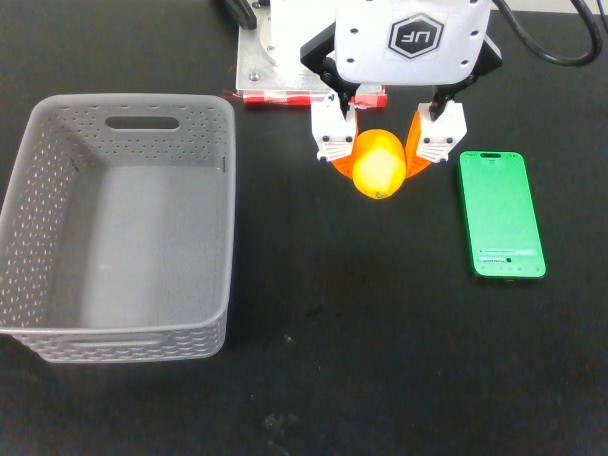} & \bcimg{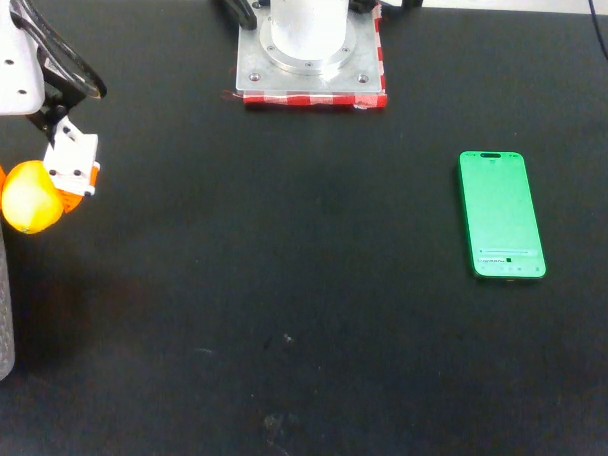} \\
Qwen3.8 27B & 10.01 & 20.93 & -- & 50 & \bcimg{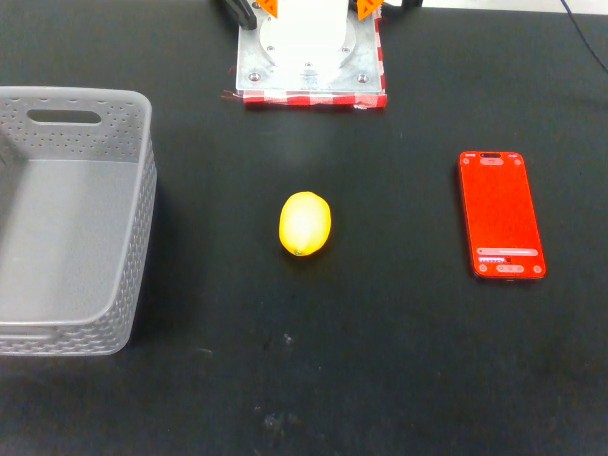} & \bcimg{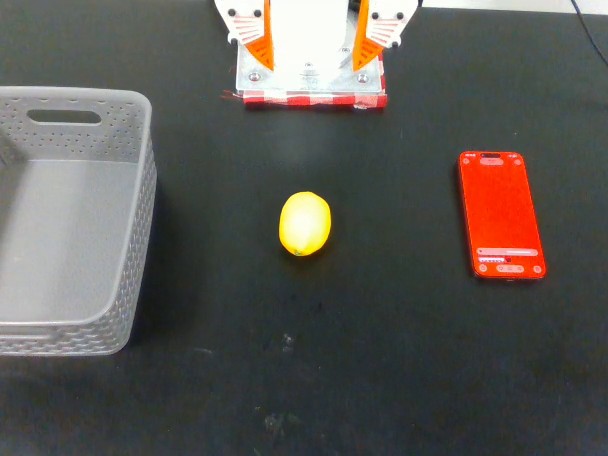} & \bcimg{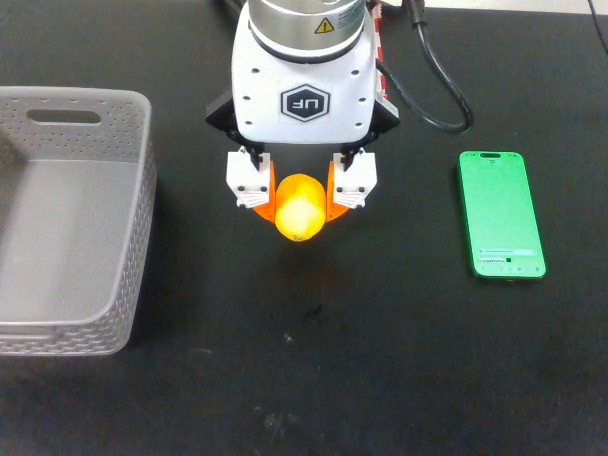} & \bcimg{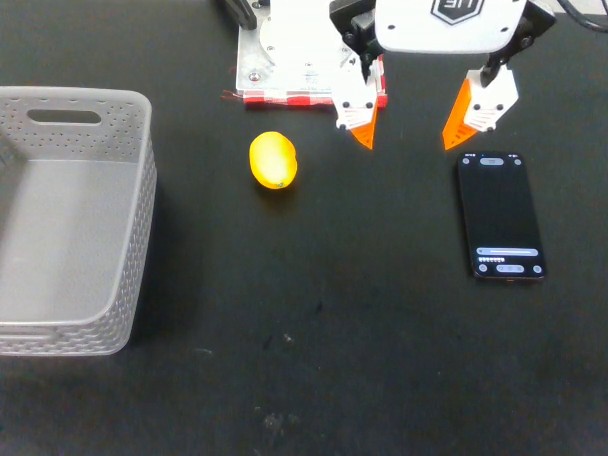} \\
GPT-6.1 Sol & 9.62 & 22.99 & 45.77 & 100 & \bcimg{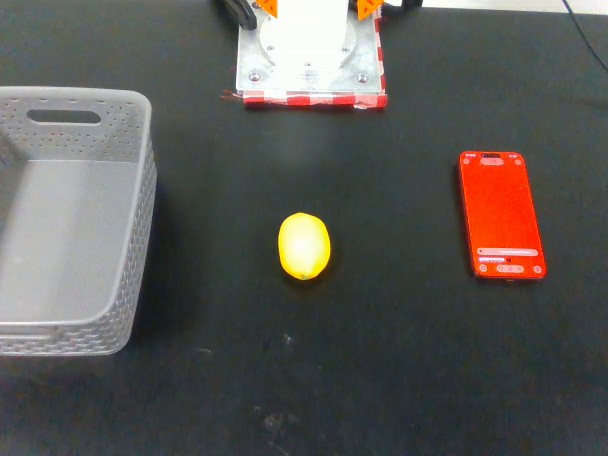} & \bcimg{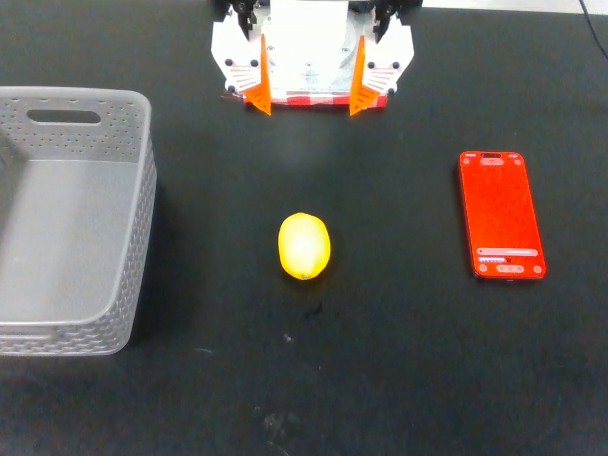} & \bcimg{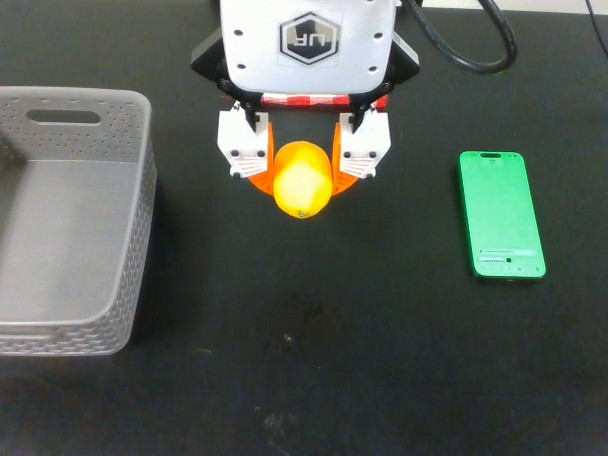} & \bcimg{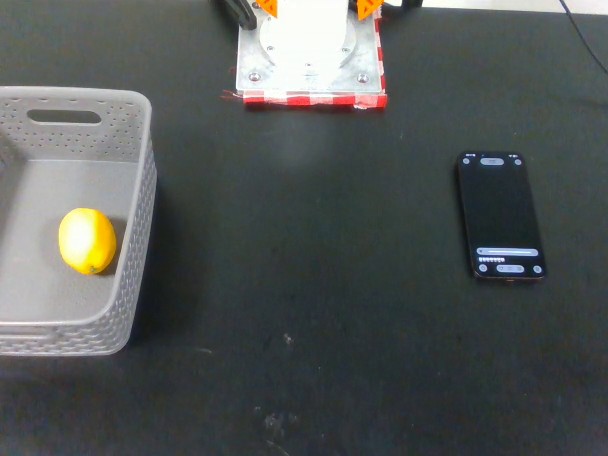} \\
\bottomrule
\end{tabular}
\end{table*}

\subsection{Scoring and Results}
\label{app:real-results}
The operator scores every episode with the rubric of Tab.~\ref{tab:real-rubric}: PS sums the step points up to 100, PSR is the share reaching the final step, and LTSR the share that also meets the temporal requirement (for A1, stopping the pour before exceeding 100\,mL). TSR is undefined on hardware. Each model runs 10 episodes per task, 360 in total.

\begin{table*}[t]
\centering\footnotesize
\setlength{\tabcolsep}{4pt}
\caption{Step scoring of the real-robot tasks. Points add up to a PS of 100 per episode.}
\label{tab:real-rubric}
\begin{tabular}{@{}l p{0.41\linewidth} @{\hspace{14pt}} l p{0.41\linewidth}@{}}
\toprule
Task & Steps and points & Task & Steps and points \\
\midrule
A1 & Can picked up: 25. Held over the bowl: 25. Tilted: 25. Water flows into the bowl and the can is returned: 25. & B2 & Fingertips aligned over the alarm clock: 50. Top pressed: 50. \\
A2 & Object picked up: 40. Held over the container: 40. Dropped into the container: 20. & C1 & Per fruit (four): picked up 10; placed in the basket 15. \\
B1 & Grasped after the red cue: 50. Held over the basket: 20. Dropped into the basket: 30. & C2 & Per peg (eight): picked up 5; inserted 5. Per column holding two pegs of one color (four): 5. \\
\bottomrule
\end{tabular}
\end{table*}

\begin{table*}[t]
\centering\scriptsize
\setlength{\tabcolsep}{2.6pt}
\caption{Real-robot results of six VLMs, scored by the operator with the rubric of Tab.~\ref{tab:real-rubric}. PS: mean step score; PSR: episodes that complete the task (\%); LTSR: episodes that complete it while satisfying the temporal requirement, including the volume-defined stopping condition of A1 (\%). Each entry is the mean over 10 episodes. Mean: unweighted mean over the six tasks. \textbf{Bold}: best per column.}
\label{tab:real-results}
\begin{tabular}{@{}l *{7}{ccc}@{}}
\toprule
& \multicolumn{3}{c}{A1} & \multicolumn{3}{c}{A2} & \multicolumn{3}{c}{B1} & \multicolumn{3}{c}{B2} & \multicolumn{3}{c}{C1} & \multicolumn{3}{c}{C2} & \multicolumn{3}{c}{Mean} \\
\cmidrule(lr){2-4}\cmidrule(lr){5-7}\cmidrule(lr){8-10}\cmidrule(lr){11-13}\cmidrule(lr){14-16}\cmidrule(lr){17-19}\cmidrule(l){20-22}
Model & PS & PSR & LTSR & PS & PSR & LTSR & PS & PSR & LTSR & PS & PSR & LTSR & PS & PSR & LTSR & PS & PSR & LTSR & PS & PSR & LTSR \\
\midrule
Gemini 4 Argon & 37.5 & 20 & 20 & \textbf{100} & \textbf{100} & 0 & \textbf{94} & \textbf{80} & 0 & \textbf{75} & \textbf{50} & \textbf{50} & 12.5 & 0 & 0 & \textbf{3.5} & 0 & 0 & 53.8 & \textbf{41.7} & \textbf{11.7} \\
GPT-6 Astra & \textbf{60} & \textbf{30} & \textbf{30} & \textbf{100} & \textbf{100} & 0 & 79 & 30 & 0 & 65 & 30 & 30 & \textbf{21.5} & 0 & 0 & 2.5 & 0 & 0 & \textbf{54.7} & 31.7 & 10.0 \\
GPT-6.1 Sol & 0 & 0 & 0 & 50 & 10 & 0 & 80 & 40 & 0 & 40 & 10 & 0 & 3 & 0 & 0 & 0 & 0 & 0 & 28.8 & 10.0 & 0.0 \\
Claude Opus 5.5 & 0 & 0 & 0 & 82 & 70 & 0 & 40 & 0 & 0 & 20 & 0 & 0 & 0 & 0 & 0 & 0 & 0 & 0 & 23.7 & 11.7 & 0.0 \\
Qwen3.8-27B & 0 & 0 & 0 & 12 & 0 & 0 & 30 & 0 & 0 & 10 & 0 & 0 & 15 & 0 & 0 & 0 & 0 & 0 & 11.2 & 0.0 & 0.0 \\
DeepSeek V4.1 Flash & 0 & 0 & 0 & 24 & 0 & 0 & 5 & 0 & 0 & 0 & 0 & 0 & 0 & 0 & 0 & 0 & 0 & 0 & 4.8 & 0.0 & 0.0 \\
\bottomrule
\end{tabular}
\end{table*}

\begin{figure*}[t]
\centering
\includegraphics[width=\linewidth]{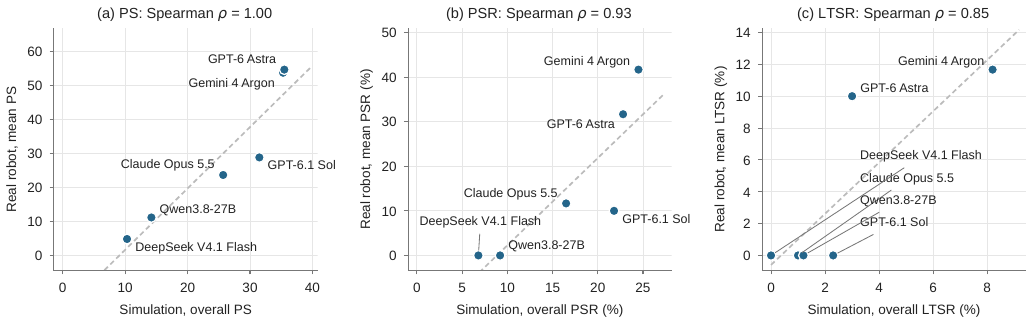}
\caption{Real-robot scores against simulated scores for the six VLMs run in both settings. Horizontal axis: simulated Overall, from Tab.~\ref{tab:api-latest} for Gemini~4~Argon, GPT-6~Astra, GPT-6.1~Sol, and Claude~Opus~5.5, and from Tab.~\ref{tab:agent-results} for Qwen3.8-27B and DeepSeek~V4.1~Flash. Vertical axis: mean over the real-robot tasks (Tab.~\ref{tab:real-results}). Dashed line: least-squares fit through the six points. $\rho$: Spearman rank correlation, with tied models sharing their mean rank.}
\label{fig:real-vs-sim}
\end{figure*}

\paragraph{Results.}
No model completes a Throughput task (best PS 21.5 on C1), while Gemini~4~Argon and GPT-6~Astra solve A2 in every episode (Tab.~\ref{tab:real-results}). On A2 and B1 every completed episode exceeds its time requirement, so LTSR is nonzero only on A1 and B2, and only for these two models. Mean PS is about 54 for the leading two, 28.8 and 23.7 for GPT-6.1~Sol and Claude~Opus~5.5, and 11.2 and 4.8 for the open-weight models.

\paragraph{Comparison with the simulated ranking.}
Fig.~\ref{fig:real-vs-sim} compares the real-robot means with the simulated Overall values: Tab.~\ref{tab:api-latest} for Gemini~4~Argon, GPT-6~Astra, GPT-6.1~Sol, and Claude~Opus~5.5, and Tab.~\ref{tab:agent-results} for Qwen3.8-27B and DeepSeek~V4.1~Flash. The rankings agree, with Spearman $\rho=1.00$ for PS, 0.93 for PSR, and 0.85 for LTSR. In LTSR, Gemini~4~Argon leads GPT-6~Astra in both settings, while the other four models score at most 2.3\% in simulation and zero on the arm. Absolute rates and interface versions differ across episodes, so this pilot comparison is descriptive and does not establish a mapping of success rates.

\end{document}